\documentclass{article}

\usepackage{arxiv}

\usepackage[utf8]{inputenc} 
\usepackage[T1]{fontenc}    
\usepackage{hyperref}       
\usepackage{url}            
\usepackage{booktabs}       
\usepackage{amsfonts}       
\usepackage{nicefrac}       
\usepackage{microtype}      
\usepackage{lipsum}		
\usepackage{graphicx}
\usepackage{natbib}
\usepackage{doi}
\usepackage{mathtools}
\usepackage{cleveref}
\usepackage{multirow}
\usepackage{subcaption}
\usepackage{pgfplotstable}
\usepackage{colortbl}
\usepackage{xcolor}
\usepackage{csvsimple}
\usepackage{colortbl}
\definecolor{lightgray}{gray}{ .9}
\definecolor{lightgreen}{HTML}{C6EFCE}
\definecolor{lightred}{HTML}{FFC7CE}
\usepackage{longtable}
\usepackage{pdflscape} 
\usepackage{pgf}
\usepackage{collcell}
\usepackage{rotating}
\usepackage{multicol}
\usepackage{adjustbox}

\usepackage{verbatim}
\usepackage{algorithm}
\usepackage{algorithmic}

\pgfplotsset{compat=1.18}
\usepackage{placeins}

\title{Content-based image retrieval for multi-class volumetric radiology images: a benchmark study}

\author{ Farnaz Khun Jush \\ 
\textit{Bayer AG}, Berlin, Germany \\
farnaz.khunjush@bayer.com 
\And
Steffen Vogler \\
\textit{Bayer AG}, Berlin, Germany \\
steffen.vogler@bayer.com
\And
Tuan Truong \\
\textit{Bayer AG}, Berlin, Germany \\
tuan.truong@bayer.com
\And
Matthias Lenga \\
\textit{Bayer AG}, Berlin, Germany\\
matthias.lenga@bayer.com
}

\begin{document}
\maketitle
\begin{abstract}

 While content-based image retrieval (CBIR) has been extensively studied in natural image retrieval, its application to medical images presents ongoing challenges, primarily due to the 3D nature of medical images. 
 Recent studies have shown the potential use of pre-trained vision embeddings for CBIR in the context of radiology image retrieval. 
 However, a benchmark for the retrieval of 3D volumetric medical images is still lacking, hindering the ability to objectively evaluate and compare the efficiency of proposed CBIR approaches in medical imaging. 
 In this study, we extend previous work and establish a benchmark for region-based and localized multi-organ retrieval using the TotalSegmentator dataset (TS) with detailed multi-organ annotations. 
 We benchmark embeddings derived from pre-trained supervised models on medical images against embeddings derived from pre-trained unsupervised models on non-medical images for 29 coarse and 104 detailed anatomical structures in volume and region levels. 
 For volumetric image retrieval, we adopt a late interaction re-ranking method inspired by text matching. We compare it against the original method proposed for volume and region retrieval and achieve a retrieval recall of 1.0 for diverse anatomical regions with a wide size range. 
 The findings and methodologies presented in this paper provide insights and benchmarks for further development and evaluation of CBIR approaches in the context of medical imaging.

\end{abstract}

\keywords{Content-based image retrieval \and Medical imaging \and Pre-trained embeddings \and Re-ranking}

\section{Introduction}

In the realm of computer vision, content-based image retrieval (CBIR) has been the subject of extensive research for several decades \citep{dubey2021decade}. 
CBIR systems typically utilize low-dimensional image representations stored in a database and subsequently retrieve similar images based on distance metrics or similarity measures of the image representations. 
Early approaches to CBIR involved manually crafting distinctive features, which led to a semantic gap, resulting in the loss of crucial image details due to the limitations of low-dimensional feature design \citep{dubey2021decade, wang2022two}. 
However, recent studies in deep learning have redirected attention towards the creation of machine-generated discriminative feature spaces, effectively addressing and bridging this semantic gap \citep{qayyum2017medical}. 
This shift has significantly enhanced the potential for more accurate and efficient CBIR methods \citep{dubey2021decade}.

While natural image retrieval has been extensively researched, the application of retrieval frameworks to medical images, particularly radiology images, presents ongoing challenges.
CBIR offers numerous advantages for medical images. 
Radiologists can utilize CBIR to search for similar cases, enabling them to review the history, reports, patient diagnoses, and prognoses, thereby enhancing their decision-making process. 
In real-world use-cases, we often encounter huge anonymized and unannotated datasets available from different studies or institutions where the available meta-information, such as DICOM header data, has been removed or is inconsistent. Manually, searching for relevant images in such databases is extremely time-consuming. 
Moreover, the development of new tools and research in the medical field requires trustable dataset sources and therefore a reliable method for retrieving images, making CBIR an essential component in advancing computer-aided medical image analysis and diagnosis.
One of the key challenges with applying standard CBIR techniques to medical images lies in the fact that algorithms developed for natural images are typically designed for 2D images, while medical images are often 3D volumes which adds a layer of complexity to the retrieval process. 

Recent studies have proposed and demonstrated the potential use of pre-trained vision embeddings for CBIR in the context of radiology image retrieval \citep{jush2023medical, abacha20233d, denner2024leveraging, truong2023benchmarking}. 
However, these studies have primarily focused on 2D images \citep{denner2024leveraging} or specific pathologies or tasks \citep{abacha20233d, jush2023medical, truong2023benchmarking}, overlooking the presence of multiple organs in the volumetric images, which is a critical aspect of real-world scenarios. 
Large multi-organ medical image datasets can be leveraged to thoroughly evaluate the efficacy of the proposed methods, enabling a more comprehensive assessment of CBIR approaches for radiology images.
Despite previous efforts, there is still no established benchmark available for comparing methods for the retrieval of 3D volumetric medical images. 
This absence of a benchmark impedes the ability to objectively evaluate and compare the efficiency of the proposed CBIR approaches in the context of medical imaging.

\begin{figure*}[!t]
  \centering
  \includegraphics[trim=1cm 3.0cm .1cm 5.6cm, clip,width=\textwidth]{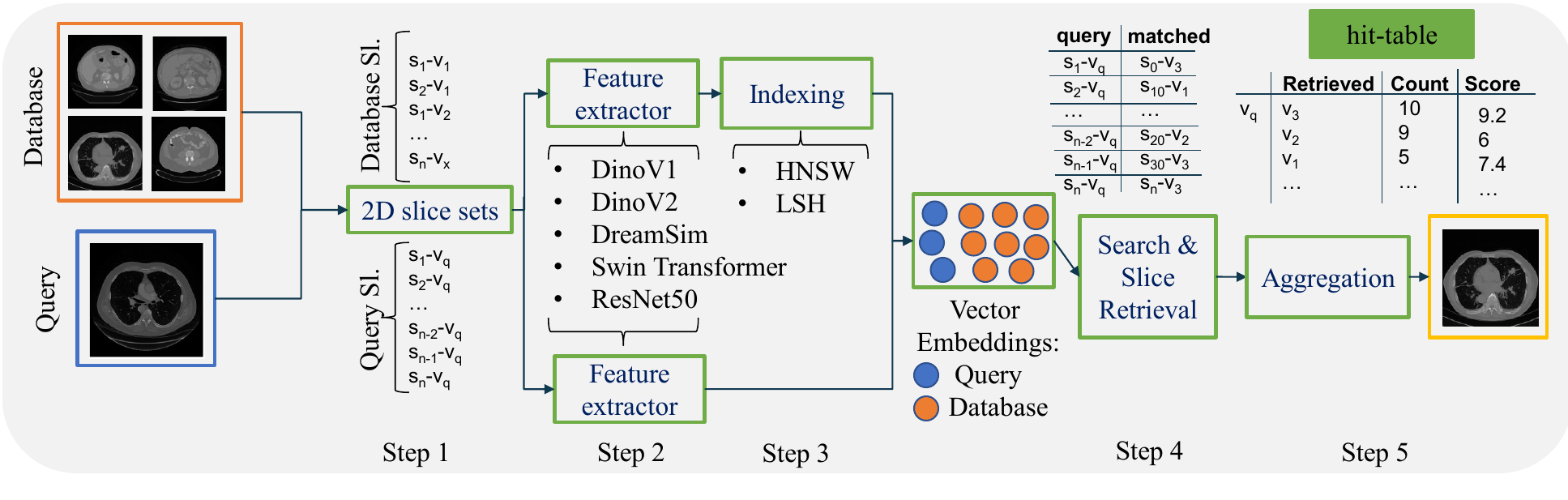}
  \caption{Overview of a retrieval system based on \cite{jush2023medical}: Step 1: 2D slices are extracted from the 3D volumes. Step 2: Feature extractors are used to extract the embeddings from the database slices and query volumes. Step 3: Database embeddings are indexed using HNSW or LSH indexing. Step 4: Search and slice retrieval is performed, and a hit-table is saved (the hit-table shows the occurrence of volume-ids per each query volume or region saved along with the sum of its total score).  Step 5: The results from slice retrieval are aggregated to retrieve the final volume.}
  \label{fig:overview-image}
\end{figure*}

Our previous work \citep{jush2023medical} demonstrated the potential of utilizing pre-trained embeddings, originally trained on natural images, for various medical image retrieval tasks using the Medical Segmentation Decathlon Challenge (MSD) dataset \citep{antonelli2022medical}. 
The approach is outlined in \Cref{fig:overview-image}. Building upon this, the current study extends the methodology proposed in \cite{jush2023medical} to establish a benchmark for anatomical region-based and localized multi-organ retrieval. 
While the focus of \cite{jush2023medical} was on evaluating the feasibility of using 2D embeddings and benchmarking different aggregation strategies of 2D information for 3D medical image retrieval within the context of the single-organ MSD dataset \citep{antonelli2022medical}, it was observed that the single-organ labeling, hinders the evaluations for images containing multiple organs. 
The main objective of this study is to set a benchmark for organ retrieval at the localized level, which is particularly valuable in practical scenarios, such as when users zoom in on specific regions of interest to retrieve similar images of the precise organ under examination. 
To achieve this, we evaluate a count-based method in regions using the TotalSegmentator dataset (TS) \citep{wasserthal2023totalsegmentator}.
TS dataset along with its detailed multi-organ annotations is a valuable resource for medical image analysis and research. 
This dataset provides comprehensive annotations for 104 organs or anatomical structures, which allow us to derive fine-grained retrieval tasks and comprehensively evaluate the proposed methods.

The contribution of this work is as follows: 
 
\begin{itemize}
    \item We benchmarked pre-trained 2D embeddings trained supervised on medical images against self-supervised pre-trained embeddings trained on non-medical images for 3D radiology image retrieval. We utilize a count-based method to aggregate search results based on slice similarity to volume-level data retrieval.
 
     \item We propose evaluation schemes based on the TotalSegmentator dataset \cite{wang2022two} for 29 aggregated coarse anatomical regions and all 104 original anatomical regions. Our proposed evaluation assesses the capabilities of a 3D image search system at different levels, including a fine-grained measure related to the localization of anatomical regions.
 
    
    \item We adopted a late interaction re-ranking method originally used for text retrieval called ColBERT \citep{khattab2020colbert} for volumetric image retrieval. For a 3D image query, this two-stage method generates a candidate 3D image result list utilizing a fast slice-wise similarity search and count-based aggregation. In the second stage, the full similarity information between all query and candidate slices is aggregated to determine re-ranking scores.
    
    \item We benchmarked the proposed re-ranking method against the original method proposed in \cite{jush2023medical} for volume, region, and localized retrieval on 29 modified coarse anatomical regions and 104 original anatomical regions from TS dataset \cite{wang2022two}.
\end{itemize}

\section{Materials and Methods}

\subsection{Vector Database and Indexing}

In the context of image search a database typically constitutes the central location where all the representations of the images, a.k.a. embeddings, and their metadata including annotations are stored.
A query allows the user or the system to request specific images in various ways, e.g., by inputting a reference image or a textual description.
The goal is to search the database for similar images that match the query. 
Similarly, in this study, the search process entails comparing a query image with images in the database to identify the most similar image using the cosine similarity of the embeddings. Throughout this process, we do not depend on any metadata information at any stage. Metadata-independence is an intended design choice and in contrast to widely used metadata-based image retrieval solutions that frequently lack the necessary specificity in real-world retrieval applications. 
In small sets, the similarity search is easy but with the growing size of the database, the complexity increases. Accuracy and speed are the key factors in search, thus, naive approaches typically fail in huge datasets.

Indexing in the context of content-based image search involves creating a structured system that allows for efficient storage and retrieval of images based on their visual content. 
A flat index is the simplest form of indexing, where no modification is made to the vectors before they are used for search. 
In flat indexing, the query vector is compared to every other full-size vector in the database and their distances are calculated. 
The nearest k of the searched spaces is then returned as the k-nearest neighbors (kNN). 
While this method is the most accurate, it comes at the cost of significant search time \citep{aumuller2020ann}. 
To improve search time, two approaches can be employed: reducing the vector size through dimensionality reduction, e.g., by reducing the number of bits representing each vector, or reducing the search scope by clustering or organizing vectors into tree structures based on similarity or distance. 
This results in the identification of an approximation of the true nearest neighbors, known as approximate nearest neighbor search (ANN) \citep{aumuller2020ann}.

There are several ANN methods available.  In the context of content-based volumetric medical image retrieval, \cite{jush2023medical} compared Locality Sensitive Hashing (LSH) \cite{charikar2002similarity} and Hierarchical Navigable Small World (HNSW) \cite{malkov2018efficient} for indexing and search. 
LSH hashes data points in a way that similar data points are mapped to the same buckets with higher probabilities. 
This allows for a more efficient search for nearest neighbors by reducing the number of candidates to be examined.
HNSW \citep{malkov2018efficient} indexing organizes data into a hierarchical graph structure where each layer of the hierarchy has a lower resolution. The top layer connects data points directly, but the lower layers have fewer connections. The graph structure is designed to 
allow for efficient navigation during the search. 
Compared to LSH, HNSW typically enables faster search and requires less memory \cite{taha2024study}. 
Based on findings in \citep{jush2023medical} HSNW was chosen as the indexing method in the setting of this study due to speed advantages over LSH at a comparable recall. 
There are various index solutions available to store and search vectors. In this study, we used the Facebook AI Similarity Search (FAISS) package that enables fast similarity search \citep{johnson2019billion}. 
The indexing process involves running the feature extractors on slices of each volumetric image and storing the output embeddings per slice. The produced representations are then added to the search index which is used later on for vector-similarity-based retrieval.

\subsection{Feature Extractors}
\label{sec:feature_extractor}

We extend the analysis of \cite{jush2023medical} by adding two ResNet50 embeddings and evaluating the performance of six different slice embedding extractors for CBIR tasks. 
All the feature extractors are based on deep-learning-based models.

\textbf{Self-supervised Models}: We employed three self-supervised models pre-trained on ImageNet \citep{deng2009imagenet}.
DINOv1 \citep{caron2021emerging}, that demonstrated learning efficient image representations from unlabeled data using self-distillation.
DINOv2 \citep{oquab2023dinov2}, is built upon DINOv1 \citep{caron2021emerging}, and this model scales the pre-training process by combining an improved training dataset, patchwise objectives during training and introducing a new regularization technique, which gives rise to superior performance on segmentation tasks.
DreamSim \citep{fu2023dreamsim}, built upon the foundation of DINOv1 \citep{caron2021emerging}, fine-tunes the model using synthetic data triplets specifically designed to be cognitively impenetrable with human judgments.
For the self-supervised models, we used the best-performing backbone reported by the developers of the models. 

\textbf{Supervised Models}: We included a SwinTransformer model \citep{liu2021swin} and a ResNet50 model \citep{he2016deep} trained in a supervised manner using the RadImageNet dataset \citep{mei2022radimagenet} that includes 5 million annotated 2D CT, MRI, and ultrasound images of musculoskeletal, neurologic, oncologic, gastrointestinal, endocrine, and pulmonary pathology. 
Furthermore, a ResNet50 model pre-trained on rendered images of fractal geometries was included based on \citep{KataokaIJCV2022}. These training images are formula-derived, non-natural, and do not require any human annotation.

\subsection{Dataset and Pre-processing}

\begin{table}[!t]
    \centering
      \caption{Mapping of the original TS classes to 29 coarse anatomical regions.}
    \tiny
    \begin{tabular}{l l||l l||l l}
    \hline
        \textbf{Anatomical region} & \textbf{Mapped class } & \textbf{Anatomical region} & \textbf{ Mapped class}  & \textbf{Anatomical region} & \textbf{ Mapped class}  \\  \hline
          adrenal gland left &  adrenal gland  &   iliopsoas right &  iliopsoas  &   rib right 11 &  rib  \\  
          adrenal gland right &  adrenal gland  &   inferior vena cava &  cardiovascular system  &   rib right 12 &  rib  \\  
          aorta &  cardiovascular system  &   kidney left &  kidney  &   sacrum &  sacrum  \\  
          autochthon left &  autochthon  &   kidney right &  kidney  &   scapula left &  scapula  \\  
          autochthon right &  autochthon  &   liver &  liver  &   scapula right &  scapula  \\  
          brain &  brain  &   lung lower lobe left &  lung  &   small bowel &  small bowel  \\  
          clavicula left &  clavicula  &   lung lower lobe right &  lung  &   spleen &  spleen  \\  
          clavicula right &  clavicula  &   lung middle lobe right &  lung  &   stomach &  stomach  \\  
          colon &  colon  &   lung upper lobe left &  lung  &   trachea &  trachea  \\  
          duodenum &  duodenum  &   lung upper lobe right &  lung  &   urinary bladder &  urinary bladder  \\  
          esophagus &  esophagus  &   pancreas &  pancreas  &   vertebrae C1 &  vertebrae  \\  
          face &  face  &   portal and splenic vein &  portal \& splenic vein  &   vertebrae C2 &  vertebrae  \\  
          femur left &  femur  &   pulmonary artery &  cardiovascular system  &   vertebrae C3 &  vertebrae  \\  
          femur right &  femur  &   rib left 1 &  rib  &   vertebrae C4 &  vertebrae  \\  
          gallbladder &  gallbladder  &   rib left 2 &  rib  &   vertebrae C5 &  vertebrae  \\  
          gluteus maximus left &  gluteus muscles  &   rib left 3 &  rib  &   vertebrae C6 &  vertebrae  \\  
          gluteus maximus right &  gluteus muscles  &   rib left 4 &  rib  &   vertebrae C7 &  vertebrae  \\  
          gluteus medius left &  gluteus muscles  &   rib left 5 &  rib  &   vertebrae L1 &  vertebrae  \\  
          gluteus medius right &  gluteus muscles  &   rib left 6 &  rib  &   vertebrae L2 &  vertebrae  \\  
          gluteus minimus left &  gluteus muscles  &   rib left 7 &  rib  &   vertebrae L3 &  vertebrae  \\  
          gluteus minimus right &  gluteus muscles  &   rib left 8 &  rib  &   vertebrae L4 &  vertebrae  \\  
          heart atrium left &  cardiovascular system  &   rib left 9 &  rib  &   vertebrae L5 &  vertebrae  \\  
          heart atrium right &  cardiovascular system  &   rib left 10 &  rib  &   vertebrae T1 &  vertebrae  \\  
          heart myocardium &  cardiovascular system  &   rib left 11 &  rib  &   vertebrae T2 &  vertebrae  \\  
          heart ventricle left &  cardiovascular system  &   rib left 12 &  rib  &   vertebrae T3 &  vertebrae  \\  
          heart ventricle right &  cardiovascular system  &   rib right 1 &  rib  &   vertebrae T4 &  vertebrae  \\  
          hip left &  hip  &   rib right 2 &  rib  &   vertebrae T5 &  vertebrae  \\  
          hip right &  hip  &   rib right 3 &  rib  &   vertebrae T6 &  vertebrae  \\  
          humerus left &  humerus  &   rib right 4 &  rib  &   vertebrae T7 &  vertebrae  \\  
          humerus right &  humerus  &   rib right 5 &  rib  &   vertebrae T8 &  vertebrae  \\  
          iliac artery left &  cardiovascular system  &   rib right 6 &  rib  &   vertebrae T9 &  vertebrae  \\  
          iliac artery right &  cardiovascular system  &   rib right 7 &  rib  &   vertebrae T10 &  vertebrae  \\  
          iliac vena left &  cardiovascular system  &   rib right 8 &  rib  &   vertebrae T11 &  vertebrae  \\  
          iliac vena right &  cardiovascular system  &   rib right 9 &  rib  &   vertebrae T12 &  vertebrae  \\ 
          iliopsoas left &  iliopsoas  &   rib right 10 &  rib  & ~ & ~ \\ 
    \label{tab: mapping table}
    \end{tabular}    
\end{table}

We designed a CBIR benchmark relying on the publicly available TotalSegmentator (TS) dataset \cite{wasserthal2023totalsegmentator}, version 1. 
This dataset comprises in total of 1204 computed tomography (CT) volumes covering 104 anatomical structure annotations (TS, V1). 
The anatomical regions presented in the original dataset include several fine-grained sub-classes for which we considered an aggregation to a coarser common class as a reasonable measure, e.g., all the rib classes are mapped to a single class `rib'. 
The coarse organ labels can help identify similarities and potential mismatches between neighboring anatomical regions, providing valuable insights into the proximity information of the target organ.
\Cref{tab: mapping table} shows the mapping of the original TS classes to the coarse aggregated classes. 
For the sake of reproducibility, the query cases are sourced from the original TS test split, while the cases contained in the original TS train and validation set serve as the database for searching.
The search is assessed on the retrieval rate of 29 coarse anatomical structures and 104 original TS anatomical structures.

The models presented in \Cref{sec:feature_extractor} are 2D models used without fine-tuning to extract the embeddings. Thus, per each 3D volume, individual 2D slices of the corresponding 3D volumes are utilized for embedding extraction.
The input size for all the used models is equal to $224\times224$ pixels with image replication along the RGB channel axis. 
For all the ViT-based models and the ResNet50 trained on fractal images, images are normalized to the ImageNet mean and standard deviation of $( .485,  .456,  .406)$ and $( .229,  .224,  .225)$, respectively. 
For the SwinTransformer and the ResNet50 model pre-trained on the RadImageNet dataset,  the images are normalized to $ .5$ mean and $ .5$ standard deviation based on \cite{mei2022radimagenet}. 
The total size of the database is $290757$ embeddings, while the final query set of the test set comprises $20442$ embeddings.

\subsection{Search and Retrieval}
\label{sec: search and retrieval}

After creating the vector database, the search is performed using the embeddings extracted from slices of query volumes.
The simplest way of retrieval is to match a 2D query slice $q$ with the most similar 2D slice in the database $s^*$ by finding the slice-embedding that maximizes the cosine similarity with respect to the embedding associated with $q$, i.e.
\begin{equation}\label{eq:similarity_search}
    s^* 
    = \underset{s \in \text{Data}}{\mathrm{argmax}} \frac{\langle\phi(s),\phi(q)\rangle}{\Vert \phi(s) \Vert_2 \Vert \phi(q) \Vert_2}
    = \underset{s \in \text{Data}}{\mathrm{argmax}} \left\langle v_s,\frac{\phi(q)}{\Vert \phi(q) \Vert_2}\right\rangle
\end{equation}
where $\langle\cdot,\cdot\rangle$ denotes standard scalar product, $\Vert\cdot\Vert_2$ the euclidean norm, $\phi$ the embedding mapping and $v_s = \phi(s) / \Vert \phi(s) \Vert_2$ the pre-computed, normalized embedding associated to slice $s$ stored in a vector index.
In \cite{jush2023medical} the slice-wise retrieval was introduced as the lower bound baseline for evaluating the proposed aggregation and sampling schemes. Similarly, in this work, we keep the slice-wise evaluation as the lower bound for the retrieval rate of our methods. 
This method is the lower bound because for each slice only one slice is retrieved and for the perfect recall all the anatomical structures visible in the query slice should match the retrieved slice.
In this baseline, each slice $q$ of the query dataset is considered as an individual search instance. In addition, we performed and evaluated image retrieval in three additional scenarios:

\subsubsection{Volume-based retrieval and evaluation} 
\label{Volume-based retrieval}

\begin{figure}[!t]
  \centering
  \includegraphics[trim=1cm 5.0cm 3cm 5cm, clip, scale=.48]{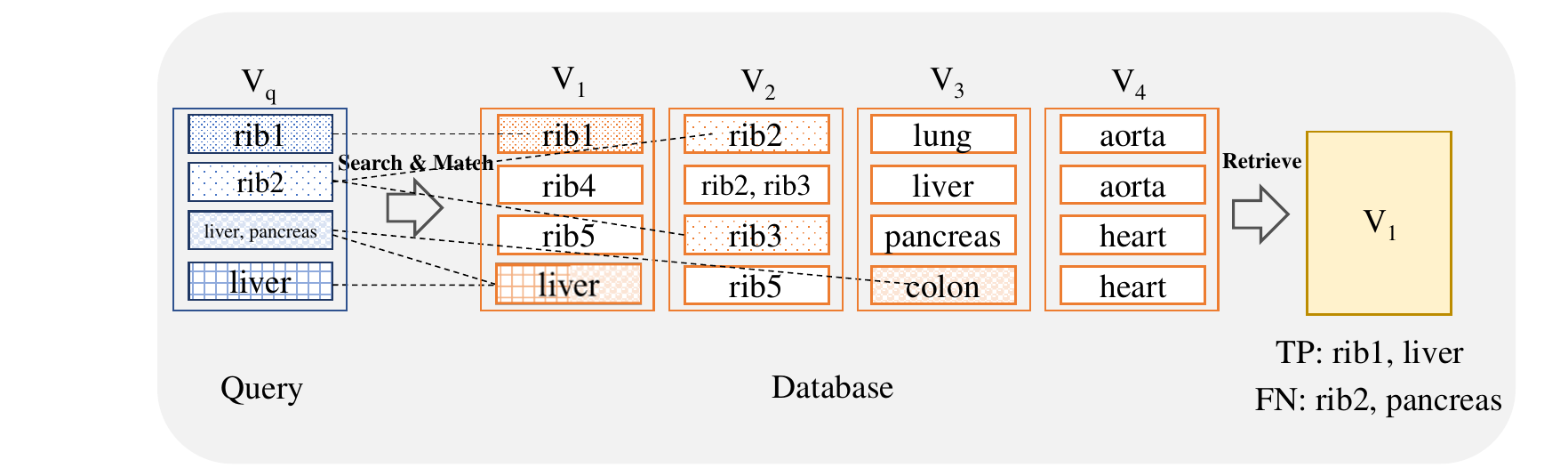}
  \caption{Volume-based retrieval: For a query volume $V_q$ covering a range of anatomical regions, a volume is retrieved that should cover the same anatomical regions. The similarity search is based on all slices from the query volume.}
  \label{fig:volume-based}
\end{figure}

In this setting, we want to assess the capability of the search system to retrieve an image volume that covers the same anatomical regions overall as a given query volume $V_Q = [q_1, ..., q_n]$. 
For every slice $q_i$ from the query volume $V_Q$, the system retrieves the most similar slice $s_i^*$ from the database using \eqref{eq:similarity_search}. Subsequently, the corresponding volume-id and its similarity score are stored in a hit-table similar to the hit-table shown in \Cref{fig:overview-image}.
We then apply the count-based aggregation scheme from \cite{jush2023medical} that uses the hit-table to identify the volume $V_R$ with the most hits given the query volume at hand, see \Cref{fig:overview-image}. The system then returns $V_R$ as the final result of the similarity search given the query volume $V_Q$. \cite{abacha20233d} refers to this method as the retrieval based on frequency.
The evaluation of the search system on the proposed benchmark dataset is done using the measure of Recall (True Positive Rate). 
To this end, the aggregated labels of the query volumes are compared against the aggregated labels of the associated retrieved volumes (see \ref{sec:volume-based-eval}).
The recall in this setting is high if the aggregated query and retrieved volume labels coincide. An overview is shown in \Cref{fig:volume-based}.

\subsubsection{Region-based retrieval and evaluation}\label{sec:region_based_retrieval}

In this setting the search system is queried with an image (sub-)volume which is constrained to a specific anatomical sub-region (e.g. liver, pancreas, heart,...). 
For each anatomical region, we want to individually assess the capability of the system to retrieve an image volume containing the anatomical region.

The query (sub-)volumes for different anatomical regions are generated as follows. Given a selected anatomical region $r$ and a query image volume $V_Q = [q_1, ..., q_n]$, the smallest subset  slices $V_{Q,r} = [q_m, ..., q_{k}] \subset V_Q$ is chosen that entirely contains the anatomical region $r$ visible in $V_Q$. Based on the sub-volume $V_{Q,r}$ a similarity search is conducted to build up a hit-table, and the count-based aggregation is conducted to finally retrieve for this query the volume with most hits, as described in \Cref{Volume-based retrieval}.

\begin{figure}[!t]
  \centering
  \includegraphics[trim=0cm   .01cm 0cm 0cm, clip, width=\textwidth]{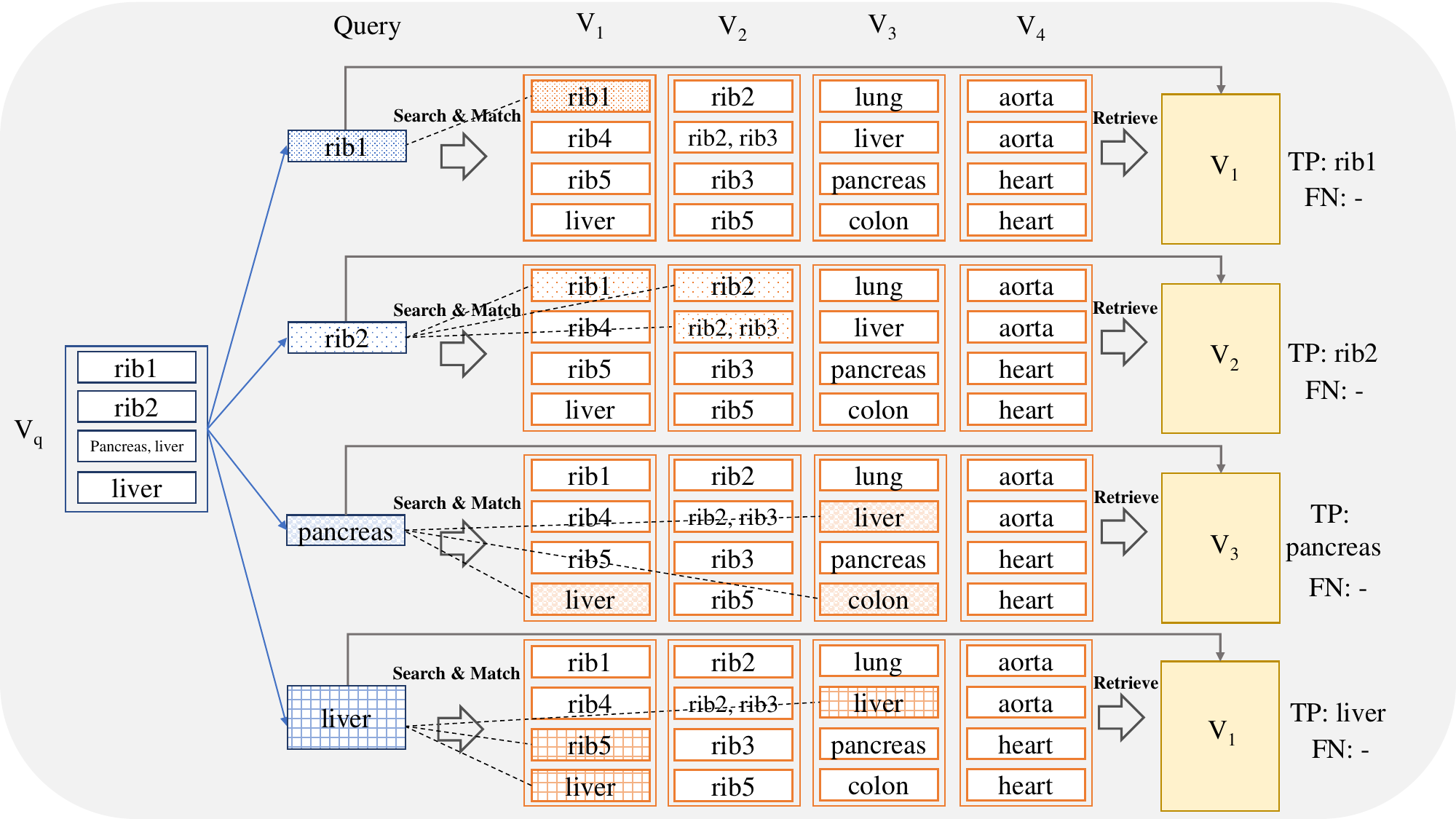}
  \caption{Region-based retrieval. Anatomical regions are considered individually. A sub-volume constrained to an anatomical region of interest $r$ is generated and fed to the search system to retrieve a volume containing the anatomical region. A case is considered a True Positive (TP) if the retrieved case contains the region $r$ at some location.}
  \label{fig:region-based}
\end{figure}

%
%
%

In this scenario, the evaluation is done for each anatomical region individually utilizing again the measure of Recall. 
To this end, for a selected anatomical region $r$ the region-centric query sub-volumes are fed to the search system and the aggregated labels of the associated retrieved volumes are compared to $r$. 
The recall in this setting is high if the aggregated retrieved volume labels contain $r$. Hence, in this evaluation setting it is only required that the retrieved volume contains the anatomical region of interest. 
It is not required that the search system identifies the exact slices where the anatomical region is visible. The overview of this method is depicted in \Cref{fig:region-based}. 
The question of whether the system can exactly localize anatomical regions will be addressed in \Cref{sec:localization}.
For example in \Cref{fig:region-based}, V3 is retrieved for the $r=\text{`pancreas'}$ anatomical sub-region. 
During the evaluation, this instance is classified as a True Positive (TP) because the retrieved volume V3 contains 'pancreas', regardless of whether the matched slices contained 'pancreas'.

\subsubsection{Localized retrieval and evaluation}\label{sec:localization}

\begin{figure}[!t]
  \centering
  \includegraphics[trim=0cm   .0cm 0cm 0cm, clip, width=\textwidth]{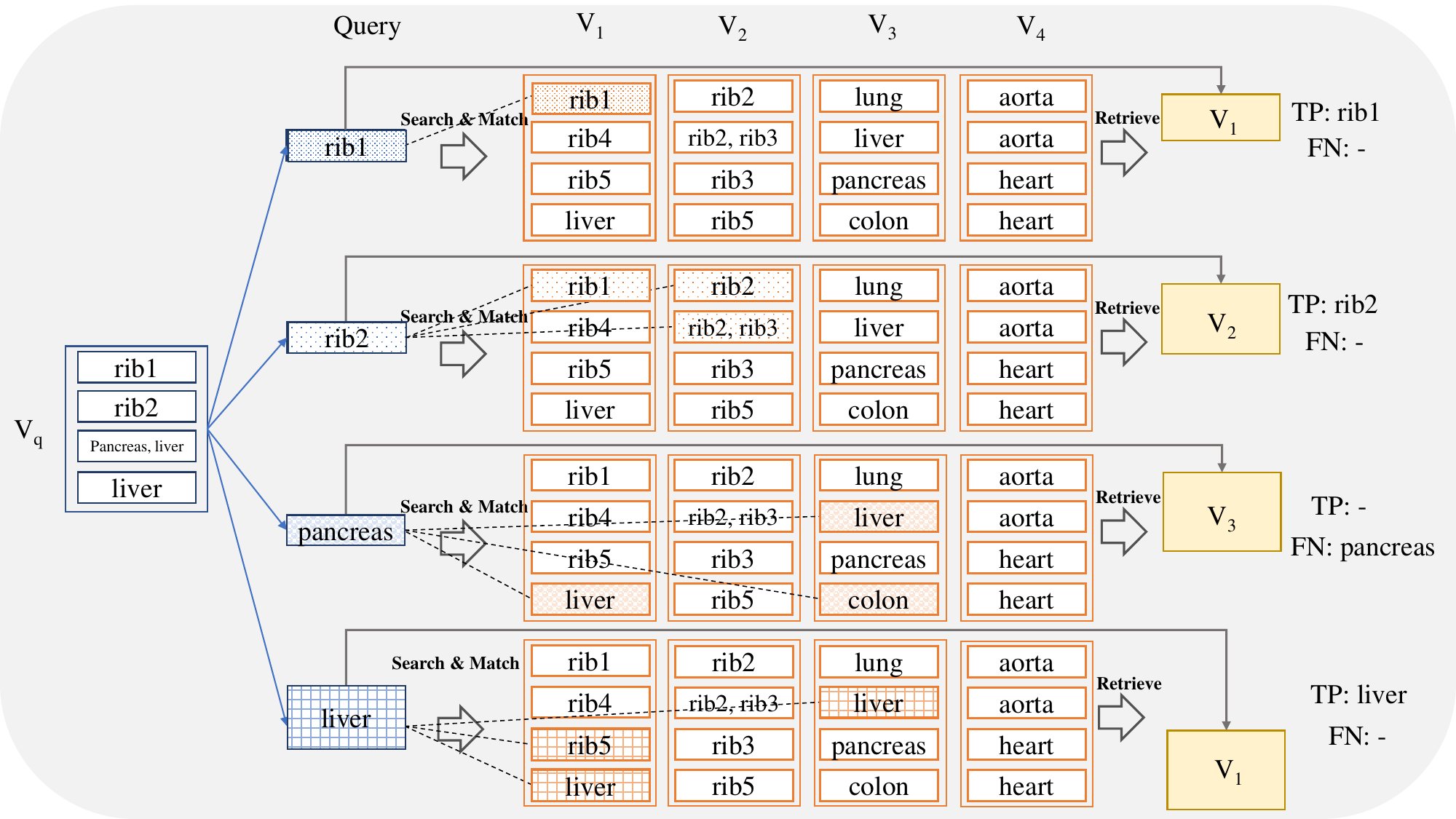}
  \caption{Localized retrieval. Anatomical regions are considered individually. A sub-volume constrained to the
anatomical region of interest $r$ is generated and fed to the search system to retrieve a volume containing the same anatomical region. A case is only considered as True Positive (TP) if at least one of the slices in the retrieved volume contains the region $r$.}
  \label{fig:localized}
\end{figure}

In this setting, the system is queried with an image sub-volume which is constrained to a specific anatomical sub-region (e.g. liver, pancreas, heart,...). 
For each anatomical region, we want to individually assess the capability of the system to retrieve an image volume containing the anatomical region and to localize the region of interest within the retrieved volume.

The query sub-volumes $V_{Q,r}$ for different anatomical regions $r$ are generated as described in detail in \Cref{sec:region_based_retrieval}.
Again, a similarity search is conducted based on the sub-volume $V_{Q,r}$ to retrieve the related volume $V_{R,r}$ with the most hits.
In this scenario, the evaluation is done for each anatomical region individually utilizing again the measure of recall. 
The evaluation criterion is stricter than the region-based evaluation from \Cref{sec:region_based_retrieval}. 
In order to be considered as a True Positive, at least one of the slices from $V_{R,r}$ that occurred in the similarity search must actually intersect with the region $r$. 
In other words, the search system is required to localize $r$ in the sense that at least one slide is identified where $r$ is visible. 
For example, for $r=\text{`pancreas'}$ if a search retrieves a volume that indeed includes the pancreas, but the specific slices hit in the similarity search do not insect the organ, the result is marked as False Negative (FN) in the evaluation, even though the pancreas is present elsewhere in the volume (see \Cref{fig:localized}).
The capability for a search system to localize an anatomical subregion of interest within a retrieved volume is particularly useful for applications with user interaction, e.g. the user marks a subregion in an image and queries the search system to retrieve similar cases from a database and localizes the corresponding subregions therein. 

Another measure to assess the capability of the system to localize a region can be defined as the ratio of the slices that actually contain the subregion $r$ in the retrieved volume to the total number of slices hit in the retrieved volume. In detail, the localization-ratio (LR) is defined as:

\begin{equation} \label{eq:normalized-location-ratio}
\mathrm{LR} =  \frac{\vert \text{slices hit in retrieved vol. that contain r}\vert }{\vert \text{slices hit in retrieved vol.}\vert }
\end{equation}

For example, the query consists of $60 $ slices related to region $r$. The table representing the top 3 volumes hit count is $[48, 21, 4]$. In the volume with the top hit count, 12 out of the 48 hit slices actually contain region $r$, indicating successful localization. The localization-ratio is then given by $12/48=.25$.

\subsection{Re-ranking retrieval and evaluation}
\label{sec: re:ranking}
Re-ranking in information retrieval involves the process of re-ordering the initially retrieved results to better align with the user's information needs. 
This can be achieved through different methods such as relevance feedback, learning to rank algorithms, or incorporating contextual information \citep{ai2018unbiased, guo2020deep, macavaney2019cedr}. 
Relevance feedback allows users to provide input on the initial results, which is then used to adjust the ranking \citep{ai2018unbiased}. 
Learning to rank algorithms utilizes machine learning techniques to re-rank results based on relevant features \citep{guo2020deep}.
Additionally, re-ranking methods may also consider contextual information such as user behavior, temporal relevance, or other relevant factors to better reflect the user's current information needs, ultimately enhancing the overall quality of retrieved results \citep{macavaney2019cedr}.
A method based on contextualized information proposed in \cite{khattab2020colbert} called ColBERT (Contextualized Late Interaction over BERT). 
ColBERT operates by generating contextualized representations of the query and the documents using BERT \citep{devlin2018bert}. 
In this method, queries and documents are encoded into more detailed multi-vector representations, and relevance is gauged through comprehensive yet scalable interactions between these sets of vectors. 
ColBERT creates an embedding for each token in the query and document, and it measures relevance as the total of maximum similarities between each query vector and all vectors within the document \citep{santhanam2021colbertv2}.
This late interaction approach allows for a more refined and contextually aware retrieval process, thereby enhancing the quality of information retrieval.

Inspired by ColBERT we introduce a two-stage method in which filtering of the search space is performed and the total similarity of the entire target volume is considered to re-rank and score the retrieved volumes. 
To create an analogy to the ColBERT method each word can be considered as one slice and each passage of the database or each question of the query can be considered as one volume. 
Instead of the BERT encoder for the image retrieval task, the pre-trained vision models can be used to create the embeddings as discussed in \Cref{sec:feature_extractor}. 

\begin{figure*}[!t]
  \centering
  \includegraphics[trim= .1cm  .1cm .1cm  .5cm, clip, width=\textwidth]{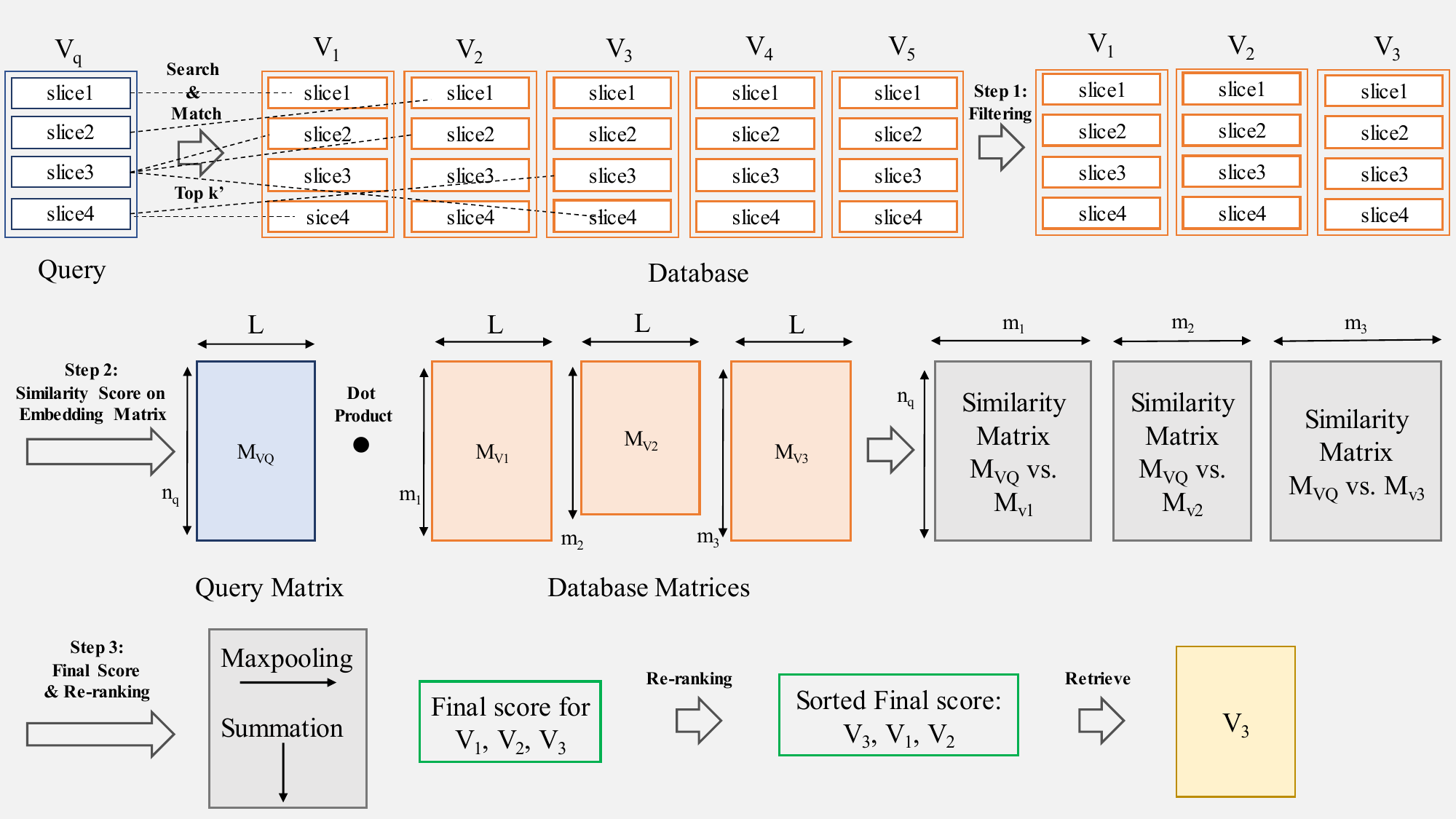}
  \caption{Overview of re-ranking. Step 1: Filtering based on at least one similar slice leads to the selection of candidate volumes Step 2:  followed by similarity score computation using dot product on the normalized embedding matrices. Step 3: The final step involves max-pooling and summation to determine the top-scoring volumes for retrieval.}
  \label{fig: overview}
\end{figure*}

An overview of the proposed method is shown in \Cref{fig: overview}. 
The method consists of the following steps: 

\subsubsection{Step 1: Filtering}
We assume that the embeddings have been computed and stored as vectors for all image data. 
In the first step, for each slice $q_i$ of the query volume $V_Q = [q_1, ..., q_N]$, the most similar slice $s_i^*$ from the database is retrieved via similarity search \eqref{eq:similarity_search}. 
We denote volumes associated to $s_1^*,...,s_N^*$ by $V_1,...,V_N$. 
Clearly, for some indices $i,j$ the retrieved sliced can belong to the same volume, i.e. $V_i=V_j$. We filter out duplicates for our list and let $V_1,..,V_M$ with $M\leq N$ denote the unique volumes associated with the retrieved slices. 
These unique volumes resulting from this initial search stage are considered as candidates for a more compute-intensive re-ranking procedure which takes all similarity interactions between all the volume slices into account.

\subsubsection{Step 2: Similarity Score on Embedding Matrix}
\label{sec: sim score}
We assume that our slice embedding $\phi$ maps each slice to a vector of fixed length $L$. 
Hence, for any image volume $V=[v_1,...,v_n]$ with $n$ slices $v_1,...,v_n$ we can compute the $L_2$-normalized embedding matrix 
\begin{equation}
    M_V = 
    \left[
    \frac{\phi(v_1)}{\Vert\phi(v_1)\Vert_2}, 
    ...,
    \frac{\phi(v_n)}{\Vert\phi(v_n)\Vert_2} 
    \right]
\end{equation}
of dimension $n \times L$. Given another image volume $W=[w_1,...,w_m]$ of size $m \times L$ we can compute the similarity matrix of size $n\times m$ via the matrix dot product:
\begin{equation}\label{eq:similarity_matrix}
    \mathrm{SIM}(M_V, M_W)
    = \left[\frac{\langle \phi(v_i),\phi(w_j) \rangle}{\Vert\phi(v_i)\Vert_2 ~ \Vert\phi(w_j)\Vert_2} \right]_{\substack{i=1,...,n\\j=1,...,m}}
\end{equation}
The entry $(i,j)$ of this matrix contains the cosine-similarity score of the embeddings related to slice $i$ of volume $V$ and slice $j$ of volume $W$. For our re-ranking,  we first compute in this step, the embedding matrix $M_{V_Q}$ for the query volume $V_Q$ and the embedding matrices $M_{V_1},...,M_{V_M}$ for all unique retrieved volumes $V_1,...,V_M$ from Step 1. Then we compute the similarity matrices $\mathrm{SIM}(M_{V_Q}, M_{V_k})$ for $k=1,...,M$.

\subsubsection{Step 3: Final Score and Re-ranking}
To compute the rank score of each volume $V_i$, the associated similarity matrix $\mathrm{SIM}(M_{V_Q}, M_{V_i})$ is max-pooled row-wise which corresponds to determine the slice in $V_i$ with highest cosine similarity to a given slice in $V_Q$. The resulting vector of length $n$ is then summed to obtain the total maximum slice similarity as final rank score (RS), i.e. for $k=1,...,M$ we compute:


\begin{equation}\label{eq:RS-score}
    \mathrm{RS}(V_k)
    = \sum_{i=1}^n \max_{j=1,...,m_k} \mathrm{SIM}(M_{V_Q}, M_{V_k})_{i,j}
\end{equation}

where ${V_k}_j$ denotes the $j$-th slice of volume $V_k$ and $m_k$ is the total number of slices of ${V_k}$. The retrieved volumes are re-ranked according to their RS, i.e. the volume with the highest score is considered as the most relevant search result.

\subsubsection{Localization with Re-ranking}
In \cref{sec:localization} we proposed a measure for assessing the capabilities of the search system to localize anatomical regions of interest. 
This localization measure relied on the hit slices that were finally retrieved by count-based aggregation. 
As described above, re-ranking may select a volume with a lower total hit count as the final search result based on a more fine-grained similarity measure. 
We propose to utilize this similarity information to identify slices of maximal similarity with the query data in order to localize the anatomical region of interest.

In detail, after having obtained the most relevant search result via re-ranking, we need to identify the most relevant slices that could contain the anatomical region of interest $r$. To this end, we consider the slices that highly contribute to the final RS score \eqref{eq:RS-score}. 
Since the full similarity matrices \eqref{eq:similarity_matrix} for all candidates from the first search stage are available, we can utilize these to identify the locations of maximum similarity between the query and candidate slices. 

Assume that the volume $V_{k^\star}$ obtained the highest RS score in the re-ranking process and therefore was returned as the final result after re-ranking, i.e. $k^* = argmax_k(RS(V_k))$. 
Further assume that $V_{k^\star}$ consists of $m_{k^\star}$ slices. First, we compute the vector $m_\mathrm{SIM}$ of size $m_{k^\star}$ containing for each slice of $V_{k^\star}$ the maximum similarity with the query slices:
%

\begin{equation}\label{eq:msim}
    m_\mathrm{SIM}(V_{k^\star})
    = \left[\max_{i=1,...,n} \mathrm{SIM}(M_{V_Q}, M_{V_{k^\star}})_{i,j} \right]_{j=1,...,m_{k^\star}}
\end{equation}

Hence, to determine the $L\leq m_{k^\star}$ slices of $V_{k^\star}$ with the maximum similarity, we determine the indices of the $L$ largest values of the vector $m_\mathrm{SIM}$. 
These slices are considered candidate locations for the anatomical region $r$. In our experiments, we chose $L=15$. 
The localization-ratio \eqref{eq:normalized-location-ratio} can be adopted in the case of re-ranking by determining how many of the $L$ most similar slices actually contained the anatomical region of interest. 

\begin{equation} \label{eq:normalized-location-ratio-rerank-version}
   \mathrm{LR} = \frac{
    \vert\text{slices among the $L$ most similar slices of $V_{k^\star}$ that contain r}\vert
   }{L}
\end{equation}

\section{Evaluation}

In this section, we evaluate the retrieval recall of the methods explained in \Cref{sec: search and retrieval} and \Cref{sec: re:ranking}.
The results related to the 29 coarse anatomical structures from \Cref{tab: mapping table} and the results related to the original 104 fine-grained anatomical structures from\cite{wasserthal2023totalsegmentator} are presented separately in the following. 
In the tables presented in this section, the average and standard deviation (STD) columns allow identifying difficult classes across models (low average) and the ones that have higher variations among models (higher STD). The average and STD rows show the average and STD over all the classes for each model.

\subsection{Search and Retrieval}
\label{sec:ablation search and retrieval}
\subsubsection{Slice-wise }

Detailed computation of the recall measure for different retrieval methods is explained in \Cref{sec: search and retrieval}.
\Cref{tab:slicewise-recall-28-regions} and \Cref{tab:slicewise-recall-104-regions} show the retrieval recall of 29 coarse anatomical regions and 104 original TS anatomical regions, respectively, using the slice-wise method. The slice-wise recall is considered the lower bound recall because for a perfect recall all the anatomical regions present in the query slice should appear in the retrieved slice.

In slice-wise retrieval, DreamSim is the best-performing model with retrieval recall of $ .863 \pm .107$ and $ .797 \pm  .129$ for coarse and original TS classes, respectively. 
ResNet50 pre-trained on fractal images has the lowest retrieval recall almost on every anatomical region for 29 and 104 classes. 
This is however expected due to the nature of synthetic generated images. 

In \Cref{tab:slicewise-recall-104-regions} the gallbladder has the lowest retrieval rate followed by vertebrae C4 and C5 (see average column).
However, in \Cref{tab:slicewise-recall-28-regions} the vertebrae class shows a higher recall which indicated that the vertebrae classes were detected but the exact location, i.e. C4 or C5 were mismatched. 
The same pattern can be observed in rib classes. 

\begin{table}[!htpb]
  \centering
  \renewcommand*{\arraystretch}{ .50} 
  \caption{Slice-wise recall of coarse anatomical regions (29 classes) using HNSW Indexing. In each row, bold numbers represent the best-performing values, while italicized numbers indicate the worst-performing. The separate average and standard deviation (STD) columns are color-coded, with blue indicating the best-performing values and yellow indicating the worst-performing values across different models. Additionally, bold numbers in colored columns represent the best classes in terms of average and standard deviation, while italicized values represent the worst-performing class across the models.}
  \scriptsize
    \begin{tabular}{lcccccc|cc}
    Model & DINOv1 & DINOv2 & DreamSim & SwinTrans. & \multicolumn{2}{c|}{ResNet50} & & \\ \cline{1-7}
      Dataset (pre-trained) & (ImgNet) & (ImgNet) & (ImgNet)& (RadImg)& 
    (Fractaldb) & (RadImg) &Average & STD \\ \hline
    
    adrenal gland & \textbf{.749} & .639 & .671 & .614 & \textit{.490} & .557 & \cellcolor[rgb]{ .871,  .875,  .847}.620 & \cellcolor[rgb]{ .961,  .925,  .816}.090 \\
    autochthon & \textbf{.980} & .974 & .979 & .976 & \textit{.941} & .965 & \cellcolor[rgb]{ .706,  .776,  .906}\textbf{.969} & \cellcolor[rgb]{ .714,  .78,  .902}.015 \\
    brain & .852 & \textit{.843} & \textbf{.901} & .894 & .850 & .863 & \cellcolor[rgb]{ .753,  .808,  .886}.867 & \cellcolor[rgb]{ .749,  .8,  .89}.025 \\
    cardiovascular system & .978 & .974 & \textbf{.979} & .970 & \textit{.941} & .953 & \cellcolor[rgb]{ .71,  .78,  .902}.966 & \cellcolor[rgb]{ .718,  .784,  .902}.015 \\
    clavicula & .886 & .884 & \textbf{.898} & .857 & \textit{.632} & .873 & \cellcolor[rgb]{ .769,  .816,  .882}.838 & \cellcolor[rgb]{ .996,  .945,  .804}.102 \\
    colon & .932 & .931 & \textbf{.945} & .912 & \textit{.830} & .905 & \cellcolor[rgb]{ .737,  .796,  .894}.909 & \cellcolor[rgb]{ .804,  .831,  .871}.042 \\
    duodenum & .678 & .682 & .719 & .697 & \textit{.605} & \textbf{.733} & \cellcolor[rgb]{ .839,  .855,  .855}.686 & \cellcolor[rgb]{ .812,  .839,  .867}.045 \\
    esophagus & .934 & .934 & \textbf{.936} & .933 & \textit{.870} & .894 & \cellcolor[rgb]{ .733,  .792,  .894}.917 & \cellcolor[rgb]{ .757,  .808,  .886}.028 \\
    face  & .854 & .840 & \textbf{.872} & .788 & \textit{.692} & .733 & \cellcolor[rgb]{ .788,  .824,  .875}.797 & \cellcolor[rgb]{ .898,  .89,  .839}.072 \\
    femur & .927 & .907 & \textbf{.953} & .914 & \textit{.778} & .860 & \cellcolor[rgb]{ .745,  .8,  .89}.890 & \cellcolor[rgb]{ .871,  .871,  .847}.063 \\
    gallbladder & \textit{.246} & .345 & .312 & .341 & .347 & \textbf{.400} & \cellcolor[rgb]{ 1,  .949,  .8}\textit{.332} & \cellcolor[rgb]{ .831,  .851,  .863}.051 \\
    gluteus muscles & .964 & .940 & \textbf{.978} & .950 & \textit{.879} & .915 & \cellcolor[rgb]{ .722,  .788,  .898}.938 & \cellcolor[rgb]{ .784,  .82,  .878}.036 \\
    hip   & .959 & .928 & \textbf{.974} & .941 & \textit{.880} & .907 & \cellcolor[rgb]{ .725,  .788,  .898}.931 & \cellcolor[rgb]{ .78,  .82,  .882}.034 \\
    humerus & .575 & .600 & \textbf{.633} & .598 & \textit{.351} & .523 & \cellcolor[rgb]{ .902,  .894,  .835}.547 & \cellcolor[rgb]{ 1,  .949,  .8}\textit{.102} \\
    iliopsoas & .950 & .933 & \textbf{.957} & .934 & \textit{.863} & .923 & \cellcolor[rgb]{ .725,  .788,  .898}.927 & \cellcolor[rgb]{ .776,  .816,  .882}.034 \\
    kidney & .759 & .771 & \textbf{.791} & .776 & \textit{.641} & .776 & \cellcolor[rgb]{ .808,  .835,  .867}.752 & \cellcolor[rgb]{ .847,  .859,  .855}.055 \\
    liver & .840 & .817 & \textbf{.844} & .841 & \textit{.814} & .839 & \cellcolor[rgb]{ .773,  .816,  .882}.833 & \cellcolor[rgb]{ .71,  .78,  .906}.013 \\
    lung  & .953 & .930 & \textbf{.958} & .940 & \textit{.890} & .898 & \cellcolor[rgb]{ .725,  .788,  .898}.928 & \cellcolor[rgb]{ .761,  .808,  .886}.028 \\
    pancreas & .720 & .685 & \textbf{.779} & .734 & \textit{.552} & .722 & \cellcolor[rgb]{ .831,  .851,  .859}.699 & \cellcolor[rgb]{ .918,  .902,  .831}.078 \\
    portal and splenic vein & \textbf{.731} & .627 & .679 & .658 & \textit{.522} & .584 & \cellcolor[rgb]{ .863,  .871,  .847}.634 & \cellcolor[rgb]{ .906,  .894,  .835}.074 \\
    rib   & .950 & .942 & \textbf{.951} & .948 & \textit{.900} & .933 & \cellcolor[rgb]{ .722,  .788,  .898}.937 & \cellcolor[rgb]{ .733,  .792,  .898}.020 \\
    sacrum & .894 & .865 & \textbf{.907} & .878 & \textit{.805} & .856 & \cellcolor[rgb]{ .753,  .808,  .886}.867 & \cellcolor[rgb]{ .784,  .82,  .878}.036 \\
    scapula & \textbf{.935} & .913 & .924 & .891 & \textit{.793} & .869 & \cellcolor[rgb]{ .745,  .8,  .89}.887 & \cellcolor[rgb]{ .835,  .851,  .859}.052 \\
    small bowel & .896 & .872 & \textbf{.900} & .894 & \textit{.783} & .892 & \cellcolor[rgb]{ .753,  .804,  .886}.873 & \cellcolor[rgb]{ .816,  .839,  .867}.045 \\
    spleen & \textbf{.774} & .719 & .735 & .699 & .731 & \textit{.693} & \cellcolor[rgb]{ .82,  .843,  .863}.725 & \cellcolor[rgb]{ .761,  .808,  .886}.029 \\
    stomach & .811 & .781 & \textbf{.844} & .778 & \textit{.741} & .752 & \cellcolor[rgb]{ .792,  .827,  .875}.784 & \cellcolor[rgb]{ .792,  .824,  .878}.038 \\
    trachea & .893 & .862 & \textbf{.903} & .863 & \textit{.762} & .816 & \cellcolor[rgb]{ .765,  .812,  .882}.850 & \cellcolor[rgb]{ .839,  .855,  .859}.053 \\
    urinary bladder & .720 & .643 & \textbf{.722} & .720 & \textit{.633} & .666 & \cellcolor[rgb]{ .839,  .855,  .855}.684 & \cellcolor[rgb]{ .8,  .831,  .875}.041 \\
    vertebrae & \textbf{.981} & .967 & .977 & .969 & \textit{.950} & .964 & \cellcolor[rgb]{ .71,  .78,  .902}.968 & \cellcolor[rgb]{ .706,  .776,  .906}\textbf{.011} \\ \hline
    Average  & .855 & .832 & \textbf{.863} & .837 & \textit{.751} & .813 &       &  \\
    STD   & .108 & .118 & \textbf{.107} & .114 & \textit{.152} & .124 &       &  \\
    \label{tab:slicewise-recall-28-regions}
    \end{tabular}%
\end{table}%

\FloatBarrier

\begingroup
\renewcommand*{\arraystretch}{ .50}
\scriptsize
\begin{longtable}{lcccccc|cc}
\caption{Slice-wise recall of all TS anatomical regions (104 classes) using HNSW Indexing. In each row, bold numbers represent the best-performing values, while italicized numbers indicate the worst-performing. The separate average and standard deviation (STD) columns are color-coded, with blue indicating the best-performing values and yellow indicating the worst-performing values across different models. Additionally, bold numbers in colored columns represent the best classes in terms of average and standard deviation, while italicized values represent the worst-performing class across the models.}  \\
Model & DINOv1 & DINOv2 & DreamSim & SwinTrans. & \multicolumn{2}{c|}{ResNet50} & & \\ \cline{1-7}
      Dataset (pre-trained) & (ImgNet) & (ImgNet) & (ImgNet)& (RadImg)& 
    (Fractaldb) & (RadImg) &Average & STD \\ \hline
    adrenal gland left & \textbf{.636} & .524 & .573 & .539 & \textit{.407} & .453 & \cellcolor[rgb]{ .914,  .898,  .831}.522 & \cellcolor[rgb]{ .847,  .859,  .855}.082 \\
    adrenal gland right & \textbf{.644} & .515 & .593 & .551 & \textit{.408} & .521 & \cellcolor[rgb]{ .906,  .894,  .831}.539 & \cellcolor[rgb]{ .843,  .855,  .859}.080 \\
    aorta & \textbf{.954} & .941 & .946 & .952 & \textit{.915} & .926 & \cellcolor[rgb]{ .722,  .788,  .898}.939 & \cellcolor[rgb]{ .71,  .776,  .906}.015 \\
    autochthon left & \textbf{.981} & .972 & .980 & .974 & \textit{.942} & .966 & \cellcolor[rgb]{ .71,  .78,  .902}.969 & \cellcolor[rgb]{ .706,  .776,  .906}.014 \\
    autochthon right & \textbf{.980} & .974 & .979 & .976 & \textit{.942} & .965 & \cellcolor[rgb]{ .706,  .776,  .906}\textbf{.969} & \cellcolor[rgb]{ .706,  .776,  .906}.014 \\
    brain & .852 & \textit{.843} & \textbf{.901} & .894 & .850 & .863 & \cellcolor[rgb]{ .757,  .808,  .886}.867 & \cellcolor[rgb]{ .729,  .788,  .898}.025 \\
    clavicula left & .866 & .875 & \textbf{.886} & .864 & \textit{.636} & .874 & \cellcolor[rgb]{ .769,  .816,  .882}.833 & \cellcolor[rgb]{ .878,  .878,  .847}.097 \\
    clavicula right & .862 & \textbf{.871} & .867 & .840 & \textit{.614} & .855 & \cellcolor[rgb]{ .776,  .82,  .878}.818 & \cellcolor[rgb]{ .886,  .882,  .843}.101 \\
    colon & .932 & .931 & \textbf{.945} & .912 & \textit{.830} & .905 & \cellcolor[rgb]{ .737,  .796,  .894}.909 & \cellcolor[rgb]{ .761,  .808,  .886}.042 \\
    duodenum & .678 & .682 & .719 & .697 & \textit{.605} & \textbf{.733} & \cellcolor[rgb]{ .839,  .855,  .855}.686 & \cellcolor[rgb]{ .769,  .812,  .882}.045 \\
    esophagus & .934 & .934 & \textbf{.936} & .933 & \textit{.870} & .894 & \cellcolor[rgb]{ .733,  .792,  .894}.917 & \cellcolor[rgb]{ .733,  .792,  .898}.028 \\
    face  & .854 & .840 & \textbf{.872} & .788 & \textit{.692} & .733 & \cellcolor[rgb]{ .788,  .824,  .875}.797 & \cellcolor[rgb]{ .827,  .847,  .863}.072 \\
    femur left & .920 & .902 & \textbf{.940} & .909 & \textit{.773} & .855 & \cellcolor[rgb]{ .749,  .8,  .89}.883 & \cellcolor[rgb]{ .804,  .831,  .871}.061 \\
    femur right & .931 & .910 & \textbf{.952} & .938 & \textit{.808} & .915 & \cellcolor[rgb]{ .737,  .796,  .894}.909 & \cellcolor[rgb]{ .784,  .82,  .878}.052 \\
    gallbladder & \textit{.246} & .345 & .312 & .341 & .347 & \textbf{.400} & \cellcolor[rgb]{ 1,  .949,  .8}\textit{.332} & \cellcolor[rgb]{ .78,  .82,  .878}.051 \\
    gluteus maximus left & .937 & .914 & \textbf{.951} & .927 & \textit{.845} & .903 & \cellcolor[rgb]{ .733,  .792,  .894}.913 & \cellcolor[rgb]{ .753,  .804,  .89}.037 \\
    gluteus maximus right & .942 & .914 & \textbf{.945} & .925 & \textit{.858} & .900 & \cellcolor[rgb]{ .733,  .792,  .894}.914 & \cellcolor[rgb]{ .745,  .796,  .894}.032 \\
    gluteus medius left & .930 & .878 & \textbf{.948} & .920 & \textit{.824} & .883 & \cellcolor[rgb]{ .741,  .796,  .89}.897 & \cellcolor[rgb]{ .769,  .812,  .882}.045 \\
    gluteus medius right & .922 & .892 & \textbf{.951} & .923 & \textit{.852} & .893 & \cellcolor[rgb]{ .737,  .796,  .894}.905 & \cellcolor[rgb]{ .749,  .8,  .89}.034 \\
    gluteus minimus left & .872 & .824 & \textbf{.894} & .855 & \textit{.795} & .876 & \cellcolor[rgb]{ .761,  .812,  .886}.853 & \cellcolor[rgb]{ .753,  .804,  .89}.037 \\
    gluteus minimus right & .876 & \textit{.811} & .878 & .874 & .819 & \textbf{.898} & \cellcolor[rgb]{ .757,  .808,  .886}.860 & \cellcolor[rgb]{ .749,  .8,  .89}.035 \\
    heart atrium left & .709 & .656 & \textbf{.800} & .680 & .588 & \textit{.542} & \cellcolor[rgb]{ .851,  .863,  .855}.663 & \cellcolor[rgb]{ .867,  .871,  .851}.091 \\
    heart atrium right & .793 & .762 & \textbf{.870} & .773 & .684 & \textit{.668} & \cellcolor[rgb]{ .804,  .835,  .871}.758 & \cellcolor[rgb]{ .831,  .847,  .863}.074 \\
    heart myocardium & .798 & .757 & \textbf{.844} & .808 & \textit{.715} & .733 & \cellcolor[rgb]{ .796,  .831,  .871}.776 & \cellcolor[rgb]{ .776,  .82,  .882}.049 \\
    heart ventricle left & .778 & .724 & \textbf{.824} & .788 & \textit{.699} & .720 & \cellcolor[rgb]{ .808,  .835,  .867}.756 & \cellcolor[rgb]{ .776,  .816,  .882}.048 \\
    heart ventricle right & .802 & .801 & \textbf{.851} & .822 & \textit{.723} & .738 & \cellcolor[rgb]{ .792,  .827,  .875}.789 & \cellcolor[rgb]{ .78,  .82,  .882}.049 \\
    hip left & .959 & .928 & \textbf{.971} & .937 & \textit{.880} & .905 & \cellcolor[rgb]{ .725,  .788,  .898}.930 & \cellcolor[rgb]{ .745,  .8,  .894}.034 \\
    hip right & .963 & .932 & \textbf{.977} & .948 & \textit{.889} & .916 & \cellcolor[rgb]{ .722,  .788,  .898}.938 & \cellcolor[rgb]{ .741,  .796,  .894}.032 \\
    humerus left & .525 & .571 & \textbf{.591} & .577 & \textit{.313} & .471 & \cellcolor[rgb]{ .922,  .902,  .827}.508 & \cellcolor[rgb]{ .894,  .886,  .839}.105 \\
    humerus right & .593 & .625 & \textbf{.627} & .567 & \textit{.314} & .529 & \cellcolor[rgb]{ .906,  .894,  .831}.543 & \cellcolor[rgb]{ .922,  .902,  .831}.118 \\
    iliac artery left & .882 & .863 & \textbf{.902} & .893 & \textit{.813} & .841 & \cellcolor[rgb]{ .757,  .808,  .886}.866 & \cellcolor[rgb]{ .745,  .8,  .894}.034 \\
    iliac artery right & .905 & .869 & \textbf{.918} & .895 & \textit{.822} & .851 & \cellcolor[rgb]{ .749,  .804,  .89}.876 & \cellcolor[rgb]{ .753,  .804,  .89}.036 \\
    iliac vena left & .903 & .868 & \textbf{.908} & .893 & \textit{.825} & .857 & \cellcolor[rgb]{ .753,  .804,  .89}.876 & \cellcolor[rgb]{ .741,  .796,  .894}.032 \\
    iliac vena right & .910 & .870 & \textbf{.923} & .891 & \textit{.831} & .873 & \cellcolor[rgb]{ .749,  .8,  .89}.883 & \cellcolor[rgb]{ .745,  .8,  .894}.033 \\
    iliopsoas left & .950 & .929 & \textbf{.958} & .932 & \textit{.861} & .924 & \cellcolor[rgb]{ .729,  .792,  .898}.926 & \cellcolor[rgb]{ .749,  .8,  .894}.034 \\
    iliopsoas right & .947 & .929 & \textbf{.951} & .932 & \textit{.854} & .922 & \cellcolor[rgb]{ .729,  .792,  .898}.923 & \cellcolor[rgb]{ .749,  .8,  .89}.035 \\
    inferior vena cava & \textbf{.928} & .896 & .922 & .923 & \textit{.841} & .893 & \cellcolor[rgb]{ .741,  .796,  .894}.901 & \cellcolor[rgb]{ .745,  .8,  .894}.033 \\
    kidney left & .719 & .708 & \textbf{.762} & .747 & \textit{.600} & \textbf{.762} & \cellcolor[rgb]{ .824,  .847,  .863}.716 & \cellcolor[rgb]{ .804,  .831,  .871}.061 \\
    kidney right & .708 & .724 & .755 & .737 & \textit{.602} & \textbf{.756} & \cellcolor[rgb]{ .827,  .847,  .863}.714 & \cellcolor[rgb]{ .796,  .827,  .875}.058 \\
    liver & .840 & .817 & \textbf{.844} & .841 & \textit{.814} & .839 & \cellcolor[rgb]{ .773,  .816,  .882}.833 & \cellcolor[rgb]{ .706,  .776,  .906}\textbf{.013} \\
    lung lower lobe left & .903 & .885 & \textbf{.908} & .887 & .826 & \textit{.811} & \cellcolor[rgb]{ .753,  .804,  .886}.870 & \cellcolor[rgb]{ .761,  .808,  .886}.041 \\
    lung lower lobe right & .903 & .880 & \textbf{.914} & .897 & \textit{.806} & .809 & \cellcolor[rgb]{ .753,  .804,  .886}.868 & \cellcolor[rgb]{ .776,  .816,  .882}.048 \\
    lung middle lobe right & .800 & .785 & \textbf{.818} & .794 & .726 & \textit{.699} & \cellcolor[rgb]{ .8,  .831,  .871}.770 & \cellcolor[rgb]{ .773,  .816,  .882}.047 \\
    lung upper lobe left & .917 & .909 & \textbf{.921} & .906 & \textit{.850} & .875 & \cellcolor[rgb]{ .741,  .8,  .89}.896 & \cellcolor[rgb]{ .733,  .792,  .898}.028 \\
    lung upper lobe right & \textbf{.928} & .883 & .919 & .885 & \textit{.818} & .848 & \cellcolor[rgb]{ .749,  .804,  .89}.880 & \cellcolor[rgb]{ .765,  .808,  .886}.042 \\
    pancreas & .720 & .685 & \textbf{.779} & .734 & \textit{.552} & .722 & \cellcolor[rgb]{ .831,  .851,  .859}.699 & \cellcolor[rgb]{ .839,  .855,  .859}.078 \\
    portal and splenic vein & \textbf{.731} & .627 & .679 & .658 & \textit{.522} & .584 & \cellcolor[rgb]{ .863,  .871,  .847}.634 & \cellcolor[rgb]{ .827,  .847,  .863}.074 \\
    pulmonary artery & \textbf{.819} & .711 & .773 & .679 & \textit{.526} & .563 & \cellcolor[rgb]{ .843,  .859,  .855}.679 & \cellcolor[rgb]{ .914,  .898,  .831}.115 \\
    rib left 1 & .855 & .824 & \textbf{.867} & .851 & \textit{.669} & .821 & \cellcolor[rgb]{ .78,  .82,  .878}.815 & \cellcolor[rgb]{ .827,  .847,  .863}.073 \\
    rib left 10 & \textbf{.827} & .775 & .803 & .823 & \textit{.742} & .747 & \cellcolor[rgb]{ .792,  .827,  .875}.786 & \cellcolor[rgb]{ .753,  .804,  .89}.037 \\
    rib left 11 & .773 & .767 & .785 & \textbf{.788} & \textit{.694} & .756 & \cellcolor[rgb]{ .804,  .835,  .871}.761 & \cellcolor[rgb]{ .749,  .8,  .89}.034 \\
    rib left 12 & .594 & .568 & \textbf{.682} & .620 & \textit{.481} & .576 & \cellcolor[rgb]{ .886,  .882,  .839}.587 & \cellcolor[rgb]{ .812,  .839,  .867}.066 \\
    rib left 2 & .841 & .804 & \textbf{.858} & .807 & \textit{.681} & .803 & \cellcolor[rgb]{ .788,  .824,  .875}.799 & \cellcolor[rgb]{ .804,  .835,  .871}.062 \\
    rib left 3 & \textbf{.832} & .803 & .808 & .805 & \textit{.728} & .789 & \cellcolor[rgb]{ .788,  .827,  .875}.794 & \cellcolor[rgb]{ .749,  .8,  .89}.035 \\
    rib left 4 & \textbf{.820} & .783 & .809 & .776 & \textit{.738} & .759 & \cellcolor[rgb]{ .796,  .831,  .875}.781 & \cellcolor[rgb]{ .741,  .796,  .894}.031 \\
    rib left 5 & .789 & .786 & \textbf{.805} & .784 & \textit{.699} & .723 & \cellcolor[rgb]{ .804,  .835,  .871}.764 & \cellcolor[rgb]{ .765,  .812,  .886}.043 \\
    rib left 6 & \textbf{.815} & .787 & .797 & .787 & \textit{.706} & .751 & \cellcolor[rgb]{ .796,  .831,  .871}.774 & \cellcolor[rgb]{ .757,  .808,  .89}.039 \\
    rib left 7 & .830 & .825 & \textbf{.834} & .829 & \textit{.734} & .778 & \cellcolor[rgb]{ .784,  .824,  .878}.805 & \cellcolor[rgb]{ .761,  .808,  .886}.040 \\
    rib left 8 & .810 & .799 & \textbf{.850} & .831 & \textit{.745} & .777 & \cellcolor[rgb]{ .784,  .824,  .875}.802 & \cellcolor[rgb]{ .753,  .804,  .89}.038 \\
    rib left 9 & .826 & .803 & .833 & \textbf{.853} & \textit{.737} & .780 & \cellcolor[rgb]{ .784,  .824,  .878}.805 & \cellcolor[rgb]{ .765,  .808,  .886}.042 \\
    rib right 1 & \textbf{.852} & .820 & .828 & .831 & \textit{.672} & .827 & \cellcolor[rgb]{ .784,  .824,  .878}.805 & \cellcolor[rgb]{ .812,  .839,  .867}.066 \\
    rib right 10 & \textbf{.827} & .768 & .804 & .814 & \textit{.728} & .747 & \cellcolor[rgb]{ .796,  .827,  .875}.781 & \cellcolor[rgb]{ .757,  .808,  .886}.040 \\
    rib right 11 & .770 & .763 & \textbf{.798} & .771 & \textit{.681} & .742 & \cellcolor[rgb]{ .808,  .835,  .867}.754 & \cellcolor[rgb]{ .761,  .808,  .886}.040 \\
    rib right 12 & .577 & .570 & .619 & \textbf{.634} & \textit{.456} & .556 & \cellcolor[rgb]{ .894,  .886,  .839}.569 & \cellcolor[rgb]{ .808,  .835,  .871}.063 \\
    rib right 2 & .839 & .820 & \textbf{.840} & .815 & \textit{.680} & .802 & \cellcolor[rgb]{ .784,  .824,  .875}.799 & \cellcolor[rgb]{ .8,  .831,  .875}.060 \\
    rib right 3 & \textbf{.850} & .794 & .826 & .795 & \textit{.725} & .780 & \cellcolor[rgb]{ .788,  .827,  .875}.795 & \cellcolor[rgb]{ .765,  .812,  .886}.043 \\
    rib right 4 & \textbf{.834} & .790 & .809 & .770 & \textit{.738} & .753 & \cellcolor[rgb]{ .796,  .827,  .875}.782 & \cellcolor[rgb]{ .749,  .804,  .89}.036 \\
    rib right 5 & .802 & .791 & \textbf{.810} & .776 & \textit{.709} & .718 & \cellcolor[rgb]{ .8,  .831,  .871}.768 & \cellcolor[rgb]{ .769,  .812,  .886}.044 \\
    rib right 6 & \textbf{.810} & .788 & .772 & .779 & \textit{.709} & .741 & \cellcolor[rgb]{ .8,  .835,  .871}.766 & \cellcolor[rgb]{ .753,  .804,  .89}.036 \\
    rib right 7 & .803 & .813 & .805 & \textbf{.817} & \textit{.731} & .765 & \cellcolor[rgb]{ .792,  .827,  .875}.789 & \cellcolor[rgb]{ .745,  .8,  .894}.034 \\
    rib right 8 & .814 & .792 & \textbf{.847} & .833 & \textit{.754} & .778 & \cellcolor[rgb]{ .784,  .824,  .875}.803 & \cellcolor[rgb]{ .749,  .8,  .89}.035 \\
    rib right 9 & .823 & .793 & .813 & \textbf{.844} & \textit{.738} & .776 & \cellcolor[rgb]{ .788,  .824,  .875}.798 & \cellcolor[rgb]{ .753,  .804,  .89}.038 \\
    sacrum & .894 & .865 & \textbf{.907} & .878 & \textit{.805} & .856 & \cellcolor[rgb]{ .753,  .808,  .886}.867 & \cellcolor[rgb]{ .753,  .804,  .89}.036 \\
    scapula left & \textbf{.922} & .891 & .908 & .891 & \textit{.798} & .884 & \cellcolor[rgb]{ .749,  .804,  .89}.882 & \cellcolor[rgb]{ .769,  .812,  .886}.044 \\
    scapula right & \textbf{.930} & .905 & .919 & .884 & \textit{.799} & .872 & \cellcolor[rgb]{ .745,  .8,  .89}.885 & \cellcolor[rgb]{ .773,  .816,  .882}.047 \\
    small bowel & .896 & .872 & \textbf{.900} & .894 & \textit{.783} & .892 & \cellcolor[rgb]{ .753,  .804,  .886}.873 & \cellcolor[rgb]{ .769,  .812,  .882}.045 \\
    spleen & \textbf{.774} & .719 & .735 & .699 & .731 & \textit{.693} & \cellcolor[rgb]{ .82,  .843,  .863}.725 & \cellcolor[rgb]{ .737,  .792,  .894}.029 \\
    stomach & .811 & .781 & \textbf{.844} & .778 & \textit{.741} & .752 & \cellcolor[rgb]{ .792,  .827,  .875}.784 & \cellcolor[rgb]{ .757,  .804,  .89}.038 \\
    trachea & .893 & .862 & \textbf{.903} & .863 & \textit{.762} & .816 & \cellcolor[rgb]{ .765,  .812,  .882}.850 & \cellcolor[rgb]{ .784,  .824,  .878}.053 \\
    urinary bladder & .720 & .643 & \textbf{.722} & .720 & \textit{.633} & .666 & \cellcolor[rgb]{ .839,  .855,  .855}.684 & \cellcolor[rgb]{ .761,  .808,  .886}.041 \\
    vertebrae C1 & .555 & .571 & \textbf{.655} & .592 & \textit{.399} & .592 & \cellcolor[rgb]{ .898,  .89,  .835}.561 & \cellcolor[rgb]{ .855,  .863,  .855}.086 \\
    vertebrae C2 & .744 & .613 & \textbf{.812} & .594 & \textit{.529} & .643 & \cellcolor[rgb]{ .851,  .863,  .851}.656 & \cellcolor[rgb]{ .894,  .886,  .839}.104 \\
    vertebrae C3 & \textbf{.677} & .566 & .586 & .414 & \textit{.271} & .359 & \cellcolor[rgb]{ .933,  .91,  .824}.479 & \cellcolor[rgb]{ 1,  .949,  .8}\textit{.155} \\
    vertebrae C4 & .427 & .377 & \textbf{.519} & .488 & .323 & \textit{.308} & \cellcolor[rgb]{ .969,  .929,  .812}.407 & \cellcolor[rgb]{ .855,  .863,  .855}.087 \\
    vertebrae C5 & .513 & .444 & \textbf{.572} & .565 & \textit{.330} & .366 & \cellcolor[rgb]{ .941,  .914,  .82}.465 & \cellcolor[rgb]{ .89,  .882,  .843}.102 \\
    vertebrae C6 & \textbf{.562} & \textbf{.562} & .536 & .423 & \textit{.220} & .445 & \cellcolor[rgb]{ .945,  .918,  .82}.458 & \cellcolor[rgb]{ .949,  .918,  .82}.131 \\
    vertebrae C7 & \textbf{.712} & .645 & .685 & .576 & \textit{.375} & .580 & \cellcolor[rgb]{ .878,  .878,  .843}.595 & \cellcolor[rgb]{ .925,  .906,  .827}.121 \\
    vertebrae L1 & .620 & .561 & \textbf{.662} & .653 & \textit{.452} & .540 & \cellcolor[rgb]{ .886,  .882,  .839}.582 & \cellcolor[rgb]{ .843,  .855,  .859}.080 \\
    vertebrae L2 & .555 & .514 & \textbf{.658} & .587 & \textit{.411} & .591 & \cellcolor[rgb]{ .902,  .89,  .835}.553 & \cellcolor[rgb]{ .851,  .863,  .855}.084 \\
    vertebrae L3 & \textbf{.747} & .533 & .608 & .593 & \textit{.503} & .629 & \cellcolor[rgb]{ .878,  .878,  .843}.602 & \cellcolor[rgb]{ .855,  .863,  .855}.086 \\
    vertebrae L4 & .612 & \textit{.449} & .639 & \textbf{.693} & .523 & .572 & \cellcolor[rgb]{ .886,  .882,  .839}.581 & \cellcolor[rgb]{ .855,  .863,  .851}.087 \\
    vertebrae L5 & .732 & \textit{.592} & \textbf{.748} & .714 & .606 & .631 & \cellcolor[rgb]{ .847,  .859,  .855}.670 & \cellcolor[rgb]{ .82,  .843,  .867}.069 \\
    vertebrae T1 & \textbf{.723} & .698 & .704 & .689 & \textit{.464} & .694 & \cellcolor[rgb]{ .851,  .863,  .851}.662 & \cellcolor[rgb]{ .878,  .878,  .843}.098 \\
    vertebrae T10 & .568 & .517 & \textbf{.583} & .565 & \textit{.449} & .518 & \cellcolor[rgb]{ .91,  .898,  .831}.533 & \cellcolor[rgb]{ .78,  .82,  .882}.050 \\
    vertebrae T11 & .555 & .539 & \textbf{.574} & .546 & \textit{.433} & .506 & \cellcolor[rgb]{ .914,  .898,  .831}.526 & \cellcolor[rgb]{ .78,  .82,  .878}.051 \\
    vertebrae T12 & .556 & .554 & \textbf{.604} & .581 & \textit{.468} & .562 & \cellcolor[rgb]{ .898,  .89,  .835}.554 & \cellcolor[rgb]{ .773,  .816,  .882}.047 \\
    vertebrae T2 & .750 & .670 & \textbf{.752} & .726 & \textit{.563} & .679 & \cellcolor[rgb]{ .835,  .855,  .859}.690 & \cellcolor[rgb]{ .824,  .843,  .863}.071 \\
    vertebrae T3 & .794 & .744 & \textbf{.814} & .736 & \textit{.621} & .668 & \cellcolor[rgb]{ .82,  .843,  .863}.729 & \cellcolor[rgb]{ .827,  .847,  .863}.073 \\
    vertebrae T4 & \textbf{.742} & .715 & .713 & .666 & \textit{.540} & .666 & \cellcolor[rgb]{ .843,  .859,  .855}.674 & \cellcolor[rgb]{ .827,  .847,  .863}.072 \\
    vertebrae T5 & .647 & .618 & \textbf{.701} & .627 & \textit{.513} & .550 & \cellcolor[rgb]{ .875,  .875,  .843}.609 & \cellcolor[rgb]{ .816,  .839,  .867}.068 \\
    vertebrae T6 & \textbf{.696} & .627 & .637 & .597 & \textit{.488} & .514 & \cellcolor[rgb]{ .882,  .878,  .843}.593 & \cellcolor[rgb]{ .839,  .855,  .859}.079 \\
    vertebrae T7 & .703 & .680 & \textbf{.705} & .613 & \textit{.460} & .516 & \cellcolor[rgb]{ .871,  .875,  .843}.613 & \cellcolor[rgb]{ .89,  .886,  .839}.104 \\
    vertebrae T8 & .595 & .590 & \textbf{.728} & .564 & \textit{.450} & .469 & \cellcolor[rgb]{ .894,  .886,  .835}.566 & \cellcolor[rgb]{ .886,  .882,  .843}.101 \\
    vertebrae T9 & .603 & .540 & \textbf{.660} & .609 & \textit{.515} & .524 & \cellcolor[rgb]{ .89,  .886,  .839}.575 & \cellcolor[rgb]{ .796,  .827,  .875}.057 \\ \hline
    Average & .784 & .750 & \textbf{.797} & .765 & \textit{.659} & .726 &       &  \\
    STD   & .137 & .144 & \textbf{.129} & .140 & \textit{.172} & .154 &       &  \\

    \label{tab:slicewise-recall-104-regions}
    \end{longtable}%
\endgroup

\subsubsection{Volume-based}
\label{sec:volume-based-eval}
This section presents the recall of volume-based retrieval explained in \Cref{Volume-based retrieval}
An overview of the evaluation is shown in \Cref{fig:volume-based}. 
In volume-based retrieval, per each query volume, one volume is retrieved. 
In the recall computation, the classes present in both the query and the retrieved volume are considered TP classes.  
The classes that are present in the query volume and are missing from the retrieved volume are considered FN.

\Cref{tab:volumewise-recall-29-regions} and \Cref{tab:volumewise-recall-104-regions} present the retrieval recall of the volume-based method on 29 and 104 classes, respectively. 
The overall recall rates are increased compared to slice-wise retrieval which is expected due to the aggregation and contextual effects of neighboring slices.

\Cref{tab:volumewise-recall-29-regions} shows that ResNet50 trained on RadImageNet outperforms other methods with an average recall of $ .952 \pm  .043$. 
However, in \Cref{tab:volumewise-recall-104-regions} DINOv1 outperforms all models including ResNet50 with an average recall of $ .923 \pm  .077$. 
This shows that the embeddings of finer classes are retrieved and assigned to a different similar class by ResNet50, thus, the performance from fine to coarse classes is improved. Whereas, all the self-supervised methods in \Cref{tab:volumewise-recall-104-regions} outperform the supervised methods. 
Although some models perform slightly better than others based on looking at isolated classes, overall models perform on par. 

\begin{table}[!htbp]
  \centering
  \renewcommand*{\arraystretch}{ .50} 
  \scriptsize
  \caption{Volume-based retrieval recall of coarse anatomical regions (29 classes) using HNSW Indexing. In each row, bold numbers represent the best-performing values, while italicized numbers indicate the worst-performing. The separate average and standard deviation (STD) columns are color-coded, with blue indicating the best-performing values and yellow indicating the worst-performing values across different models. Additionally, bold numbers in colored columns represent the best classes in terms of average and standard deviation, while italicized values represent the worst-performing class across the models.}
    \begin{tabular}{lrrrrrr|rr}
    Model & DINOv1 & DINOv2 & DreamSim & SwinTrans. & \multicolumn{2}{c|}{ResNet50} & & \\ \cline{1-7}
      Dataset (pre-trained) & (ImgNet) & (ImgNet) & (ImgNet)& (RadImg)& 
    (Fractaldb) & (RadImg) &Average & STD \\ \hline
    adrenal gland & \textbf{1.000} & \textit{ .960} & \textit{ .960} & \textit{ .960} &  .980 &  .980 & \cellcolor[rgb]{ .729,  .792,  .898} .973 & \cellcolor[rgb]{ .733,  .792,  .898} .016 \\
    autochthon &  .985 & \textit{ .969} & \textit{ .969} &  .985 &  .985 & \textbf{1.000} & \cellcolor[rgb]{ .718,  .784,  .902} .982 & \cellcolor[rgb]{ .71,  .776,  .906} .012 \\
    brain & \textit{ .692} &  .769 &  .769 & \textbf{ .846} &  .769 & \textbf{ .846} & \cellcolor[rgb]{ .969,  .929,  .812} .782 & \cellcolor[rgb]{ .945,  .914,  .824} .058 \\
    cardiovascular system & \textbf{1.000} & \textbf{1.000} & \textit{ .968} & \textbf{1.000} & \textbf{1.000} & \textit{ .968} & \cellcolor[rgb]{ .71,  .78,  .902} .989 & \cellcolor[rgb]{ .733,  .792,  .898} .016 \\
    clavicula & \textbf{ .949} & \textbf{ .949} & \textbf{ .949} &  .897 & \textit{ .821} & \textbf{ .949} & \cellcolor[rgb]{ .796,  .831,  .871} .919 & \cellcolor[rgb]{ .914,  .898,  .831} .052 \\
    colon & \textbf{1.000} & \textit{ .943} &  .981 & \textit{ .943} & \textbf{1.000} &  .962 & \cellcolor[rgb]{ .729,  .792,  .894} .972 & \cellcolor[rgb]{ .784,  .82,  .878} .026 \\
    duodenum &  .940 & \textit{ .860} &  .900 &  .920 & \textbf{ .980} &  .920 & \cellcolor[rgb]{ .796,  .831,  .871} .920 & \cellcolor[rgb]{ .855,  .863,  .855} .040 \\
    esophagus &  .964 &  .964 & \textit{ .946} &  .982 & \textit{ .946} & \textbf{1.000} & \cellcolor[rgb]{ .737,  .796,  .894} .967 & \cellcolor[rgb]{ .757,  .804,  .89} .021 \\
    face  &  .765 &  .765 & \textit{ .706} & \textit{ .706} & \textit{ .706} & \textbf{ .882} & \cellcolor[rgb]{ 1,  .949,  .8}\textit{ .755} & \cellcolor[rgb]{ 1,  .949,  .8}\textit{ .069} \\
    femur &  .933 &  .933 & \textit{ .911} & \textit{ .911} & \textbf{ .956} &  .933 & \cellcolor[rgb]{ .784,  .824,  .878} .930 & \cellcolor[rgb]{ .737,  .792,  .898} .017 \\
    gallbladder &  .846 & \textit{ .795} &  .872 &  .821 &  .846 & \textbf{ .897} & \cellcolor[rgb]{ .886,  .882,  .839} .846 & \cellcolor[rgb]{ .835,  .851,  .863} .036 \\
    gluteus muscles & \textbf{1.000} &  .977 &  .977 &  .977 & \textbf{1.000} & \textit{ .955} & \cellcolor[rgb]{ .718,  .784,  .902} .981 & \cellcolor[rgb]{ .737,  .796,  .894} .017 \\
    hip   & \textbf{1.000} &  .977 &  .977 &  .977 & \textbf{1.000} & \textit{ .955} & \cellcolor[rgb]{ .718,  .784,  .902} .981 & \cellcolor[rgb]{ .737,  .796,  .894} .017 \\
    humerus &  .898 & \textit{ .857} & \textbf{ .980} &  .898 &  .878 &  .878 & \cellcolor[rgb]{ .824,  .847,  .863} .898 & \cellcolor[rgb]{ .867,  .871,  .847} .043 \\
    iliopsoas & \textbf{ .981} & \textbf{ .981} & \textit{ .962} & \textit{ .962} & \textbf{ .981} & \textit{ .962} & \cellcolor[rgb]{ .729,  .792,  .894} .972 & \cellcolor[rgb]{ .706,  .776,  .906}\textbf{ .010} \\
    kidney &  .945 &  .927 &  .945 & \textit{ .891} &  .927 & \textbf{ .964} & \cellcolor[rgb]{ .776,  .82,  .878} .933 & \cellcolor[rgb]{ .776,  .816,  .882} .025 \\
    liver &  .964 & \textit{ .945} & \textbf{ .982} & \textit{ .945} & \textbf{ .982} & \textbf{ .982} & \cellcolor[rgb]{ .737,  .796,  .894} .967 & \cellcolor[rgb]{ .741,  .796,  .894} .018 \\
    lung  & \textbf{ .983} & \textbf{ .983} & \textit{ .931} & \textbf{ .983} & \textbf{ .983} & \textbf{ .983} & \cellcolor[rgb]{ .725,  .788,  .898} .974 & \cellcolor[rgb]{ .757,  .808,  .89} .021 \\
    pancreas &  .940 & \textit{ .920} & \textit{ .920} &  .940 &  .960 & \textbf{ .980} & \cellcolor[rgb]{ .765,  .812,  .882} .943 & \cellcolor[rgb]{ .769,  .812,  .882} .023 \\
    portal and splenic vein & \textbf{ .980} &  .960 & \textit{ .940} & \textbf{ .980} &  .960 & \textbf{ .980} & \cellcolor[rgb]{ .737,  .796,  .894} .967 & \cellcolor[rgb]{ .733,  .792,  .898} .016 \\
    rib   &  .983 &  .983 & \textit{ .949} & \textbf{1.000} &  .966 & \textbf{1.000} & \cellcolor[rgb]{ .718,  .784,  .898} .980 & \cellcolor[rgb]{ .753,  .804,  .89} .020 \\
    sacrum &  .977 & \textit{ .955} &  .977 & \textit{ .955} & \textbf{1.000} & \textit{ .955} & \cellcolor[rgb]{ .733,  .792,  .894} .970 & \cellcolor[rgb]{ .745,  .8,  .894} .019 \\
    scapula & \textbf{ .909} & \textbf{ .909} & \textbf{ .909} & \textit{ .818} &  .886 &  .886 & \cellcolor[rgb]{ .835,  .855,  .859} .886 & \cellcolor[rgb]{ .827,  .847,  .863} .035 \\
    small bowel &  .958 & \textit{ .896} &  .917 &  .958 & \textbf{ .979} &  .938 & \cellcolor[rgb]{ .769,  .816,  .882} .941 & \cellcolor[rgb]{ .808,  .835,  .871} .031 \\
    spleen & \textbf{1.000} &  .980 & \textit{ .960} &  .980 &  .980 & \textbf{1.000} & \cellcolor[rgb]{ .714,  .784,  .902} .983 & \cellcolor[rgb]{ .729,  .788,  .898} .015 \\
    stomach & \textbf{1.000} &  .980 & \textit{ .961} &  .980 &  .980 & \textbf{1.000} & \cellcolor[rgb]{ .714,  .784,  .902} .984 & \cellcolor[rgb]{ .725,  .788,  .898} .015 \\
    trachea & \textbf{ .951} & \textbf{ .951} & \textbf{ .951} &  .878 & \textit{ .805} &  .902 & \cellcolor[rgb]{ .812,  .839,  .867} .907 & \cellcolor[rgb]{ .945,  .918,  .82} .059 \\
    urinary bladder & \textbf{1.000} &  .977 &  .977 & \textit{ .953} &  .977 & \textit{ .953} & \cellcolor[rgb]{ .729,  .792,  .898} .973 & \cellcolor[rgb]{ .741,  .796,  .894} .018 \\
    vertebrae &  .984 & \textit{ .969} & \textbf{1.000} &  .984 & \textbf{1.000} & \textbf{1.000} & \cellcolor[rgb]{ .706,  .776,  .906}\textbf{ .990} & \cellcolor[rgb]{ .718,  .78,  .902} .013 \\ \hline
    Average &  .949 &  .932 &  .936 & \textit{ .932} &  .939 & \textbf{ .952} &       &  \\
    STD   &  .072 &  .064 &  .063 &  .067 & \textit{ .078} & \textbf{ .043} &       &  \\

    \label{tab:volumewise-recall-29-regions}
    \end{tabular}%
\end{table}%

\FloatBarrier
\begingroup
\renewcommand*{\arraystretch}{ .50} 
\scriptsize
\begin{longtable}{lcccccc|cc}
\caption{Volume-based retrieval recall of all TS anatomical regions (104 classes) using HNSW Indexing. In each row, bold numbers represent the best-performing values, while italicized numbers indicate the worst-performing. The separate average and standard deviation (STD) columns are color-coded, with blue indicating the best-performing values and yellow indicating the worst-performing values across different models. Additionally, bold numbers in colored columns represent the best classes in terms of average and standard deviation, while italicized values represent the worst-performing class across the models.}   \\
Model & DINOv1 & DINOv2 & DreamSim & SwinTrans. & \multicolumn{2}{c|}{ResNet50} & & \\ \cline{1-7}
      Dataset (pre-trained) & (ImgNet) & (ImgNet) & (ImgNet)& (RadImg)& 
    (Fractaldb) & (RadImg) & Average & STD \\ \hline
    adrenal gland left & \textbf{ .960} & \textit{ .900} &  .920 & \textit{ .900} & \textit{ .900} & \textbf{ .960} & \cellcolor[rgb]{ .729,  .792,  .898} .923 & \cellcolor[rgb]{ .753,  .804,  .89} .029 \\
    adrenal gland right & \textbf{ .980} & \textit{ .900} &  .940 & \textit{ .900} & \textit{ .900} &  .960 & \cellcolor[rgb]{ .722,  .788,  .898} .930 & \cellcolor[rgb]{ .773,  .816,  .882} .035 \\
    aorta & \textbf{ .984} &  .934 &  .934 &  .934 & \textit{ .918} &  .934 & \cellcolor[rgb]{ .714,  .784,  .902} .940 & \cellcolor[rgb]{ .733,  .792,  .898} .022 \\
    autochthon left & \textbf{ .969} &  .923 &  .938 &  .923 & \textit{ .908} &  .954 & \cellcolor[rgb]{ .718,  .784,  .902} .936 & \cellcolor[rgb]{ .733,  .792,  .898} .023 \\
    autochthon right & \textbf{ .969} &  .923 &  .938 &  .923 & \textit{ .908} &  .954 & \cellcolor[rgb]{ .718,  .784,  .902} .936 & \cellcolor[rgb]{ .733,  .792,  .898} .023 \\
    brain & \textit{ .692} & \textit{ .692} & \textit{ .692} & \textit{ .692} & \textbf{ .769} & \textit{ .692} & \cellcolor[rgb]{ .918,  .902,  .827} .705 & \cellcolor[rgb]{ .761,  .808,  .886} .031 \\
    clavicula left & \textbf{ .949} &  .923 &  .897 & \textit{ .821} & \textit{ .821} &  .897 & \cellcolor[rgb]{ .761,  .812,  .882} .885 & \cellcolor[rgb]{ .824,  .847,  .863} .053 \\
    clavicula right & \textbf{ .974} &  .947 &  .921 & \textit{ .816} & \textit{ .816} &  .895 & \cellcolor[rgb]{ .753,  .804,  .886} .895 & \cellcolor[rgb]{ .863,  .871,  .851} .067 \\
    colon & \textbf{ .981} & \textit{ .906} &  .962 & \textit{ .906} & \textit{ .906} &  .925 & \cellcolor[rgb]{ .722,  .788,  .898} .931 & \cellcolor[rgb]{ .765,  .812,  .886} .033 \\
    duodenum & \textbf{ .920} & \textit{ .820} &  .880 &  .880 &  .880 &  .880 & \cellcolor[rgb]{ .769,  .816,  .882} .877 & \cellcolor[rgb]{ .761,  .808,  .886} .032 \\
    esophagus &  .946 &  .911 &  .911 &  .911 & \textit{ .875} & \textbf{ .964} & \cellcolor[rgb]{ .733,  .792,  .894} .920 & \cellcolor[rgb]{ .761,  .808,  .886} .031 \\
    face  & \textbf{ .765} &  .706 &  .647 & \textit{ .588} &  .706 &  .765 & \cellcolor[rgb]{ .925,  .906,  .824} .696 & \cellcolor[rgb]{ .871,  .871,  .847} .069 \\
    femur left & \textbf{ .911} & \textbf{ .911} &  .889 & \textit{ .867} & \textit{ .867} & \textbf{ .911} & \cellcolor[rgb]{ .757,  .808,  .886} .893 & \cellcolor[rgb]{ .733,  .792,  .898} .022 \\
    femur right & \textbf{ .927} & \textbf{ .927} & \textit{ .902} & \textit{ .902} & \textbf{ .927} & \textbf{ .927} & \cellcolor[rgb]{ .733,  .792,  .894} .919 & \cellcolor[rgb]{ .706,  .776,  .906}\textbf{ .013} \\
    gallbladder &  .846 & \textit{ .744} &  .846 &  .795 &  .795 & \textbf{ .872} & \cellcolor[rgb]{ .824,  .847,  .863} .816 & \cellcolor[rgb]{ .808,  .835,  .871} .047 \\
    gluteus maximus left & \textbf{ .977} &  .953 &  .953 & \textit{ .907} &  .930 &  .930 & \cellcolor[rgb]{ .714,  .78,  .902} .942 & \cellcolor[rgb]{ .737,  .796,  .894} .024 \\
    gluteus maximus right & \textbf{ .977} &  .953 &  .930 &  .930 & \textit{ .907} &  .930 & \cellcolor[rgb]{ .718,  .784,  .902} .938 & \cellcolor[rgb]{ .737,  .796,  .894} .024 \\
    gluteus medius left & \textbf{ .977} &  .932 &  .955 & \textit{ .909} & \textit{ .909} &  .932 & \cellcolor[rgb]{ .718,  .784,  .902} .936 & \cellcolor[rgb]{ .745,  .8,  .894} .027 \\
    gluteus medius right & \textbf{ .977} &  .930 &  .953 &  .930 & \textit{ .907} &  .953 & \cellcolor[rgb]{ .714,  .78,  .902} .942 & \cellcolor[rgb]{ .737,  .796,  .894} .024 \\
    gluteus minimus left & \textbf{ .977} &  .953 &  .953 & \textit{ .907} &  .930 &  .930 & \cellcolor[rgb]{ .714,  .78,  .902} .942 & \cellcolor[rgb]{ .737,  .796,  .894} .024 \\
    gluteus minimus right & \textbf{ .976} &  .952 &  .952 & \textit{ .905} &  .929 &  .952 & \cellcolor[rgb]{ .71,  .78,  .902} .944 & \cellcolor[rgb]{ .741,  .796,  .894} .025 \\
    heart atrium left &  .915 & \textit{ .830} &  .872 &  .936 & \textit{ .830} & \textbf{ .979} & \cellcolor[rgb]{ .753,  .808,  .886} .894 & \cellcolor[rgb]{ .847,  .859,  .859} .060 \\
    heart atrium right &  .939 &  .898 &  .898 &  .939 & \textit{ .816} & \textbf{ .980} & \cellcolor[rgb]{ .737,  .796,  .894} .912 & \cellcolor[rgb]{ .831,  .851,  .863} .056 \\
    heart myocardium &  .939 &  .898 &  .898 &  .939 & \textit{ .816} & \textbf{ .980} & \cellcolor[rgb]{ .737,  .796,  .894} .912 & \cellcolor[rgb]{ .831,  .851,  .863} .056 \\
    heart ventricle left &  .939 &  .898 &  .878 &  .939 & \textit{ .816} & \textbf{ .980} & \cellcolor[rgb]{ .741,  .8,  .89} .908 & \cellcolor[rgb]{ .835,  .851,  .859} .057 \\
    heart ventricle right &  .939 &  .898 &  .898 &  .939 & \textit{ .816} & \textbf{ .980} & \cellcolor[rgb]{ .737,  .796,  .894} .912 & \cellcolor[rgb]{ .831,  .851,  .863} .056 \\
    hip left & \textbf{ .977} &  .932 &  .955 &  .932 & \textit{ .909} &  .932 & \cellcolor[rgb]{ .714,  .784,  .902} .939 & \cellcolor[rgb]{ .737,  .792,  .898} .023 \\
    hip right & \textbf{ .977} &  .932 &  .955 &  .932 & \textit{ .886} &  .932 & \cellcolor[rgb]{ .718,  .784,  .902} .936 & \cellcolor[rgb]{ .757,  .804,  .89} .030 \\
    humerus left & \textbf{ .949} &  .897 & \textbf{ .949} &  .872 & \textit{ .846} &  .897 & \cellcolor[rgb]{ .749,  .8,  .89} .902 & \cellcolor[rgb]{ .788,  .824,  .878} .041 \\
    humerus right &  .875 &  .854 & \textbf{ .917} &  .833 &  .833 & \textit{ .813} & \cellcolor[rgb]{ .788,  .827,  .875} .854 & \cellcolor[rgb]{ .776,  .816,  .882} .037 \\
    iliac artery left & \textbf{ .977} &  .932 &  .955 &  .932 & \textit{ .909} &  .955 & \cellcolor[rgb]{ .714,  .78,  .902} .943 & \cellcolor[rgb]{ .737,  .792,  .894} .024 \\
    iliac artery right & \textbf{ .955} &  .909 &  .932 &  .909 & \textit{ .886} &  .932 & \cellcolor[rgb]{ .733,  .792,  .894} .920 & \cellcolor[rgb]{ .737,  .792,  .894} .024 \\
    iliac vena left & \textbf{ .977} &  .932 &  .955 &  .932 & \textit{ .909} &  .932 & \cellcolor[rgb]{ .714,  .784,  .902} .939 & \cellcolor[rgb]{ .737,  .792,  .898} .023 \\
    iliac vena right & \textbf{ .955} &  .909 &  .932 &  .909 & \textit{ .886} &  .932 & \cellcolor[rgb]{ .733,  .792,  .894} .920 & \cellcolor[rgb]{ .737,  .792,  .894} .024 \\
    iliopsoas left & \textbf{ .943} & \textit{ .887} &  .925 &  .925 &  .906 &  .925 & \cellcolor[rgb]{ .733,  .792,  .894} .918 & \cellcolor[rgb]{ .725,  .788,  .902} .019 \\
    iliopsoas right & \textbf{ .961} &  .922 &  .941 &  .922 & \textit{ .902} &  .941 & \cellcolor[rgb]{ .722,  .788,  .898} .931 & \cellcolor[rgb]{ .729,  .788,  .898} .021 \\
    inferior vena cava & \textbf{ .982} &  .930 &  .965 &  .930 & \textit{ .912} &  .965 & \cellcolor[rgb]{ .71,  .78,  .902} .947 & \cellcolor[rgb]{ .749,  .8,  .894} .027 \\
    kidney left &  .906 &  .887 &  .906 & \textit{ .849} &  .868 & \textbf{ .943} & \cellcolor[rgb]{ .757,  .808,  .886} .893 & \cellcolor[rgb]{ .765,  .812,  .886} .033 \\
    kidney right & \textbf{ .900} & \textit{ .820} &  .880 &  .860 & \textit{ .820} & \textbf{ .900} & \cellcolor[rgb]{ .78,  .82,  .878} .863 & \cellcolor[rgb]{ .776,  .816,  .882} .037 \\
    liver &  .945 &  .909 & \textbf{ .964} &  .909 & \textit{ .891} &  .945 & \cellcolor[rgb]{ .725,  .788,  .898} .927 & \cellcolor[rgb]{ .749,  .8,  .89} .028 \\
    lung lower lobe left & \textbf{ .912} &  .877 & \textit{ .842} &  .895 &  .895 & \textbf{ .912} & \cellcolor[rgb]{ .757,  .808,  .886} .889 & \cellcolor[rgb]{ .745,  .8,  .894} .026 \\
    lung lower lobe right & \textbf{ .946} &  .911 & \textit{ .875} &  .893 & \textit{ .875} &  .929 & \cellcolor[rgb]{ .745,  .8,  .89} .905 & \cellcolor[rgb]{ .753,  .804,  .89} .029 \\
    lung middle lobe right &  .939 &  .918 &  .878 &  .959 & \textit{ .837} & \textbf{ .980} & \cellcolor[rgb]{ .733,  .792,  .894} .918 & \cellcolor[rgb]{ .824,  .847,  .863} .053 \\
    lung upper lobe left &  .929 &  .911 &  .911 &  .911 & \textit{ .839} & \textbf{ .946} & \cellcolor[rgb]{ .741,  .8,  .89} .908 & \cellcolor[rgb]{ .776,  .816,  .882} .036 \\
    lung upper lobe right & \textbf{ .891} &  .870 &  .870 & \textit{ .804} & \textbf{ .891} &  .870 & \cellcolor[rgb]{ .78,  .82,  .878} .866 & \cellcolor[rgb]{ .761,  .808,  .886} .032 \\
    pancreas &  .920 & \textit{ .880} &  .900 &  .900 &  .900 & \textbf{ .960} & \cellcolor[rgb]{ .741,  .796,  .89} .910 & \cellcolor[rgb]{ .749,  .8,  .89} .028 \\
     portal and splenic vein & \textbf{ .960} &  .920 &  .920 &  .940 & \textit{ .880} & \textbf{ .960} & \cellcolor[rgb]{ .722,  .788,  .898} .930 & \cellcolor[rgb]{ .757,  .804,  .89} .030 \\
    pulmonary artery & \textbf{ .850} &  .825 & \textbf{ .850} & \textit{ .750} &  .800 &  .800 & \cellcolor[rgb]{ .824,  .847,  .863} .813 & \cellcolor[rgb]{ .78,  .82,  .882} .038 \\
    rib left 1 & \textbf{ .974} &  .947 &  .921 &  .842 & \textit{ .816} &  .895 & \cellcolor[rgb]{ .749,  .804,  .89} .899 & \cellcolor[rgb]{ .847,  .859,  .855} .061 \\
    rib left 10 &  .961 &  .922 &  .922 &  .941 & \textit{ .882} & \textbf{ .980} & \cellcolor[rgb]{ .718,  .784,  .898} .935 & \cellcolor[rgb]{ .769,  .812,  .886} .034 \\
    rib left 11 &  .961 &  .922 &  .922 &  .941 & \textit{ .882} & \textbf{ .980} & \cellcolor[rgb]{ .718,  .784,  .898} .935 & \cellcolor[rgb]{ .769,  .812,  .886} .034 \\
    rib left 12 &  .896 &  .938 &  .896 &  .917 & \textit{ .875} & \textbf{ .958} & \cellcolor[rgb]{ .737,  .796,  .894} .913 & \cellcolor[rgb]{ .757,  .808,  .89} .031 \\
    rib left 2 & \textbf{ .950} & \textbf{ .950} &  .925 & \textit{ .825} & \textit{ .825} &  .875 & \cellcolor[rgb]{ .757,  .808,  .886} .892 & \cellcolor[rgb]{ .839,  .855,  .859} .058 \\
    rib left 3 & \textbf{ .951} &  .927 &  .927 & \textit{ .829} &  .854 &  .878 & \cellcolor[rgb]{ .753,  .804,  .886} .894 & \cellcolor[rgb]{ .808,  .835,  .871} .048 \\
    rib left 4 & \textbf{ .900} &  .875 & \textbf{ .900} & \textit{ .825} &  .850 &  .875 & \cellcolor[rgb]{ .773,  .816,  .878} .871 & \cellcolor[rgb]{ .753,  .804,  .89} .029 \\
    rib left 5 & \textbf{ .909} & \textit{ .841} &  .864 &  .864 & \textit{ .841} & \textbf{ .909} & \cellcolor[rgb]{ .773,  .816,  .878} .871 & \cellcolor[rgb]{ .757,  .808,  .886} .031 \\
    rib left 6 &  .880 &  .840 &  .860 &  .940 & \textit{ .820} & \textbf{ .960} & \cellcolor[rgb]{ .765,  .812,  .882} .883 & \cellcolor[rgb]{ .831,  .851,  .863} .056 \\
    rib left 7 &  .959 &  .918 &  .918 &  .959 & \textit{ .857} & \textbf{ .980} & \cellcolor[rgb]{ .722,  .788,  .898} .932 & \cellcolor[rgb]{ .796,  .831,  .875} .044 \\
    rib left 8 &  .961 &  .922 &  .922 &  .961 & \textit{ .882} & \textbf{ .980} & \cellcolor[rgb]{ .718,  .784,  .902} .938 & \cellcolor[rgb]{ .773,  .816,  .882} .036 \\
    rib left 9 &  .961 &  .922 &  .922 &  .961 & \textit{ .902} & \textbf{ .980} & \cellcolor[rgb]{ .714,  .78,  .902} .941 & \cellcolor[rgb]{ .757,  .804,  .89} .030 \\
    rib right 1 & \textbf{ .974} &  .947 &  .921 & \textit{ .842} & \textit{ .842} &  .895 & \cellcolor[rgb]{ .745,  .8,  .89} .904 & \cellcolor[rgb]{ .827,  .847,  .863} .054 \\
    rib right 10 &  .961 &  .922 &  .922 &  .941 & \textit{ .882} & \textbf{ .980} & \cellcolor[rgb]{ .718,  .784,  .898} .935 & \cellcolor[rgb]{ .769,  .812,  .886} .034 \\
    rib right 11 &  .961 &  .922 &  .922 &  .941 & \textit{ .882} & \textbf{ .980} & \cellcolor[rgb]{ .718,  .784,  .898} .935 & \cellcolor[rgb]{ .769,  .812,  .886} .034 \\
    rib right 12 &  .872 &  .915 &  .872 &  .936 & \textit{ .830} & \textbf{ .957} & \cellcolor[rgb]{ .753,  .804,  .886} .897 & \cellcolor[rgb]{ .808,  .835,  .871} .047 \\
    rib right 2 & \textbf{ .974} &  .949 &  .923 & \textit{ .846} & \textit{ .846} &  .897 & \cellcolor[rgb]{ .745,  .8,  .89} .906 & \cellcolor[rgb]{ .824,  .843,  .863} .053 \\
    rib right 3 & \textbf{ .927} &  .902 &  .902 & \textit{ .805} &  .854 &  .854 & \cellcolor[rgb]{ .773,  .816,  .882} .874 & \cellcolor[rgb]{ .8,  .831,  .875} .045 \\
    rib right 4 & \textbf{ .927} &  .902 &  .878 & \textit{ .829} &  .854 &  .854 & \cellcolor[rgb]{ .773,  .816,  .882} .874 & \cellcolor[rgb]{ .773,  .816,  .882} .036 \\
    rib right 5 & \textbf{ .932} & \textit{ .841} &  .864 &  .864 &  .864 &  .886 & \cellcolor[rgb]{ .773,  .816,  .882} .875 & \cellcolor[rgb]{ .761,  .808,  .886} .031 \\
    rib right 6 & \textbf{ .918} &  .857 &  .857 & \textbf{ .918} & \textit{ .796} & \textbf{ .918} & \cellcolor[rgb]{ .769,  .816,  .882} .878 & \cellcolor[rgb]{ .816,  .839,  .867} .050 \\
    rib right 7 &  .959 &  .918 &  .918 &  .959 & \textit{ .816} & \textbf{ .980} & \cellcolor[rgb]{ .725,  .788,  .898} .925 & \cellcolor[rgb]{ .839,  .855,  .859} .059 \\
    rib right 8 &  .961 &  .922 &  .922 &  .961 & \textit{ .882} & \textbf{ .980} & \cellcolor[rgb]{ .718,  .784,  .902} .938 & \cellcolor[rgb]{ .773,  .816,  .882} .036 \\
    rib right 9 &  .941 & \textit{ .902} & \textit{ .902} &  .941 & \textit{ .902} & \textbf{ .961} & \cellcolor[rgb]{ .729,  .788,  .898} .925 & \cellcolor[rgb]{ .745,  .796,  .894} .026 \\
    sacrum & \textbf{ .955} & \textit{ .909} & \textbf{ .955} & \textit{ .909} & \textit{ .909} &  .932 & \cellcolor[rgb]{ .725,  .788,  .898} .928 & \cellcolor[rgb]{ .733,  .792,  .898} .022 \\
    scapula left & \textbf{ .902} &  .878 & \textbf{ .902} & \textit{ .780} &  .854 &  .854 & \cellcolor[rgb]{ .78,  .824,  .878} .862 & \cellcolor[rgb]{ .8,  .831,  .875} .045 \\
    scapula right & \textbf{ .930} &  .884 &  .884 & \textit{ .767} &  .860 &  .860 & \cellcolor[rgb]{ .78,  .82,  .878} .864 & \cellcolor[rgb]{ .827,  .847,  .863} .054 \\
    small bowel & \textbf{ .938} & \textit{ .854} &  .896 &  .917 &  .896 &  .896 & \cellcolor[rgb]{ .749,  .804,  .89} .899 & \cellcolor[rgb]{ .749,  .8,  .89} .028 \\
    spleen & \textbf{ .980} &  .940 &  .940 &  .940 & \textit{ .900} & \textbf{ .980} & \cellcolor[rgb]{ .71,  .78,  .902} .947 & \cellcolor[rgb]{ .757,  .804,  .89} .030 \\
    stomach & \textbf{ .980} &  .941 &  .941 &  .941 & \textit{ .902} & \textbf{ .980} & \cellcolor[rgb]{ .706,  .776,  .906}\textbf{ .948} & \cellcolor[rgb]{ .753,  .804,  .89} .030 \\
    trachea & \textbf{ .951} &  .927 &  .902 & \textit{ .805} & \textit{ .805} &  .854 & \cellcolor[rgb]{ .773,  .816,  .882} .874 & \cellcolor[rgb]{ .851,  .863,  .855} .063 \\
    urinary bladder & \textbf{ .977} &  .953 &  .953 & \textit{ .907} & \textit{ .907} &  .930 & \cellcolor[rgb]{ .718,  .784,  .902} .938 & \cellcolor[rgb]{ .749,  .8,  .89} .028 \\
    vertebrae C1 & \textit{ .643} & \textit{ .643} & \textit{ .643} & \textit{ .643} & \textbf{ .714} & \textit{ .643} & \cellcolor[rgb]{ .961,  .925,  .812} .655 & \cellcolor[rgb]{ .753,  .804,  .89} .029 \\
    vertebrae C2 & \textit{ .692} & \textit{ .692} & \textit{ .692} & \textit{ .692} & \textbf{ .769} & \textit{ .692} & \cellcolor[rgb]{ .918,  .902,  .827} .705 & \cellcolor[rgb]{ .761,  .808,  .886} .031 \\
    vertebrae C3 &  .643 &  .714 & \textit{ .571} &  .714 & \textbf{ .857} &  .714 & \cellcolor[rgb]{ .922,  .902,  .827} .702 & \cellcolor[rgb]{ .949,  .918,  .82} .095 \\
    vertebrae C4 &  .600 &  .667 & \textit{ .533} &  .667 & \textbf{ .867} &  .667 & \cellcolor[rgb]{ .953,  .922,  .816} .667 & \cellcolor[rgb]{ 1,  .949,  .8}\textit{ .112} \\
    vertebrae C5 &  .650 &  .600 &  .600 & \textit{ .500} & \textbf{ .700} &  .600 & \cellcolor[rgb]{ 1,  .949,  .8}\textit{ .608} & \cellcolor[rgb]{ .863,  .867,  .851} .066 \\
    vertebrae C6 & \textbf{ .818} &  .758 &  .788 &  .758 & \textit{ .606} &  .636 & \cellcolor[rgb]{ .898,  .89,  .835} .727 & \cellcolor[rgb]{ .922,  .902,  .831} .086 \\
    vertebrae C7 & \textbf{ .972} &  .944 &  .917 &  .833 & \textit{ .806} &  .861 & \cellcolor[rgb]{ .757,  .808,  .886} .889 & \cellcolor[rgb]{ .863,  .867,  .851} .066 \\
    vertebrae L1 & \textbf{ .959} &  .918 &  .918 &  .918 & \textit{ .878} & \textbf{ .959} & \cellcolor[rgb]{ .725,  .788,  .898} .925 & \cellcolor[rgb]{ .757,  .808,  .89} .031 \\
    vertebrae L2 &  .909 & \textit{ .886} &  .909 & \textit{ .886} & \textit{ .886} & \textbf{ .977} & \cellcolor[rgb]{ .741,  .8,  .89} .909 & \cellcolor[rgb]{ .773,  .816,  .882} .035 \\
    vertebrae L3 &  .932 &  .841 &  .932 &  .886 & \textit{ .818} & \textbf{ .955} & \cellcolor[rgb]{ .753,  .804,  .886} .894 & \cellcolor[rgb]{ .831,  .847,  .863} .055 \\
    vertebrae L4 &  .955 & \textit{ .864} &  .955 &  .909 &  .909 & \textbf{ .977} & \cellcolor[rgb]{ .725,  .788,  .898} .928 & \cellcolor[rgb]{ .792,  .824,  .878} .042 \\
    vertebrae L5 & \textbf{ .953} & \textit{ .884} & \textbf{ .953} &  .907 &  .907 & \textbf{ .953} & \cellcolor[rgb]{ .725,  .788,  .898} .926 & \cellcolor[rgb]{ .757,  .808,  .89} .031 \\
    vertebrae T1 & \textbf{ .973} &  .946 &  .919 &  .838 & \textit{ .811} &  .892 & \cellcolor[rgb]{ .753,  .804,  .886} .896 & \cellcolor[rgb]{ .851,  .863,  .855} .063 \\
    vertebrae T10 &  .918 &  .898 &  .918 &  .918 & \textit{ .837} & \textbf{ .980} & \cellcolor[rgb]{ .737,  .796,  .894} .912 & \cellcolor[rgb]{ .804,  .831,  .871} .046 \\
    vertebrae T11 &  .958 &  .917 &  .917 &  .938 & \textit{ .875} & \textbf{ .979} & \cellcolor[rgb]{ .722,  .788,  .898} .931 & \cellcolor[rgb]{ .776,  .816,  .882} .036 \\
    vertebrae T12 &  .960 & \textit{ .900} &  .920 &  .920 & \textit{ .900} & \textbf{ .980} & \cellcolor[rgb]{ .722,  .788,  .898} .930 & \cellcolor[rgb]{ .765,  .812,  .886} .033 \\
    vertebrae T2 & \textbf{ .974} &  .947 &  .921 &  .842 & \textit{ .816} &  .895 & \cellcolor[rgb]{ .749,  .804,  .89} .899 & \cellcolor[rgb]{ .847,  .859,  .855} .061 \\
    vertebrae T3 & \textbf{ .947} &  .921 &  .895 & \textit{ .816} & \textit{ .816} &  .868 & \cellcolor[rgb]{ .769,  .816,  .882} .877 & \cellcolor[rgb]{ .827,  .847,  .863} .054 \\
    vertebrae T4 & \textbf{ .949} & \textbf{ .949} &  .923 & \textit{ .821} & \textit{ .821} &  .872 & \cellcolor[rgb]{ .757,  .808,  .886} .889 & \cellcolor[rgb]{ .843,  .859,  .859} .060 \\
    vertebrae T5 & \textbf{ .949} &  .923 &  .872 & \textit{ .821} & \textit{ .821} & \textit{ .821} & \cellcolor[rgb]{ .776,  .82,  .878} .868 & \cellcolor[rgb]{ .835,  .851,  .859} .057 \\
    vertebrae T6 & \textbf{ .944} & \textbf{ .944} &  .917 & \textit{ .833} & \textit{ .833} &  .889 & \cellcolor[rgb]{ .753,  .808,  .886} .894 & \cellcolor[rgb]{ .82,  .843,  .867} .051 \\
    vertebrae T7 & \textbf{ .872} & \textit{ .821} &  .846 & \textit{ .821} & \textit{ .821} &  .846 & \cellcolor[rgb]{ .804,  .835,  .871} .838 & \cellcolor[rgb]{ .729,  .788,  .898} .021 \\
    vertebrae T8 &  .867 & \textit{ .800} &  .822 &  .844 &  .822 & \textbf{ .889} & \cellcolor[rgb]{ .8,  .831,  .871} .841 & \cellcolor[rgb]{ .765,  .808,  .886} .033 \\
    vertebrae T9 &  .878 &  .857 &  .857 &  .898 & \textit{ .796} & \textbf{ .939} & \cellcolor[rgb]{ .776,  .816,  .878} .871 & \cellcolor[rgb]{ .808,  .835,  .871} .048 \\ \hline
    Average & \textbf{ .923} &  .887 &  .892 &  .873 & \textit{ .856} &  .908 &       &  \\
    STD   &  .077 &  .071 &  .080 & \textit{ .082} & \textbf{ .054} &  .081 &       &  \\

    \label{tab:volumewise-recall-104-regions}
    \end{longtable}
\endgroup

\subsubsection{Region-based}

This section presents the recall of region-based retrieval. 
An overview of the evaluation is shown in \Cref{fig:region-based}. 
In region-based retrieval, per each anatomical region in the query volume, one volume is retrieved. 
In the recall computation, the classes present in both the sub-volume of the query and the corresponding retrieved volume are considered TP classes.  
The classes that are present in the query sub-volume and are missing from the retrieved volume are considered FN.

\Cref{tab:regionbased-recall-29-regions} and \Cref{tab:regionbased-recall-104-regions} present the retrieval recalls. 
Compared to volume-based retrieval the average retrieval for the regions is higher. 
The performance of the models is very close. 
DreamSim performs slightly better with an average recall of $ .979 \pm  .037$ for coarse anatomical regions and $ .983 \pm  .032$ for 104 anatomical regions. 
The retrieval recall for many classes is $1.0$. The standard deviation among classes and the models is low, with the highest standard deviation of $ .076$ and $ .092$, for coarse and fine classes respectively.

\begin{table}[!htbp]
 \centering
 \renewcommand*{\arraystretch}{ .50} 
  \caption{Region-based retrieval recall of coarse anatomical regions (29 classes) using HNSW Indexing. In each row, bold numbers represent the best-performing values, while italicized numbers indicate the worst-performing. The separate average and standard deviation (STD) columns are color-coded, with blue indicating the best-performing values and yellow indicating the worst-performing values across different models. Additionally, bold numbers in colored columns represent the best classes in terms of average and standard deviation, while italicized values represent the worst-performing class across the models.}
  \scriptsize
\begin{tabular}{lrrrrrr|rr}
    Model & DINOv1 & DINOv2 & DreamSim & SwinTrans. & \multicolumn{2}{c|}{ResNet50} & & \\ \cline{1-7}
      Dataset (pre-trained) & (ImgNet) & (ImgNet) & (ImgNet)& (RadImg)& 
    (Fractaldb) & (RadImg) &Average & STD \\ \hline
    adrenal gland & \textbf{1.000} & \textbf{1.000} & \textbf{1.000} & \textit{ .970} &  .990 & \textbf{1.000} & \cellcolor[rgb]{ .722,  .784,  .898} .993 & \cellcolor[rgb]{ .753,  .804,  .89} .012 \\
    autochthon & \textit{ .992} & \textit{ .992} & \textit{ .992} & \textit{ .992} & \textit{ .992} & \textit{ .992} & \cellcolor[rgb]{ .722,  .788,  .898} .992 & \cellcolor[rgb]{ .706,  .776,  .906}\textbf{ .000} \\
    brain & \textit{ .846} & \textit{ .846} & \textbf{1.000} &  .923 & \textbf{1.000} & \textbf{1.000} & \cellcolor[rgb]{ .835,  .855,  .859} .936 & \cellcolor[rgb]{ 1,  .949,  .8}\textit{ .076} \\
    cardiovascular system & \textbf{1.000} & \textbf{1.000} & \textbf{1.000} & \textbf{1.000} & \textbf{1.000} & \textbf{1.000} & \cellcolor[rgb]{ .706,  .776,  .906}\textbf{1.000} & \cellcolor[rgb]{ .706,  .776,  .906}\textbf{ .000} \\
    clavicula &  .987 &  .987 & \textbf{1.000} & \textbf{1.000} & \textit{ .961} &  .987 & \cellcolor[rgb]{ .733,  .792,  .894} .987 & \cellcolor[rgb]{ .761,  .808,  .886} .014 \\
    colon & \textbf{1.000} & \textbf{1.000} & \textbf{1.000} & \textbf{1.000} & \textbf{1.000} & \textbf{1.000} & \cellcolor[rgb]{ .706,  .776,  .906}\textbf{1.000} & \cellcolor[rgb]{ .706,  .776,  .906}\textbf{ .000} \\
\end{tabular}%
\end{table}%

\FloatBarrier

\begingroup
\renewcommand*{\arraystretch}{ .50} 
\scriptsize
\begin{longtable}{lcccccc|cc}
\caption{Regiond-based retrieval recall of all TS anatomical regions (104 classes) using HNSW Indexing. In each row, bold numbers represent the best-performing values, while italicized numbers indicate the worst-performing. The separate average and standard deviation (STD) columns are color-coded, with blue indicating the best-performing values and yellow indicating the worst-performing values across different models. Additionally, bold numbers in colored columns represent the best classes in terms of average and standard deviation, while italicized values represent the worst-performing class across the models.}  \\
Model & DINOv1 & DINOv2 & DreamSim & SwinTrans. & \multicolumn{2}{c|}{ResNet50} & & \\ \cline{1-7}
      Dataset (pre-trained) & (ImgNet) & (ImgNet) & (ImgNet)& (RadImg)& 
    (Fractaldb) & (RadImg) & Average & STD \\ \hline

    adrenal gland left &  .980 & \textbf{1.000} & \textbf{1.000} & \textit{ .940} & \textbf{1.000} & \textbf{1.000} & \cellcolor[rgb]{ .729,  .792,  .894} .987 & \cellcolor[rgb]{ .78,  .82,  .878} .024 \\
    adrenal gland right & \textbf{1.000} & \textbf{1.000} & \textbf{1.000} &  .980 & \textit{ .960} & \textbf{1.000} & \cellcolor[rgb]{ .725,  .788,  .898} .990 & \cellcolor[rgb]{ .757,  .808,  .89} .017 \\
    aorta & \textbf{1.000} & \textbf{1.000} & \textbf{1.000} & \textit{ .984} & \textbf{1.000} & \textbf{1.000} & \cellcolor[rgb]{ .714,  .78,  .902} .997 & \cellcolor[rgb]{ .725,  .788,  .902} .007 \\
    autochthon left & \textbf{1.000} & \textbf{1.000} & \textbf{1.000} & \textit{ .985} & \textbf{1.000} & \textit{ .985} & \cellcolor[rgb]{ .718,  .784,  .902} .995 & \cellcolor[rgb]{ .729,  .788,  .898} .008 \\
    autochthon right & \textit{ .985} & \textit{ .985} & \textit{ .985} & \textit{ .985} & \textit{ .985} & \textbf{1.000} & \cellcolor[rgb]{ .729,  .792,  .894} .987 & \cellcolor[rgb]{ .725,  .788,  .902} .006 \\
\end{longtable}
\endgroup

\subsubsection{Localized}

This section presents the recall and localization-ratio of localized retrieval. 
An overview of the evaluation is shown in \Cref{fig:localized}. 
In localized retrieval, per each anatomical region in the query volume, one volume is retrieved. 

\paragraph{Localized Retrieval Recall} 
The recall calculation for localized retrieval is explained in \Cref{sec:localization} and an overview is shown in \Cref{fig:localized}. \Cref{tab:localized-recall-29-regions} and \Cref{tab:localized-recall-104-regions} present the retrieval recalls. Compared to region-based retrieval the average retrieval for regions is lower which is expected based on the more strict metric defined. 
The performance of models is close, especially the self-supervised models. 
DINOv2 performs best for 29 anatomical regions with an average recall of $.941 \pm .077$. 
For 104 regions the performance of models is even closer with DINOv1 performing slightly better with an average recall of $.929 \pm .085$.

\begin{table}[!htbp]
 \centering
 \renewcommand*{\arraystretch}{ .50} 
  \caption{Localized retrieval recall of coarse anatomical regions (29 classes) using HNSW Indexing, $L=15$. In each row, bold numbers represent the best-performing values, while italicized numbers indicate the worst-performing. The separate average and standard deviation (STD) columns are color-coded, with blue indicating the best-performing values and yellow indicating the worst-performing values across different models. Additionally, bold numbers in colored columns represent the best classes in terms of average and standard deviation, while italicized values represent the worst-performing class across the models.}
  \scriptsize
\begin{tabular}{lrrrrrr|rr}
    Model & DINOv1 & DINOv2 & DreamSim & SwinTrans. & \multicolumn{2}{c|}{ResNet50} & & \\ \cline{1-7}
      Dataset (pre-trained) & (ImgNet) & (ImgNet) & (ImgNet)& (RadImg)& 
    (Fractaldb) & (RadImg) &Average & STD \\ \hline
    adrenal gland & \textbf{  .960} &   .920 &   .920 &   .850 & \textit{  .790} &   .860 & \cellcolor[rgb]{ .78,  .82,  .878}  .883 & \cellcolor[rgb]{ .886,  .882,  .843}  .062 \\
    autochthon &   .985 & \textbf{  .992} &   .969 &   .985 & \textit{  .954} & \textbf{  .992} & \cellcolor[rgb]{ .714,  .784,  .902}  .979 & \cellcolor[rgb]{ .722,  .784,  .902}  .015 \\
    brain & \textit{  .846} & \textit{  .846} & \textbf{1.000} &   .923 & \textit{  .846} & \textbf{1.000} & \cellcolor[rgb]{ .761,  .812,  .886}  .910 & \cellcolor[rgb]{ .937,  .91,  .824}  .076 \\
    cardiovascular system & \textbf{1.000} &   .998 &   .995 &   .997 &   .983 & \textit{  .976} & \cellcolor[rgb]{ .706,  .776,  .906}  .991 & \cellcolor[rgb]{ .706,  .776,  .906}  .010 \\
    clavicula &   .935 & \textbf{  .987} &   .974 &   .961 & \textit{  .857} & \textbf{  .987} & \cellcolor[rgb]{ .733,  .796,  .894}  .950 & \cellcolor[rgb]{ .843,  .859,  .859}  .050 \\
    colon &   .962 & \textbf{1.000} & \textbf{1.000} &   .942 & \textit{  .923} &   .962 & \cellcolor[rgb]{ .725,  .788,  .898}  .965 & \cellcolor[rgb]{ .78,  .82,  .882}  .031 \\
\end{tabular}%
  \label{tab:localized-recall-29-regions}%
\end{table}%

\FloatBarrier

\begingroup
\renewcommand*{\arraystretch}{ .50} 
\scriptsize
\begin{longtable}{lcccccc|cc}
\caption{Localized retrieval recall of all TS anatomical regions (104 classes) using HNSW Indexing, $L=15$. In each row, bold numbers represent the best-performing values, while italicized numbers indicate the worst-performing. The separate average and standard deviation (STD) columns are color-coded, with blue indicating the best-performing values and yellow indicating the worst-performing values across different models. Additionally, bold numbers in colored columns represent the best classes in terms of average and standard deviation, while italicized values represent the worst-performing class across the models.}  \\
Model & DINOv1 & DINOv2 & DreamSim & SwinTrans. & \multicolumn{2}{c|}{ResNet50} & & \\ \cline{1-7}
      Dataset (pre-trained) & (ImgNet) & (ImgNet) & (ImgNet)& (RadImg)& 
    (Fractaldb) & (RadImg) & Average & STD \\ \hline
    adrenal  gland  left &   .900 & \textbf{  .920} &   .860 & \textit{  .740} &   .760 & \textit{  .740} & \cellcolor[rgb]{ .82,  .843,  .863}  .820 & \cellcolor[rgb]{ .827,  .847,  .863}  .083 \\
    adrenal  gland  right & \textbf{  .880} &   .840 & \textbf{  .880} &   .840 & \textit{  .620} &   .840 & \cellcolor[rgb]{ .824,  .847,  .863}  .817 & \cellcolor[rgb]{ .855,  .863,  .855}  .098 \\
    aorta & \textbf{1.000} & \textit{  .967} & \textbf{1.000} & \textit{  .967} &   .984 & \textit{  .967} & \cellcolor[rgb]{ .714,  .78,  .902}  .981 & \cellcolor[rgb]{ .71,  .776,  .906}  .016 \\
    autochthon  left &   .985 & \textbf{1.000} &   .969 &   .985 & \textit{  .954} &   .985 & \cellcolor[rgb]{ .714,  .78,  .902}  .979 & \cellcolor[rgb]{ .71,  .776,  .906}  .016 \\
    autochthon  right &   .985 &   .985 &   .969 &   .985 & \textit{  .954} & \textbf{1.000} & \cellcolor[rgb]{ .714,  .78,  .902}  .979 & \cellcolor[rgb]{ .71,  .776,  .906}  .016 \\
\end{longtable}
\endgroup

\paragraph{Localization-ratio}

The localization-ratio is computed based on \eqref{eq:normalized-location-ratio}.
This measure shows how many slices that contributed to the retrieval of the volume actually contained the desired organ. 
\Cref{tab:localization-ratio-29-regions} and \Cref{tab:localization-ratio-104-regions} show the localization-ratio for 29 coarse and 104 TS original classes. 
DreamSim shows the best average localization-ratio with an average localization-ratio of $.864\pm.145$ and $.803\pm.130$ for coarse and original TS classes, respectively.

\begin{table}[!htbp]
 \centering
 \renewcommand*{\arraystretch}{ .50} 
  \caption{Localization-ratio of coarse anatomical regions (29 classes) using HNSW Indexing, $L=15$. In each row, bold numbers represent the best-performing values, while italicized numbers indicate the worst-performing. The separate average and standard deviation (STD) columns are color-coded, with blue indicating the best-performing values and yellow indicating the worst-performing values across different models. Additionally, bold numbers in colored columns represent the best classes in terms of average and standard deviation, while italicized values represent the worst-performing class across the models.}
  \scriptsize
    \begin{tabular}{lrrrrrr|rr}
    Model & DINOv1 & DINOv2 & DreamSim & SwinTrans. & \multicolumn{2}{c|}{ResNet50} & & \\ \cline{1-7}
      Dataset (pre-trained) & (ImgNet) & (ImgNet) & (ImgNet)& (RadImg)& 
    (Fractaldb) & (RadImg) &Average & STD \\ \hline
    adrenal gland & \textbf{.801} & .651 & .724 & .669 & \textit{.525} & .617 & \cellcolor[rgb]{ .851,  .863,  .851}.665 & \cellcolor[rgb]{ .996,  .945,  .804}.094 \\
    autochthon & .971 & .968 & .960 & \textbf{.973} & \textit{.938} & .963 & \cellcolor[rgb]{ .718,  .784,  .898}.962 & \cellcolor[rgb]{ .706,  .776,  .906}\textbf{.013} \\
    brain & .842 & .837 & \textbf{.941} & .869 & \textit{.835} & .873 & \cellcolor[rgb]{ .761,  .812,  .886}.866 & \cellcolor[rgb]{ .804,  .831,  .871}.040 \\
    cardiovascular system & \textbf{.998} & .993 & .995 & .995 & .954 & \textit{.954} & \cellcolor[rgb]{ .71,  .78,  .902}.982 & \cellcolor[rgb]{ .733,  .792,  .898}.021 \\
    clavicula & .890 & .945 & \textbf{.961} & .923 & \textit{.788} & .952 & \cellcolor[rgb]{ .741,  .8,  .89}.910 & \cellcolor[rgb]{ .89,  .886,  .839}.065 \\
    colon & .914 & .963 & \textbf{.964} & .865 & \textit{.831} & .892 & \cellcolor[rgb]{ .745,  .8,  .89}.905 & \cellcolor[rgb]{ .851,  .859,  .855}.053 \\
    duodenum & \textit{.627} & .696 & \textbf{.765} & .656 & .654 & .747 & \cellcolor[rgb]{ .839,  .855,  .855}.691 & \cellcolor[rgb]{ .859,  .863,  .851}.055 \\
    esophagus & .895 & .913 & .931 & \textbf{.961} & .924 & \textit{.858} & \cellcolor[rgb]{ .741,  .796,  .89}.914 & \cellcolor[rgb]{ .784,  .824,  .878}.035 \\
    face  & .748 & .774 & .787 & \textit{.733} & \textbf{.804} & .736 & \cellcolor[rgb]{ .808,  .835,  .867}.763 & \cellcolor[rgb]{ .761,  .808,  .886}.029 \\
    femur & \textit{.873} & .948 & \textbf{.990} & .948 & .925 & .960 & \cellcolor[rgb]{ .729,  .792,  .898}.940 & \cellcolor[rgb]{ .8,  .831,  .875}.039 \\
    gallbladder & \textit{.219} & \textbf{.388} & .302 & .320 & .363 & .369 & \cellcolor[rgb]{ 1,  .949,  .8}\textit{.327} & \cellcolor[rgb]{ .882,  .878,  .843}.062 \\
    gluteus muscles & .986 & \textit{.954} & .990 & \textbf{.991} & .976 & .989 & \cellcolor[rgb]{ .71,  .78,  .902}.981 & \cellcolor[rgb]{ .71,  .776,  .906}.014 \\
    hip   & \textbf{.978} & .951 & .976 & .945 & .951 & \textit{.931} & \cellcolor[rgb]{ .722,  .788,  .898}.955 & \cellcolor[rgb]{ .725,  .784,  .902}.018 \\
    humerus & .592 & .581 & \textbf{.749} & .583 & \textit{.452} & .571 & \cellcolor[rgb]{ .886,  .882,  .839}.588 & \cellcolor[rgb]{ 1,  .949,  .8}\textit{.094} \\
    iliopsoas & .929 & .912 & \textbf{.937} & .912 & .909 & \textit{.900} & \cellcolor[rgb]{ .741,  .796,  .894}.916 & \cellcolor[rgb]{ .706,  .776,  .906}.014 \\
    kidney & .813 & .819 & .804 & \textbf{.859} & \textit{.757} & .824 & \cellcolor[rgb]{ .784,  .824,  .875}.813 & \cellcolor[rgb]{ .776,  .82,  .882}.033 \\
    liver & \textbf{.873} & .841 & \textit{.830} & .831 & .832 & .860 & \cellcolor[rgb]{ .773,  .816,  .882}.844 & \cellcolor[rgb]{ .722,  .784,  .902}.018 \\
    lung  & \textbf{.979} & .976 & .962 & .969 & .928 & \textit{.878} & \cellcolor[rgb]{ .725,  .788,  .898}.949 & \cellcolor[rgb]{ .8,  .831,  .875}.039 \\
    pancreas & .697 & \textbf{.787} & .778 & .708 & \textit{.573} & .770 & \cellcolor[rgb]{ .827,  .847,  .859}.719 & \cellcolor[rgb]{ .949,  .918,  .82}.081 \\
    portal and splenic vein & \textbf{.725} & .652 & .692 & .630 & .552 & \textit{.493} & \cellcolor[rgb]{ .871,  .875,  .847}.624 & \cellcolor[rgb]{ .973,  .933,  .812}.087 \\
    rib   & .964 & .970 & \textbf{.999} & .972 & .972 & \textit{.933} & \cellcolor[rgb]{ .718,  .784,  .902}.968 & \cellcolor[rgb]{ .733,  .792,  .898}.021 \\
    sacrum & .875 & .860 & \textbf{.962} & .889 & .932 & \textit{.855} & \cellcolor[rgb]{ .749,  .804,  .89}.895 & \cellcolor[rgb]{ .812,  .839,  .871}.043 \\
    scapula & \textbf{.936} & .897 & .902 & .889 & .908 & \textit{.821} & \cellcolor[rgb]{ .749,  .804,  .89}.892 & \cellcolor[rgb]{ .796,  .827,  .875}.038 \\
    small bowel & \textbf{.905} & \textit{.818} & .889 & .881 & .847 & .895 & \cellcolor[rgb]{ .761,  .808,  .886}.872 & \cellcolor[rgb]{ .776,  .816,  .882}.033 \\
    spleen & \textbf{.864} & .742 & .736 & .735 & .815 & \textit{.710} & \cellcolor[rgb]{ .808,  .835,  .867}.767 & \cellcolor[rgb]{ .871,  .871,  .847}.059 \\
    stomach & \textbf{.876} & \textit{.780} & .827 & .815 & .833 & .807 & \cellcolor[rgb]{ .78,  .824,  .878}.823 & \cellcolor[rgb]{ .773,  .816,  .882}.032 \\
    trachea & .907 & .874 & \textbf{.915} & .912 & \textit{.797} & .805 & \cellcolor[rgb]{ .761,  .808,  .886}.868 & \cellcolor[rgb]{ .855,  .863,  .855}.055 \\
    urinary bladder & .752 & .709 & \textbf{.804} & .776 & \textit{.670} & .685 & \cellcolor[rgb]{ .82,  .847,  .863}.733 & \cellcolor[rgb]{ .847,  .859,  .855}.053 \\
    vertebrae & .998 & .998 & .972 & \textbf{.998} & \textit{.970} & .995 & \cellcolor[rgb]{ .706,  .776,  .906}\textbf{.988} & \cellcolor[rgb]{ .706,  .776,  .906}.014 \\ \hline
    Average & .842 & .834 & \textbf{.864} & .835 & \textit{.800} & .815 &       &  \\ 
    STD   & .161 & \textbf{.144} & \textbf{.145} & .155 & \textit{.168} & .152 &       &  \\
        \end{tabular}%
  \label{tab:localization-ratio-29-regions}%
\end{table}%

\FloatBarrier

\begingroup
\renewcommand*{\arraystretch}{ .50} 
\scriptsize
\begin{longtable}{lcccccc|cc}
\caption{Localization-ratio of all TS anatomical regions (104 classes) using HNSW Indexing, $L=15$. In each row, bold numbers represent the best-performing values, while italicized numbers indicate the worst-performing. The separate average and standard deviation (STD) columns are color-coded, with blue indicating the best-performing values and yellow indicating the worst-performing values across different models. Additionally, bold numbers in colored columns represent the best classes in terms of average and standard deviation, while italicized values represent the worst-performing class across the models.}  \\
Model & DINOv1 & DINOv2 & DreamSim & SwinTrans. & \multicolumn{2}{c|}{ResNet50} & & \\ \cline{1-7}
      Dataset (pre-trained) & (ImgNet) & (ImgNet) & (ImgNet)& (RadImg)& 
    (Fractaldb) & (RadImg) & Average & STD \\ \hline
    adrenal gland left & \textbf{.683} & .569 & .602 & .526 & .488 & \textit{.431} & \cellcolor[rgb]{ .898,  .89,  .835}.550 & \cellcolor[rgb]{ .855,  .863,  .855}.089 \\
    adrenal gland right & \textbf{.654} & .525 & .641 & .578 & \textit{.410} & .535 & \cellcolor[rgb]{ .894,  .89,  .835}.557 & \cellcolor[rgb]{ .855,  .863,  .855}.089 \\
    aorta & .943 & .921 & \textbf{.967} & .936 & .927 & \textit{.889} & \cellcolor[rgb]{ .725,  .788,  .898}.931 & \cellcolor[rgb]{ .729,  .788,  .898}.026 \\
    autochthon left & .969 & .980 & .960 & \textbf{.983} & \textit{.938} & .964 & \cellcolor[rgb]{ .706,  .776,  .906}\textbf{.966} & \cellcolor[rgb]{ .714,  .78,  .906}.016 \\
    autochthon right & \textbf{.972} & .956 & .954 & .968 & \textit{.938} & .964 & \cellcolor[rgb]{ .71,  .78,  .902}.959 & \cellcolor[rgb]{ .706,  .776,  .906}\textbf{.012} \\
    brain & .842 & .837 & \textbf{.941} & .869 & \textit{.835} & .873 & \cellcolor[rgb]{ .753,  .804,  .886}.866 & \cellcolor[rgb]{ .761,  .808,  .886}.040 \\
    clavicula left & .871 & \textbf{.908} & .891 & .882 & \textit{.827} & .885 & \cellcolor[rgb]{ .749,  .804,  .89}.877 & \cellcolor[rgb]{ .733,  .792,  .898}.027 \\
    clavicula right & .859 & \textbf{.924} & .923 & .886 & \textit{.752} & .924 & \cellcolor[rgb]{ .749,  .804,  .89}.878 & \cellcolor[rgb]{ .812,  .839,  .871}.067 \\
    colon & .914 & .963 & \textbf{.964} & .865 & \textit{.831} & .892 & \cellcolor[rgb]{ .737,  .796,  .894}.905 & \cellcolor[rgb]{ .784,  .824,  .878}.053 \\
    duodenum & \textit{.627} & .696 & \textbf{.765} & .656 & .654 & .747 & \cellcolor[rgb]{ .835,  .851,  .859}.691 & \cellcolor[rgb]{ .788,  .824,  .878}.055 \\
    esophagus & .895 & .913 & .931 & \textbf{.961} & .924 & \textit{.858} & \cellcolor[rgb]{ .733,  .792,  .894}.914 & \cellcolor[rgb]{ .749,  .8,  .89}.035 \\
    face  & .748 & .774 & .787 & \textit{.733} & \textbf{.804} & .736 & \cellcolor[rgb]{ .8,  .831,  .871}.763 & \cellcolor[rgb]{ .737,  .792,  .894}.029 \\
    femur left & \textit{.836} & .907 & .908 & \textbf{.928} & .868 & .862 & \cellcolor[rgb]{ .745,  .8,  .89}.885 & \cellcolor[rgb]{ .749,  .8,  .89}.035 \\
    femur right & \textit{.870} & .922 & .949 & \textbf{.951} & .916 & .947 & \cellcolor[rgb]{ .725,  .788,  .898}.926 & \cellcolor[rgb]{ .741,  .796,  .894}.031 \\
    gallbladder & \textit{.219} & \textbf{.388} & .302 & .320 & .363 & .369 & \cellcolor[rgb]{ 1,  .949,  .8}\textit{.327} & \cellcolor[rgb]{ .8,  .831,  .875}.062 \\
    gluteus maximus left & \textbf{.981} & .921 & .956 & .947 & \textit{.868} & .896 & \cellcolor[rgb]{ .725,  .788,  .898}.928 & \cellcolor[rgb]{ .761,  .808,  .886}.041 \\
    gluteus maximus right & \textbf{.992} & .928 & .945 & .916 & .937 & \textit{.913} & \cellcolor[rgb]{ .722,  .784,  .898}.938 & \cellcolor[rgb]{ .737,  .792,  .894}.029 \\
    gluteus medius left & \textbf{.966} & .877 & .947 & .936 & .904 & \textit{.865} & \cellcolor[rgb]{ .729,  .792,  .894}.916 & \cellcolor[rgb]{ .761,  .808,  .886}.040 \\
    gluteus medius right & .932 & .911 & \textbf{.973} & .912 & .944 & \textit{.900} & \cellcolor[rgb]{ .725,  .788,  .898}.929 & \cellcolor[rgb]{ .733,  .792,  .898}.027 \\
    gluteus minimus left & \textbf{.907} & .834 & .907 & .903 & \textit{.834} & .899 & \cellcolor[rgb]{ .745,  .8,  .89}.881 & \cellcolor[rgb]{ .749,  .804,  .89}.036 \\
    gluteus minimus right & .915 & \textit{.790} & .852 & .906 & .861 & \textbf{.947} & \cellcolor[rgb]{ .749,  .804,  .89}.878 & \cellcolor[rgb]{ .788,  .824,  .878}.056 \\
    heart atrium left & .727 & .703 & \textbf{.756} & .671 & .672 & \textit{.452} & \cellcolor[rgb]{ .847,  .859,  .855}.664 & \cellcolor[rgb]{ .894,  .886,  .839}.109 \\
    heart atrium right & .866 & \textit{.785} & \textbf{.912} & .794 & .787 & .794 & \cellcolor[rgb]{ .773,  .816,  .878}.823 & \cellcolor[rgb]{ .784,  .824,  .878}.053 \\
    heart myocardium & .881 & .794 & \textbf{.903} & .840 & .815 & \textit{.782} & \cellcolor[rgb]{ .769,  .812,  .882}.836 & \cellcolor[rgb]{ .773,  .816,  .882}.048 \\
    heart ventricle left & .850 & \textit{.764} & \textbf{.879} & .864 & .807 & .814 & \cellcolor[rgb]{ .769,  .816,  .882}.830 & \cellcolor[rgb]{ .765,  .808,  .886}.043 \\
    heart ventricle right & .857 & .833 & \textbf{.876} & .839 & .850 & \textit{.769} & \cellcolor[rgb]{ .769,  .812,  .882}.837 & \cellcolor[rgb]{ .753,  .804,  .89}.037 \\
    hip left & \textbf{.981} & .949 & .974 & .929 & .952 & \textit{.913} & \cellcolor[rgb]{ .714,  .784,  .902}.950 & \cellcolor[rgb]{ .729,  .788,  .898}.026 \\
    hip right & \textbf{.980} & .941 & .972 & \textit{.941} & .972 & .948 & \cellcolor[rgb]{ .71,  .78,  .902}.959 & \cellcolor[rgb]{ .714,  .78,  .906}.017 \\
    humerus left & .559 & .619 & \textbf{.679} & .625 & \textit{.438} & .583 & \cellcolor[rgb]{ .882,  .882,  .839}.584 & \cellcolor[rgb]{ .839,  .855,  .859}.082 \\
    humerus right & .567 & .563 & \textbf{.686} & .515 & \textit{.390} & .542 & \cellcolor[rgb]{ .902,  .894,  .835}.544 & \cellcolor[rgb]{ .867,  .871,  .851}.096 \\
    iliac artery left & .888 & .895 & .918 & \textbf{.933} & .890 & \textit{.847} & \cellcolor[rgb]{ .741,  .796,  .894}.895 & \cellcolor[rgb]{ .737,  .796,  .894}.030 \\
    iliac artery right & .929 & .885 & \textbf{.935} & .911 & .916 & \textit{.833} & \cellcolor[rgb]{ .737,  .796,  .894}.901 & \cellcolor[rgb]{ .753,  .804,  .89}.038 \\
    iliac vena left & .906 & .869 & \textbf{.927} & .916 & .867 & \textit{.860} & \cellcolor[rgb]{ .741,  .8,  .89}.891 & \cellcolor[rgb]{ .737,  .792,  .898}.029 \\
    iliac vena right & .918 & .883 & \textbf{.940} & .901 & .904 & \textit{.839} & \cellcolor[rgb]{ .741,  .796,  .894}.897 & \cellcolor[rgb]{ .745,  .8,  .894}.034 \\
    iliopsoas left & .920 & .926 & \textbf{.954} & .890 & .871 & \textit{.870} & \cellcolor[rgb]{ .737,  .796,  .894}.905 & \cellcolor[rgb]{ .745,  .8,  .894}.034 \\
    iliopsoas right & .922 & .908 & \textbf{.934} & .888 & \textit{.885} & .900 & \cellcolor[rgb]{ .733,  .796,  .894}.906 & \cellcolor[rgb]{ .718,  .78,  .902}.019 \\
    inferior vena cava & \textbf{.944} & .927 & \textit{.894} & .936 & .922 & .912 & \cellcolor[rgb]{ .729,  .788,  .898}.923 & \cellcolor[rgb]{ .714,  .78,  .906}.018 \\
    kidney left & .764 & .663 & \textbf{.798} & .717 & \textit{.637} & .763 & \cellcolor[rgb]{ .82,  .843,  .863}.724 & \cellcolor[rgb]{ .804,  .831,  .871}.063 \\
    kidney right & .720 & .751 & .773 & .752 & \textit{.690} & \textbf{.783} & \cellcolor[rgb]{ .808,  .839,  .867}.745 & \cellcolor[rgb]{ .749,  .8,  .89}.035 \\
    liver & \textbf{.873} & .841 & \textit{.830} & .831 & .832 & .860 & \cellcolor[rgb]{ .765,  .812,  .882}.844 & \cellcolor[rgb]{ .714,  .78,  .906}.018 \\
    lung lower lobe left & .876 & \textbf{.879} & .873 & .865 & .835 & \textit{.813} & \cellcolor[rgb]{ .757,  .808,  .886}.857 & \cellcolor[rgb]{ .733,  .792,  .898}.027 \\
    lung lower lobe right & \textbf{.912} & .855 & .866 & .865 & .817 & \textit{.779} & \cellcolor[rgb]{ .761,  .812,  .886}.849 & \cellcolor[rgb]{ .769,  .812,  .882}.046 \\
    lung middle lobe right & .802 & .832 & .830 & \textbf{.856} & .750 & \textit{.738} & \cellcolor[rgb]{ .784,  .824,  .878}.802 & \cellcolor[rgb]{ .773,  .816,  .882}.048 \\
    lung upper lobe left & .921 & .938 & .925 & \textbf{.945} & \textit{.831} & .844 & \cellcolor[rgb]{ .737,  .796,  .894}.901 & \cellcolor[rgb]{ .776,  .816,  .882}.050 \\
    lung upper lobe right & \textbf{.887} & .879 & .881 & .873 & .848 & \textit{.779} & \cellcolor[rgb]{ .757,  .808,  .886}.858 & \cellcolor[rgb]{ .761,  .808,  .886}.041 \\
    pancreas & .697 & \textbf{.787} & .778 & .708 & \textit{.573} & .770 & \cellcolor[rgb]{ .82,  .843,  .863}.719 & \cellcolor[rgb]{ .839,  .855,  .859}.081 \\
    portal and splenic vein & \textbf{.725} & .652 & .692 & .630 & .552 & \textit{.493} & \cellcolor[rgb]{ .867,  .871,  .847}.624 & \cellcolor[rgb]{ .851,  .859,  .855}.087 \\
    pulmonary artery & .689 & .697 & \textbf{.758} & .647 & \textit{.493} & .575 & \cellcolor[rgb]{ .855,  .867,  .851}.643 & \cellcolor[rgb]{ .867,  .871,  .851}.095 \\
    rib left 1 & .861 & .878 & \textbf{.922} & .848 & .827 & \textit{.826} & \cellcolor[rgb]{ .757,  .808,  .886}.860 & \cellcolor[rgb]{ .753,  .804,  .89}.036 \\
    rib left 10 & \textbf{.878} & .792 & .823 & .829 & .815 & \textit{.790} & \cellcolor[rgb]{ .773,  .816,  .878}.821 & \cellcolor[rgb]{ .741,  .796,  .894}.032 \\
    rib left 11 & .820 & \textit{.768} & .827 & .799 & .854 & \textbf{.861} & \cellcolor[rgb]{ .773,  .816,  .878}.821 & \cellcolor[rgb]{ .749,  .8,  .894}.034 \\
    rib left 12 & .656 & \textit{.480} & \textbf{.697} & .643 & .561 & .649 & \cellcolor[rgb]{ .871,  .875,  .847}.614 & \cellcolor[rgb]{ .835,  .851,  .859}.080 \\
    rib left 2 & \textbf{.860} & .801 & .819 & .785 & \textit{.736} & .770 & \cellcolor[rgb]{ .788,  .824,  .875}.795 & \cellcolor[rgb]{ .765,  .808,  .886}.043 \\
    rib left 3 & .837 & .806 & .799 & .824 & \textbf{.854} & \textit{.743} & \cellcolor[rgb]{ .78,  .82,  .878}.811 & \cellcolor[rgb]{ .757,  .804,  .89}.038 \\
    rib left 4 & \textit{.801} & .829 & \textbf{.861} & .816 & .820 & .816 & \cellcolor[rgb]{ .773,  .816,  .882}.824 & \cellcolor[rgb]{ .722,  .784,  .902}.020 \\
    rib left 5 & .785 & .763 & \textbf{.810} & .774 & \textit{.732} & .753 & \cellcolor[rgb]{ .8,  .831,  .871}.769 & \cellcolor[rgb]{ .733,  .792,  .898}.027 \\
    rib left 6 & \textbf{.824} & .789 & .768 & .774 & \textit{.723} & .789 & \cellcolor[rgb]{ .796,  .827,  .875}.778 & \cellcolor[rgb]{ .745,  .8,  .894}.033 \\
    rib left 7 & .887 & \textbf{.888} & .842 & .856 & \textit{.771} & .808 & \cellcolor[rgb]{ .765,  .812,  .882}.842 & \cellcolor[rgb]{ .769,  .812,  .882}.046 \\
    rib left 8 & .847 & \textbf{.881} & .826 & .826 & \textit{.717} & .819 & \cellcolor[rgb]{ .776,  .82,  .878}.819 & \cellcolor[rgb]{ .788,  .824,  .878}.055 \\
    rib left 9 & .831 & \textbf{.862} & .822 & .860 & \textit{.758} & .845 & \cellcolor[rgb]{ .769,  .816,  .882}.830 & \cellcolor[rgb]{ .757,  .804,  .89}.039 \\
    rib right 1 & .864 & .826 & \textbf{.871} & \textit{.790} & .826 & .857 & \cellcolor[rgb]{ .765,  .812,  .882}.839 & \cellcolor[rgb]{ .741,  .796,  .894}.031 \\
    rib right 10 & \textbf{.890} & \textit{.810} & .826 & .820 & .825 & .840 & \cellcolor[rgb]{ .769,  .812,  .882}.835 & \cellcolor[rgb]{ .737,  .792,  .898}.029 \\
    rib right 11 & \textbf{.863} & \textit{.785} & .855 & .792 & .831 & .794 & \cellcolor[rgb]{ .776,  .82,  .878}.820 & \cellcolor[rgb]{ .749,  .8,  .894}.034 \\
    rib right 12 & .554 & \textit{.488} & .643 & \textbf{.677} & .536 & .562 & \cellcolor[rgb]{ .886,  .882,  .839}.577 & \cellcolor[rgb]{ .816,  .843,  .867}.070 \\
    rib right 2 & \textbf{.883} & .817 & .834 & .844 & \textit{.726} & .850 & \cellcolor[rgb]{ .773,  .816,  .882}.826 & \cellcolor[rgb]{ .784,  .824,  .878}.053 \\
    rib right 3 & \textbf{.867} & .814 & .803 & .827 & .775 & \textit{.724} & \cellcolor[rgb]{ .784,  .824,  .878}.801 & \cellcolor[rgb]{ .776,  .816,  .882}.049 \\
    rib right 4 & .850 & .836 & \textbf{.880} & .813 & .842 & \textit{.766} & \cellcolor[rgb]{ .769,  .816,  .882}.831 & \cellcolor[rgb]{ .757,  .804,  .89}.038 \\
    rib right 5 & .750 & .774 & \textbf{.819} & .753 & .761 & \textit{.746} & \cellcolor[rgb]{ .8,  .831,  .871}.767 & \cellcolor[rgb]{ .733,  .792,  .898}.027 \\
    rib right 6 & \textbf{.793} & .790 & .789 & .757 & .773 & \textit{.729} & \cellcolor[rgb]{ .796,  .831,  .871}.772 & \cellcolor[rgb]{ .729,  .788,  .898}.025 \\
    rib right 7 & .849 & \textbf{.878} & .810 & .853 & \textit{.754} & .824 & \cellcolor[rgb]{ .773,  .816,  .882}.828 & \cellcolor[rgb]{ .765,  .812,  .886}.043 \\
    rib right 8 & .867 & .840 & \textbf{.868} & .846 & \textit{.763} & .834 & \cellcolor[rgb]{ .769,  .812,  .882}.836 & \cellcolor[rgb]{ .757,  .804,  .89}.039 \\
    rib right 9 & .860 & .855 & .804 & \textbf{.865} & \textit{.790} & .814 & \cellcolor[rgb]{ .769,  .816,  .882}.831 & \cellcolor[rgb]{ .745,  .796,  .894}.033 \\
    sacrum & .875 & .860 & \textbf{.962} & .889 & .932 & \textit{.855} & \cellcolor[rgb]{ .741,  .796,  .894}.895 & \cellcolor[rgb]{ .765,  .808,  .886}.043 \\
    scapula left & .891 & .869 & .869 & \textbf{.909} & .892 & \textit{.850} & \cellcolor[rgb]{ .749,  .8,  .89}.880 & \cellcolor[rgb]{ .722,  .784,  .902}.021 \\
    scapula right & \textbf{.936} & .875 & .907 & .896 & .895 & \textit{.823} & \cellcolor[rgb]{ .745,  .8,  .89}.889 & \cellcolor[rgb]{ .753,  .804,  .89}.038 \\
    small bowel & \textbf{.905} & \textit{.818} & .889 & .881 & .847 & .895 & \cellcolor[rgb]{ .749,  .804,  .89}.872 & \cellcolor[rgb]{ .745,  .8,  .894}.033 \\
    spleen & \textbf{.864} & .742 & .736 & .735 & .815 & \textit{.710} & \cellcolor[rgb]{ .8,  .831,  .871}.767 & \cellcolor[rgb]{ .796,  .827,  .875}.059 \\
    stomach & \textbf{.876} & \textit{.780} & .827 & .815 & .833 & .807 & \cellcolor[rgb]{ .773,  .816,  .878}.823 & \cellcolor[rgb]{ .741,  .796,  .894}.032 \\
    trachea & .907 & .874 & \textbf{.915} & .912 & \textit{.797} & .805 & \cellcolor[rgb]{ .753,  .804,  .886}.868 & \cellcolor[rgb]{ .788,  .824,  .878}.055 \\
    urinary bladder & .752 & .709 & \textbf{.804} & .776 & \textit{.670} & .685 & \cellcolor[rgb]{ .816,  .843,  .867}.733 & \cellcolor[rgb]{ .784,  .82,  .878}.053 \\
    vertebrae C1 & .581 & .582 & \textbf{.621} & .509 & \textit{.502} & .564 & \cellcolor[rgb]{ .894,  .886,  .835}.560 & \cellcolor[rgb]{ .769,  .812,  .882}.046 \\
    vertebrae C2 & .713 & .719 & \textbf{.828} & .595 & .656 & \textit{.576} & \cellcolor[rgb]{ .839,  .855,  .855}.681 & \cellcolor[rgb]{ .863,  .867,  .851}.093 \\
    vertebrae C3 & \textbf{.670} & .659 & .597 & .524 & .433 & \textit{.243} & \cellcolor[rgb]{ .914,  .898,  .831}.521 & \cellcolor[rgb]{ 1,  .949,  .8}\textit{.163} \\
    vertebrae C4 & .435 & .350 & .475 & \textbf{.497} & \textit{.267} & .276 & \cellcolor[rgb]{ .976,  .937,  .808}.383 & \cellcolor[rgb]{ .875,  .875,  .847}.100 \\
    vertebrae C5 & .559 & .408 & .613 & \textbf{.631} & .330 & \textit{.286} & \cellcolor[rgb]{ .937,  .914,  .824}.471 & \cellcolor[rgb]{ .973,  .933,  .812}.149 \\
    vertebrae C6 & .525 & .495 & .504 & .360 & \textit{.183} & \textbf{.555} & \cellcolor[rgb]{ .953,  .922,  .816}.437 & \cellcolor[rgb]{ .957,  .922,  .816}.141 \\
    vertebrae C7 & \textbf{.719} & .679 & .704 & .618 & \textit{.382} & .623 & \cellcolor[rgb]{ .867,  .871,  .847}.621 & \cellcolor[rgb]{ .922,  .902,  .827}.124 \\
    vertebrae L1 & .645 & .579 & .613 & \textbf{.655} & \textit{.536} & .582 & \cellcolor[rgb]{ .875,  .878,  .843}.602 & \cellcolor[rgb]{ .769,  .812,  .886}.045 \\
    vertebrae L2 & .542 & .599 & .659 & .579 & \textit{.535} & \textbf{.700} & \cellcolor[rgb]{ .875,  .878,  .843}.602 & \cellcolor[rgb]{ .808,  .835,  .871}.066 \\
    vertebrae L3 & \textbf{.720} & \textit{.494} & .579 & .593 & .595 & .655 & \cellcolor[rgb]{ .875,  .875,  .843}.606 & \cellcolor[rgb]{ .827,  .847,  .863}.076 \\
    vertebrae L4 & .638 & \textit{.381} & .619 & \textbf{.721} & .637 & .642 & \cellcolor[rgb]{ .875,  .875,  .843}.606 & \cellcolor[rgb]{ .906,  .894,  .835}.116 \\
    vertebrae L5 & \textbf{.782} & \textit{.651} & .757 & .773 & .722 & .698 & \cellcolor[rgb]{ .816,  .843,  .867}.731 & \cellcolor[rgb]{ .776,  .82,  .882}.050 \\
    vertebrae T1 & .741 & \textbf{.745} & \textit{.634} & .700 & .663 & .731 & \cellcolor[rgb]{ .827,  .851,  .859}.702 & \cellcolor[rgb]{ .769,  .812,  .882}.046 \\
    vertebrae T10 & .553 & .545 & .613 & \textbf{.649} & \textit{.489} & .556 & \cellcolor[rgb]{ .89,  .886,  .839}.568 & \cellcolor[rgb]{ .788,  .824,  .878}.056 \\
    vertebrae T11 & \textbf{.580} & .528 & .538 & .470 & \textit{.447} & .498 & \cellcolor[rgb]{ .918,  .902,  .827}.510 & \cellcolor[rgb]{ .776,  .816,  .882}.049 \\
    vertebrae T12 & .564 & .553 & .624 & \textbf{.638} & \textit{.490} & .538 & \cellcolor[rgb]{ .89,  .886,  .839}.568 & \cellcolor[rgb]{ .788,  .824,  .878}.055 \\
    vertebrae T2 & \textbf{.814} & .724 & .758 & .742 & \textit{.669} & .728 & \cellcolor[rgb]{ .812,  .839,  .867}.739 & \cellcolor[rgb]{ .773,  .816,  .882}.047 \\
    vertebrae T3 & \textbf{.794} & .732 & .792 & .740 & .683 & \textit{.598} & \cellcolor[rgb]{ .82,  .843,  .863}.723 & \cellcolor[rgb]{ .824,  .847,  .863}.074 \\
    vertebrae T4 & \textbf{.739} & .728 & .700 & .713 & \textit{.561} & .622 & \cellcolor[rgb]{ .839,  .855,  .855}.677 & \cellcolor[rgb]{ .82,  .843,  .867}.070 \\
    vertebrae T5 & .684 & .687 & \textbf{.775} & .700 & .536 & \textit{.497} & \cellcolor[rgb]{ .855,  .863,  .851}.647 & \cellcolor[rgb]{ .89,  .882,  .843}.107 \\
    vertebrae T6 & \textbf{.674} & .596 & .595 & .585 & .536 & \textit{.525} & \cellcolor[rgb]{ .882,  .882,  .839}.585 & \cellcolor[rgb]{ .784,  .82,  .878}.053 \\
    vertebrae T7 & .706 & \textbf{.743} & .686 & .583 & \textit{.507} & .543 & \cellcolor[rgb]{ .863,  .871,  .847}.628 & \cellcolor[rgb]{ .871,  .871,  .847}.097 \\
    vertebrae T8 & .572 & .538 & \textbf{.756} & .588 & \textit{.432} & .438 & \cellcolor[rgb]{ .898,  .89,  .835}.554 & \cellcolor[rgb]{ .914,  .898,  .831}.119 \\
    vertebrae T9 & .633 & .589 & \textbf{.755} & .550 & .578 & \textit{.522} & \cellcolor[rgb]{ .875,  .875,  .843}.605 & \cellcolor[rgb]{ .843,  .855,  .859}.083 \\ \hline
    Average & .793 & .763 & \textbf{.803} & .773 & \textit{.722} & .736 &       &  \\
    STD   & .140 & .148 & \textbf{.130} & .143 & \textit{.178} & .162 &       &  \\
    \label{tab:localization-ratio-104-regions}
    \end{longtable}
\endgroup

\subsection{Re-ranking}

This section presents the retrieval recalls after applying the re-ranking method of \Cref{sec: re:ranking}. 

\subsubsection{Volume-based}

\Cref{tab:volumewise-recall-29-regions-reranking} and \Cref{tab:volumewise-recall-104-regions-reranking} show the retrieval recalls for 29 coarse anatomical structures and 104 original TS anatomical structures using the proposed re-ranking method. 
All the recalls are improved using re-ranking. 
The performance of the models for 29 classes is close with only slight differences. DINOv1 and DreamSim have a slightly better recall in comparison, with an average recall of $ .967$ but the standard deviation of DINOv1 is slightly lower ($ .040$ vs. $ .045$).
In 104 anatomical regions, SwinTransformer performs better than the other models with an average recall of $.924$ but its standard deviation ($.072$) is the lowest. 

\begin{table}[htbp]
  \centering
  \renewcommand*{\arraystretch}{ .50} 
  \caption{Volume-based retrieval recall of coarse anatomical regions (29 classes) using HNSW Indexing and re-ranking. In each row, bold numbers represent the best-performing values, while italicized numbers indicate the worst-performing. The separate average and standard deviation (STD) columns are color-coded, with blue indicating the best-performing values and yellow indicating the worst-performing values across different models. Additionally, bold numbers in colored columns represent the best classes in terms of average and standard deviation, while italicized values represent the worst-performing class across the models.}
  \scriptsize
\begin{tabular}{lrrrrrr|rr}
    Model & DINOv1 & DINOv2 & DreamSim & SwinTrans. & \multicolumn{2}{c|}{ResNet50} & & \\ \cline{1-7}
      Dataset (pre-trained) & (ImgNet) & (ImgNet) & (ImgNet)& (RadImg)& 
    (Fractaldb) & (RadImg) &Average & STD \\ \hline
    adrenal gland & \textbf{1.000} & \textbf{1.000} &  .980 & \textit{ .960} & \textbf{1.000} & \textit{ .960} & \cellcolor[rgb]{ .729,  .792,  .894} .983 & \cellcolor[rgb]{ .745,  .8,  .894} .020 \\
    autochthon &  .985 & \textit{ .969} &  .985 &  .985 & \textbf{1.000} & \textbf{1.000} & \cellcolor[rgb]{ .725,  .788,  .898} .987 & \cellcolor[rgb]{ .722,  .784,  .902} .012 \\
    brain & \textbf{ .923} & \textbf{ .923} &  .846 & \textbf{ .923} & \textit{ .692} &  .846 & \cellcolor[rgb]{ .929,  .91,  .824} .859 & \cellcolor[rgb]{ .961,  .925,  .816} .090 \\
    cardiovascular system & \textbf{1.000} & \textbf{1.000} & \textit{ .984} & \textbf{1.000} & \textbf{1.000} & \textbf{1.000} & \cellcolor[rgb]{ .706,  .776,  .906} .997 & \cellcolor[rgb]{ .706,  .776,  .906} .006 \\
    clavicula & \textit{ .974} & \textit{ .974} & \textit{ .974} & \textit{ .974} & \textbf{1.000} & \textbf{1.000} & \cellcolor[rgb]{ .729,  .792,  .894} .983 & \cellcolor[rgb]{ .725,  .788,  .902} .013 \\
    colon & \textbf{1.000} &  .981 & \textbf{1.000} & \textit{ .962} & \textit{ .962} &  .981 & \cellcolor[rgb]{ .733,  .792,  .894} .981 & \cellcolor[rgb]{ .737,  .792,  .898} .017 \\
\end{tabular}%
\end{table}%

\FloatBarrier

\begingroup
\renewcommand*{\arraystretch}{ .50} 
\scriptsize
\begin{longtable}{lcccccc|cc}
\caption{Volume-based retrieval recall of all TS anatomical regions (104 classes) using HNSW Indexing and re-ranking. In each row, bold numbers represent the best-performing values, while italicized numbers indicate the worst-performing. The separate average and standard deviation (STD) columns are color-coded, with blue indicating the best-performing values and yellow indicating the worst-performing values across different models. Additionally, bold numbers in colored columns represent the best classes in terms of average and standard deviation, while italicized values represent the worst-performing class across the models.}  \\
Model & DINOv1 & DINOv2 & DreamSim & SwinTrans. & \multicolumn{2}{c|}{ResNet50} & & \\ \cline{1-7}
      Dataset (pre-trained) & (ImgNet) & (ImgNet) & (ImgNet)& (RadImg)& 
    (Fractaldb) & (RadImg) & Average & STD \\ \hline
    adrenal gland left &  .920 & \textit{ .880} & \textit{ .880} & \textbf{ .940} &  .920 & \textit{ .880} & \cellcolor[rgb]{ .765,  .812,  .882} .903 & \cellcolor[rgb]{ .753,  .804,  .89} .027 \\
    adrenal gland right & \textbf{ .940} &  .920 & \textit{ .880} & \textbf{ .940} &  .920 &  .900 & \cellcolor[rgb]{ .753,  .804,  .886} .917 & \cellcolor[rgb]{ .745,  .796,  .894} .023 \\
    aorta & \textbf{ .951} &  .902 & \textit{ .885} & \textbf{ .951} & \textit{ .885} &  .918 & \cellcolor[rgb]{ .753,  .804,  .886} .915 & \cellcolor[rgb]{ .761,  .808,  .886} .030 \\
    autochthon left & \textbf{ .938} & \textit{ .877} &  .892 & \textbf{ .938} &  .908 &  .923 & \cellcolor[rgb]{ .757,  .808,  .886} .913 & \cellcolor[rgb]{ .749,  .8,  .89} .025 \\
    autochthon right & \textbf{ .938} & \textit{ .877} &  .892 & \textbf{ .938} &  .908 &  .923 & \cellcolor[rgb]{ .757,  .808,  .886} .913 & \cellcolor[rgb]{ .749,  .8,  .89} .025 \\
\end{longtable}
\endgroup

\subsubsection{Region-based}

\Cref{tab:regionbased-recall-29-regions-reranking} and \Cref{tab:regionwise-recall-104-regions-reranking} show the retrieval recall for 29 coarse anatomical structures and 104 original TS anatomical structures employing the proposed re-ranking method. 
Using the re-ranking, the overall performance of all the models has improved. 
DreamSim performs the best with the average retrieval recall of $ .987\pm .027$ and $ .987 \pm  .024$ for 29 and 104 classes, respectively. 
There are only slight variations between the performance on coarse and all the original TS classes.  
Similar to the count-based method in the anatomical region retrieval many classes are perfectly retrieved (recall of 1.0).
There is a low variation among models and between classes where the highest standard deviation is $ .064$ and $ .042$.

\begin{table}[!htpb]
 \centering
 \renewcommand*{\arraystretch}{ .50} 
  \caption{Region-based retrieval recall of coarse anatomical regions (29 classes) using HNSW Indexing and re-ranking. In each row, bold numbers represent the best-performing values, while italicized numbers indicate the worst-performing. The separate average and standard deviation (STD) columns are color-coded, with blue indicating the best-performing values and yellow indicating the worst-performing values across different models. Additionally, bold numbers in colored columns represent the best classes in terms of average and standard deviation, while italicized values represent the worst-performing class across the models.}
  \scriptsize
\begin{tabular}{lrrrrrr|rr}
    Model & DINOv1 & DINOv2 & DreamSim & SwinTrans. & \multicolumn{2}{c|}{ResNet50} & & \\ \cline{1-7}
      Dataset (pre-trained) & (ImgNet) & (ImgNet) & (ImgNet)& (RadImg)& 
    (Fractaldb) & (RadImg) &Average & STD \\ \hline
    adrenal gland & \textbf{1.000} & \textbf{1.000} & \textbf{1.000} & \textit{ .970} &  .990 & \textbf{1.000} & \cellcolor[rgb]{ .722,  .784,  .898} .993 & \cellcolor[rgb]{ .761,  .808,  .886} .012 \\
    autochthon &  .992 & \textbf{1.000} &  .992 & \textit{ .985} & \textbf{1.000} &  .992 & \cellcolor[rgb]{ .722,  .784,  .898} .994 & \cellcolor[rgb]{ .729,  .792,  .898} .006 \\
    brain &  .923 & \textit{ .846} & \textbf{1.000} &  .923 & \textbf{1.000} & \textbf{1.000} & \cellcolor[rgb]{ .804,  .835,  .871} .949 & \cellcolor[rgb]{ 1,  .949,  .8}\textit{ .063} \\
    cardiovascular system & \textbf{1.000} & \textbf{1.000} & \textbf{1.000} & \textbf{1.000} & \textbf{1.000} & \textbf{1.000} & \cellcolor[rgb]{ .706,  .776,  .906}1.000 & \cellcolor[rgb]{ .706,  .776,  .906} \textbf{.000} \\
    clavicula &  .987 &  .987 & \textbf{1.000} & \textbf{1.000} & \textit{ .974} &  .987 & \cellcolor[rgb]{ .729,  .792,  .898} .989 & \cellcolor[rgb]{ .749,  .8,  .89} .010 \\
    colon & \textbf{1.000} & \textbf{1.000} & \textbf{1.000} & \textit{ .981} & \textit{ .981} & \textit{ .981} & \cellcolor[rgb]{ .725,  .788,  .898} .991 & \cellcolor[rgb]{ .753,  .804,  .89} .010 \\
\end{tabular}%
\end{table}%

\FloatBarrier

\begingroup
\renewcommand*{\arraystretch}{ .5} 
\scriptsize
\begin{longtable}{lcccccc|cc}
\caption{Region-based retrieval recall of all TS anatomical regions (104 classes) using HNSW Indexing and re-ranking. In each row, bold numbers represent the best-performing values, while italicized numbers indicate the worst-performing. The separate average and standard deviation (STD) columns are color-coded, with blue indicating the best-performing values and yellow indicating the worst-performing values across different models. Additionally, bold numbers in colored columns represent the best classes in terms of average and standard deviation, while italicized values represent the worst-performing class across the models.}   \\
Model & DINOv1 & DINOv2 & DreamSim & SwinTrans. & \multicolumn{2}{c|}{ResNet50} & & \\ \cline{1-7}
      Dataset (pre-trained) & (ImgNet) & (ImgNet) & (ImgNet)& (RadImg)& 
    (Fractaldb) & (RadImg) & Average & STD \\ \hline
    adrenal gland left &  .980 & \textbf{1.000} & \textbf{1.000} & \textit{ .940} & \textbf{1.000} & \textbf{1.000} & \cellcolor[rgb]{ .733,  .792,  .894} .987 & \cellcolor[rgb]{ .816,  .839,  .867} .024 \\
    adrenal gland right & \textbf{1.000} & \textbf{1.000} & \textbf{1.000} &  .980 & \textit{ .960} & \textbf{1.000} & \cellcolor[rgb]{ .725,  .788,  .898} .990 & \cellcolor[rgb]{ .78,  .82,  .878} .017 \\
    aorta & \textbf{1.000} & \textbf{1.000} & \textbf{1.000} & \textbf{1.000} & \textbf{1.000} & \textbf{1.000} & \cellcolor[rgb]{ .706,  .776,  .906}\textbf{1.000} & \cellcolor[rgb]{ .706,  .776,  .906}\textbf{ .000} \\
    autochthon left & \textbf{1.000} & \textbf{1.000} & \textbf{1.000} & \textit{ .985} & \textbf{1.000} & \textit{ .985} & \cellcolor[rgb]{ .718,  .784,  .902} .995 & \cellcolor[rgb]{ .741,  .796,  .894} .008 \\
    autochthon right & \textit{ .985} & \textbf{1.000} & \textit{ .985} & \textit{ .985} & \textbf{1.000} & \textbf{1.000} & \cellcolor[rgb]{ .722,  .788,  .898} .992 & \cellcolor[rgb]{ .741,  .796,  .894} .008 \\
\end{longtable}
\endgroup

\subsubsection{Localized}

\paragraph{Localized Retrieval Recall}
\Cref{tab:localized-recall-29-regions-reranking} and \Cref{tab:localized-recall-104-regions-reranking} show the retrieval recall for 29 coarse anatomical structures and 104 original TS anatomical structures after re-ranking for $L=15$. 
Re-ranking improved the localization for all the models.
DreamsSim is the best-performing model with an average recall of $.955 \pm .062$ for coarse anatomical structures and  $.956 \pm .055$ for original TS classes. 
Although the retrieval is lower compared to region-based or volume-based evaluation, it is still high which shows that the pretrained vision embeddings not only can retrieve similar cases but also can localize the corresponding region of interest.

\begin{table}[!htpb]
 \centering
 \renewcommand*{\arraystretch}{ .50} 
  \caption{Localized retrieval recall of coarse anatomical regions (29 classes) using HNSW Indexing and re-ranking, $L=15$. In each row, bold numbers represent the best-performing values, while italicized numbers indicate the worst-performing. The separate average and standard deviation (STD) columns are color-coded, with blue indicating the best-performing values and yellow indicating the worst-performing values across different models. Additionally, bold numbers in colored columns represent the best classes in terms of average and standard deviation, while italicized values represent the worst-performing class across the models.}
  \scriptsize
\begin{tabular}{lrrrrrr|rr}
    Model & DINOv1 & DINOv2 & DreamSim & SwinTrans. & \multicolumn{2}{c|}{ResNet50} & & \\ \cline{1-7}
      Dataset (pre-trained) & (ImgNet) & (ImgNet) & (ImgNet)& (RadImg)& 
    (Fractaldb) & (RadImg) &Average & STD \\ \hline
    adrenal gland & \textbf{.980} & \textbf{.980} & \textbf{.980} & .940 & \textit{.860} & .930 & \cellcolor[rgb]{ .749,  .804,  .89}.945 & \cellcolor[rgb]{ .863,  .867,  .851}.047 \\
    autochthon & .992 & .985 & .985 & .985 & \textbf{1.000} & \textit{.977} & \cellcolor[rgb]{ .714,  .784,  .902}.987 & \cellcolor[rgb]{ .714,  .78,  .906}.008 \\
    brain & .923 & \textit{.846} & \textbf{1.000} & .923 & .923 & \textbf{1.000} & \cellcolor[rgb]{ .757,  .808,  .886}.936 & \cellcolor[rgb]{ .902,  .89,  .835}.058 \\
    cardiovascular system & \textbf{1.000} & .997 & .998 & .998 & \textit{.986} & .993 & \cellcolor[rgb]{ .71,  .78,  .902}.995 & \cellcolor[rgb]{ .706,  .776,  .906}\textbf{.005} \\
    clavicula & .961 & \textbf{.987} & \textbf{.987} & .961 & \textit{.909} & .961 & \cellcolor[rgb]{ .737,  .796,  .894}.961 & \cellcolor[rgb]{ .792,  .827,  .875}.028 \\
    colon & .962 & \textbf{1.000} & \textbf{1.000} & .981 & \textit{.925} & \textit{.925} & \cellcolor[rgb]{ .733,  .792,  .894}.965 & \cellcolor[rgb]{ .816,  .839,  .867}.035 \\
\end{tabular}%
\end{table}%

\FloatBarrier

\begingroup
\renewcommand*{\arraystretch}{ .5} 
\scriptsize
\begin{longtable}{lcccccc|cc}
\caption{Localized retrieval recall of all TS anatomical regions (104 classes) using HNSW Indexing and re-ranking, $L=15$. In each row, bold numbers represent the best-performing values, while italicized numbers indicate the worst-performing. The separate average and standard deviation (STD) columns are color-coded, with blue indicating the best-performing values and yellow indicating the worst-performing values across different models. Additionally, bold numbers in colored columns represent the best classes in terms of average and standard deviation, while italicized values represent the worst-performing class across the models.}   \\
Model & DINOv1 & DINOv2 & DreamSim & SwinTrans. & \multicolumn{2}{c|}{ResNet50} & & \\ \cline{1-7}
      Dataset (pre-trained) & (ImgNet) & (ImgNet) & (ImgNet)& (RadImg)& 
    (Fractaldb) & (RadImg) & Average & STD \\ \hline
    adrenal gland left & .960 & \textbf{.980} & .940 & .860 & \textit{.840} & .860 & \cellcolor[rgb]{ .784,  .824,  .875}.907 & \cellcolor[rgb]{ .82,  .843,  .867}.060 \\
    adrenal gland right & \textbf{.940} & \textbf{.940} & \textbf{.940} & .900 & \textit{.820} & \textbf{.940} & \cellcolor[rgb]{ .78,  .82,  .878}.913 & \cellcolor[rgb]{ .796,  .827,  .875}.048 \\
    aorta & \textbf{1.000} & \textbf{1.000} & \textbf{1.000} & \textbf{1.000} & \textbf{1.000} & \textit{.984} & \cellcolor[rgb]{ .71,  .78,  .902}.997 & \cellcolor[rgb]{ .718,  .78,  .902}.007 \\
    autochthon left & \textbf{1.000} & .985 & \textbf{1.000} & .985 & \textbf{1.000} & \textit{.954} & \cellcolor[rgb]{ .718,  .784,  .902}.987 & \cellcolor[rgb]{ .737,  .796,  .894}.018 \\
    autochthon right & .985 & .985 & \textit{.969} & .985 & \textbf{1.000} & .985 & \cellcolor[rgb]{ .722,  .784,  .898}.985 & \cellcolor[rgb]{ .722,  .784,  .902}.010 \\
\end{longtable}
\endgroup

\paragraph{Localization-ratio}

The localization-ratio is calculated based on \eqref{eq:normalized-location-ratio-rerank-version}. 
\Cref{tab:localiation-ratio-29-regions-reranking} and \Cref{tab:localization-ratio-104-regions-reranking} demonstrate the localization-ratio for 29 coarse and 104 original TS anatomical regions after re-ranking for $L=15$.
The best-performing embedding is still DreamSim with a localization-ratio of $.837\pm.159$ and $.790\pm.142$ for 29 coarse and 104 original TS classes. 
After re-ranking the overall localization-ratio is reduced (previously, $.864\pm.0145$ and $.803\pm.130$, respectively).

\begin{table}[!htpb]
 \centering
 \renewcommand*{\arraystretch}{ .50} 
  \caption{Localization-ratio of coarse anatomical regions (29 classes) using HNSW Indexing and re-ranking, $L=15$. In each row, bold numbers represent the best-performing values, while italicized numbers indicate the worst-performing. The separate average and standard deviation (STD) columns are color-coded, with blue indicating the best-performing values and yellow indicating the worst-performing values across different models. Additionally, bold numbers in colored columns represent the best classes in terms of average and standard deviation, while italicized values represent the worst-performing class across the models.}
  \scriptsize
    \begin{tabular}{lrrrrrr|rr}
    Model & DINOv1 & DINOv2 & DreamSim & SwinTrans. & \multicolumn{2}{c|}{ResNet50} & & \\ \cline{1-7}
      Dataset (pre-trained) & (ImgNet) & (ImgNet) & (ImgNet)& (RadImg)& 
    (Fractaldb) & (RadImg) &Average & STD \\ \hline    \multicolumn{1}{l}{adrenal gland} & .697 & .607 & \textbf{.712} & .537 & \textit{.523} & \textit{.523} & \cellcolor[rgb]{ .878,  .878,  .843}.600 & \cellcolor[rgb]{ .91,  .898,  .831}.087 \\
    \multicolumn{1}{l}{autochthon} & \textbf{.970} & .956 & .951 & .959 & \textit{.926} & .939 & \cellcolor[rgb]{ .725,  .788,  .898}.950 & \cellcolor[rgb]{ .71,  .78,  .906}.016 \\
    \multicolumn{1}{l}{brain} & .903 & .759 & \textbf{.990} & .846 & \textit{.651} & .785 & \cellcolor[rgb]{ .78,  .82,  .878}.822 & \cellcolor[rgb]{ 1,  .949,  .8}\textit{.118} \\
    \multicolumn{1}{l}{cardiovascular system} & \textit{.945} & .977 & \textbf{.989} & .968 & .983 & .962 & \cellcolor[rgb]{ .718,  .784,  .902}.971 & \cellcolor[rgb]{ .714,  .78,  .906}.016 \\
    \multicolumn{1}{l}{clavicula} & .819 & .805 & .841 & .764 & \textit{.619} & \textbf{.848} & \cellcolor[rgb]{ .8,  .831,  .871}.783 & \cellcolor[rgb]{ .906,  .894,  .835}.086 \\
    \multicolumn{1}{l}{colon} & .874 & .886 & \textbf{.952} & \textit{.808} & .891 & .843 & \cellcolor[rgb]{ .757,  .808,  .886}.875 & \cellcolor[rgb]{ .804,  .831,  .871}.049 \\
    \multicolumn{1}{l}{duodenum} & .645 & .627 & .636 & .656 & \textit{.517} & \textbf{.697} & \cellcolor[rgb]{ .863,  .871,  .847}.630 & \cellcolor[rgb]{ .835,  .851,  .859}.060 \\
    \multicolumn{1}{l}{esophagus} & .920 & .956 & \textbf{.957} & .910 & \textit{.887} & .920 & \cellcolor[rgb]{ .737,  .796,  .894}.925 & \cellcolor[rgb]{ .745,  .796,  .894}.027 \\
    \multicolumn{1}{l}{face} & .682 & .702 & .569 & \textbf{.729} & \textit{.561} & .576 & \cellcolor[rgb]{ .863,  .867,  .847}.637 & \cellcolor[rgb]{ .882,  .878,  .843}.076 \\
    \multicolumn{1}{l}{femur} & .896 & .881 & \textbf{.954} & .921 & .861 & \textit{.847} & \cellcolor[rgb]{ .749,  .804,  .89}.894 & \cellcolor[rgb]{ .776,  .82,  .882}.040 \\
    \multicolumn{1}{l}{gallbladder} & \textit{.224} & \textbf{.376} & .356 & .268 & .321 & .320 & \cellcolor[rgb]{ 1,  .949,  .8}\textit{.311} & \cellcolor[rgb]{ .824,  .847,  .863}.056 \\
    \multicolumn{1}{l}{gluteus muscles} & \textbf{1.000} & .992 & \textbf{1.000} & .977 & \textit{.960} & .978 & \cellcolor[rgb]{ .71,  .78,  .902}.984 & \cellcolor[rgb]{ .71,  .78,  .906}.016 \\
    \multicolumn{1}{l}{hip} & .968 & .929 & \textbf{.970} & .924 & \textit{.795} & .932 & \cellcolor[rgb]{ .737,  .796,  .894}.920 & \cellcolor[rgb]{ .847,  .859,  .855}.064 \\
    \multicolumn{1}{l}{humerus} & .529 & .554 & \textbf{.638} & .550 & \textit{.417} & .525 & \cellcolor[rgb]{ .906,  .894,  .831}.535 & \cellcolor[rgb]{ .867,  .871,  .851}.071 \\
    \multicolumn{1}{l}{iliopsoas} & .926 & .883 & \textbf{.942} & \textit{.869} & .926 & .887 & \cellcolor[rgb]{ .745,  .8,  .89}.905 & \cellcolor[rgb]{ .749,  .8,  .89}.029 \\
    \multicolumn{1}{l}{kidney} & .770 & .793 & .819 & .826 & \textit{.709} & \textbf{.831} & \cellcolor[rgb]{ .796,  .827,  .875}.791 & \cellcolor[rgb]{ .796,  .827,  .875}.047 \\
    \multicolumn{1}{l}{liver} & \textbf{.804} & .788 & .742 & .737 & \textit{.604} & .770 & \cellcolor[rgb]{ .816,  .843,  .867}.741 & \cellcolor[rgb]{ .867,  .871,  .847}.072 \\
    \multicolumn{1}{l}{lung} & .970 & .948 & \textbf{.980} & .929 & \textit{.876} & .906 & \cellcolor[rgb]{ .733,  .792,  .894}.935 & \cellcolor[rgb]{ .776,  .82,  .882}.039 \\
    \multicolumn{1}{l}{pancreas} & .772 & .721 & \textbf{.773} & .720 & \textit{.505} & .651 & \cellcolor[rgb]{ .839,  .855,  .859}.690 & \cellcolor[rgb]{ .949,  .918,  .82}.101 \\
    \multicolumn{1}{l}{portal and splenic vein} & \textbf{.664} & .613 & .643 & .560 & \textit{.449} & .525 & \cellcolor[rgb]{ .886,  .882,  .839}.576 & \cellcolor[rgb]{ .894,  .886,  .839}.081 \\
    \multicolumn{1}{l}{rib} & \textbf{1.000} & .982 & .992 & .970 & \textit{.918} & .992 & \cellcolor[rgb]{ .714,  .784,  .902}.976 & \cellcolor[rgb]{ .753,  .804,  .89}.030 \\
    \multicolumn{1}{l}{sacrum} & .862 & .844 & \textbf{.892} & .882 & \textit{.830} & .883 & \cellcolor[rgb]{ .761,  .812,  .882}.866 & \cellcolor[rgb]{ .737,  .792,  .898}.025 \\
    \multicolumn{1}{l}{scapula} & \textbf{.940} & .913 & .904 & \textit{.806} & .816 & .912 & \cellcolor[rgb]{ .757,  .808,  .886}.882 & \cellcolor[rgb]{ .824,  .847,  .863}.056 \\
    \multicolumn{1}{l}{small bowel} & .925 & \textit{.849} & .860 & .885 & \textbf{.949} & .918 & \cellcolor[rgb]{ .749,  .804,  .89}.897 & \cellcolor[rgb]{ .776,  .82,  .882}.039 \\
    \multicolumn{1}{l}{spleen} & \textbf{.775} & .757 & .708 & .715 & \textit{.579} & .669 & \cellcolor[rgb]{ .835,  .851,  .859}.700 & \cellcolor[rgb]{ .863,  .871,  .851}.070 \\
    \multicolumn{1}{l}{stomach} & \textbf{.816} & .689 & .784 & .719 & \textit{.557} & .686 & \cellcolor[rgb]{ .831,  .851,  .859}.708 & \cellcolor[rgb]{ .922,  .902,  .831}.091 \\
    \multicolumn{1}{l}{trachea} & .881 & \textbf{.899} & .862 & \textit{.767} & .816 & .859 & \cellcolor[rgb]{ .769,  .816,  .882}.847 & \cellcolor[rgb]{ .8,  .831,  .875}.048 \\
    \multicolumn{1}{l}{urinary bladder} & .702 & \textit{.595} & \textbf{.845} & .780 & .662 & .668 & \cellcolor[rgb]{ .831,  .851,  .859}.709 & \cellcolor[rgb]{ .918,  .902,  .831}.090 \\
    \multicolumn{1}{l}{vertebrae} & \textbf{1.000} & \textbf{1.000} & \textbf{1.000} & \textbf{1.000} & .981 & \textit{.970} & \cellcolor[rgb]{ .706,  .776,  .906}\textbf{.992} & \cellcolor[rgb]{ .706,  .776,  .906}\textbf{.013} \\ \hline
      Average    & .823 & .803 & \textbf{.837} & .792 & \textit{.727} & .787 &       &  \\
     STD     & .169 & \textbf{.157} & \textbf{.159} & .164 & \textit{.195} & .171 &       &  \\
        \label{tab:localiation-ratio-29-regions-reranking}
    \end{tabular}%
\end{table}%

\FloatBarrier

\begingroup
\renewcommand*{\arraystretch}{ .5} 
\scriptsize
\begin{longtable}{lcccccc|cc}
\caption{Localization-ratio of all TS anatomical regions (104 classes) using HNSW Indexing and re-ranking, $L=15$. In each row, bold numbers represent the best-performing values, while italicized numbers indicate the worst-performing. The separate average and standard deviation (STD) columns are color-coded, with blue indicating the best-performing values and yellow indicating the worst-performing values across different models. Additionally, bold numbers in colored columns represent the best classes in terms of average and standard deviation, while italicized values represent the worst-performing class across the models.}   \\
Model & DINOv1 & DINOv2 & DreamSim & SwinTrans. & \multicolumn{2}{c|}{ResNet50} & & \\ \cline{1-7}
      Dataset (pre-trained) & (ImgNet) & (ImgNet) & (ImgNet)& (RadImg)& 
    (Fractaldb) & (RadImg) & Average & STD \\ \hline
    adrenal gland left & \textbf{.673} & .573 & .600 & .507 & .480 & \textit{.387} & \cellcolor[rgb]{ .898,  .89,  .835}.537 & \cellcolor[rgb]{ .886,  .882,  .843}.101 \\
    adrenal gland right & \textbf{.647} & .467 & .580 & .547 & \textit{.300} & .473 & \cellcolor[rgb]{ .914,  .898,  .831}.502 & \cellcolor[rgb]{ .929,  .906,  .827}.120 \\
    aorta & \textbf{.940} & .929 & .923 & .902 & .874 & \textit{.814} & \cellcolor[rgb]{ .729,  .792,  .898}.897 & \cellcolor[rgb]{ .776,  .816,  .882}.047 \\
    autochthon left & \textbf{.985} & .959 & .964 & .954 & .903 & \textit{.897} & \cellcolor[rgb]{ .706,  .776,  .906}\textbf{.944} & \cellcolor[rgb]{ .753,  .804,  .89}.035 \\
    autochthon right & .933 & \textbf{.964} & .928 & .954 & \textit{.908} & \textit{.908} & \cellcolor[rgb]{ .714,  .78,  .902}.932 & \cellcolor[rgb]{ .729,  .788,  .898}.023 \\
    brain & .923 & .769 & \textbf{1.000} & .897 & \textit{.667} & .821 & \cellcolor[rgb]{ .753,  .804,  .886}.846 & \cellcolor[rgb]{ .925,  .906,  .827}.119 \\
    clavicula left & .744 & .803 & \textbf{.838} & .752 & \textit{.658} & .812 & \cellcolor[rgb]{ .788,  .827,  .875}.768 & \cellcolor[rgb]{ .816,  .839,  .867}.065 \\
    clavicula right & .816 & .825 & \textbf{.868} & .754 & \textit{.667} & .860 & \cellcolor[rgb]{ .776,  .82,  .878}.798 & \cellcolor[rgb]{ .839,  .851,  .859}.076 \\
    colon & .881 & .855 & \textbf{.931} & \textit{.818} & .899 & .855 & \cellcolor[rgb]{ .741,  .796,  .89}.873 & \cellcolor[rgb]{ .761,  .808,  .886}.039 \\
    duodenum & .647 & .700 & .547 & .673 & \textit{.500} & \textbf{.713} & \cellcolor[rgb]{ .855,  .863,  .851}.630 & \cellcolor[rgb]{ .859,  .867,  .851}.087 \\
    esophagus & .929 & \textbf{.982} & .929 & .881 & \textit{.869} & .899 & \cellcolor[rgb]{ .722,  .788,  .898}.915 & \cellcolor[rgb]{ .765,  .812,  .886}.041 \\
    face  & .608 & \textbf{.706} & \textit{.549} & \textbf{.706} & \textit{.549} & .647 & \cellcolor[rgb]{ .855,  .863,  .851}.627 & \cellcolor[rgb]{ .827,  .847,  .863}.071 \\
    femur left & .881 & .896 & \textbf{.956} & .911 & \textit{.837} & .859 & \cellcolor[rgb]{ .733,  .792,  .894}.890 & \cellcolor[rgb]{ .765,  .812,  .886}.041 \\
    femur right & \textbf{.951} & \textbf{.951} & .927 & \textbf{.951} & .919 & \textit{.886} & \cellcolor[rgb]{ .714,  .78,  .902}.931 & \cellcolor[rgb]{ .733,  .792,  .898}.026 \\
    gallbladder & \textit{.162} & \textbf{.453} & .350 & .282 & .282 & .333 & \cellcolor[rgb]{ 1,  .949,  .8}\textit{.311} & \cellcolor[rgb]{ .878,  .878,  .843}.096 \\
    gluteus maximus left & \textbf{.953} & .938 & .946 & .946 & \textit{.922} & \textbf{.953} & \cellcolor[rgb]{ .71,  .78,  .902}.943 & \cellcolor[rgb]{ .706,  .776,  .906}\textbf{.012} \\
    gluteus maximus right & .946 & \textit{.930} & \textit{.930} & .946 & .938 & \textbf{.961} & \cellcolor[rgb]{ .71,  .78,  .902}.942 & \cellcolor[rgb]{ .706,  .776,  .906}.012 \\
    gluteus medius left & \textbf{.947} & .924 & .909 & \textit{.864} & \textit{.864} & .917 & \cellcolor[rgb]{ .725,  .788,  .898}.904 & \cellcolor[rgb]{ .749,  .8,  .89}.034 \\
    gluteus medius right & .899 & .907 & \textbf{.946} & .899 & \textit{.837} & \textbf{.946} & \cellcolor[rgb]{ .725,  .788,  .898}.906 & \cellcolor[rgb]{ .761,  .808,  .886}.040 \\
    gluteus minimus left & .806 & .899 & \textit{.798} & .907 & \textbf{.938} & .922 & \cellcolor[rgb]{ .737,  .796,  .894}.879 & \cellcolor[rgb]{ .804,  .835,  .871}.061 \\
    gluteus minimus right & .865 & .841 & \textit{.802} & .905 & .881 & \textbf{.976} & \cellcolor[rgb]{ .737,  .796,  .894}.878 & \cellcolor[rgb]{ .804,  .831,  .871}.060 \\
    heart atrium left & .645 & .745 & \textbf{.837} & .596 & .667 & \textit{.525} & \cellcolor[rgb]{ .835,  .855,  .859}.669 & \cellcolor[rgb]{ .906,  .894,  .835}.110 \\
    heart atrium right & .721 & .776 & \textbf{.898} & .741 & \textit{.646} & .762 & \cellcolor[rgb]{ .796,  .827,  .875}.757 & \cellcolor[rgb]{ .851,  .859,  .855}.082 \\
    heart myocardium & .748 & .769 & \textbf{.939} & .782 & \textit{.687} & .830 & \cellcolor[rgb]{ .776,  .82,  .878}.793 & \cellcolor[rgb]{ .855,  .863,  .851}.085 \\
    heart ventricle left & .776 & .701 & \textbf{.918} & .741 & \textit{.619} & .816 & \cellcolor[rgb]{ .792,  .827,  .875}.762 & \cellcolor[rgb]{ .89,  .886,  .839}.102 \\
    heart ventricle right & .769 & .830 & \textbf{.912} & .803 & \textit{.687} & .782 & \cellcolor[rgb]{ .776,  .82,  .878}.797 & \cellcolor[rgb]{ .831,  .851,  .863}.074 \\
    hip left & .962 & .932 & \textbf{.977} & .939 & \textit{.727} & .924 & \cellcolor[rgb]{ .722,  .788,  .898}.910 & \cellcolor[rgb]{ .871,  .871,  .847}.092 \\
    hip right & \textbf{.970} & .917 & .962 & .886 & \textit{.826} & .955 & \cellcolor[rgb]{ .718,  .784,  .898}.919 & \cellcolor[rgb]{ .796,  .827,  .875}.056 \\
    humerus left & .530 & .607 & \textbf{.615} & .590 & \textit{.368} & .513 & \cellcolor[rgb]{ .898,  .89,  .835}.537 & \cellcolor[rgb]{ .871,  .875,  .847}.093 \\
    humerus right & .521 & .528 & \textbf{.625} & .472 & \textit{.417} & .549 & \cellcolor[rgb]{ .906,  .894,  .831}.519 & \cellcolor[rgb]{ .827,  .847,  .863}.071 \\
    iliac artery left & .894 & .841 & \textbf{.962} & .902 & \textit{.712} & .894 & \cellcolor[rgb]{ .745,  .8,  .89}.867 & \cellcolor[rgb]{ .855,  .863,  .855}.085 \\
    iliac artery right & .818 & .902 & \textbf{.977} & .902 & \textit{.742} & .894 & \cellcolor[rgb]{ .741,  .796,  .89}.872 & \cellcolor[rgb]{ .847,  .859,  .855}.081 \\
    iliac vena left & .886 & .848 & \textbf{.939} & .848 & \textit{.735} & .917 & \cellcolor[rgb]{ .745,  .8,  .89}.862 & \cellcolor[rgb]{ .827,  .847,  .863}.072 \\
    iliac vena right & .871 & .879 & \textbf{.955} & .871 & \textit{.773} & .841 & \cellcolor[rgb]{ .745,  .8,  .89}.865 & \cellcolor[rgb]{ .8,  .831,  .875}.059 \\
    iliopsoas left & .899 & .881 & \textbf{.937} & .868 & .893 & \textit{.843} & \cellcolor[rgb]{ .733,  .792,  .894}.887 & \cellcolor[rgb]{ .745,  .8,  .894}.032 \\
    iliopsoas right & \textit{.889} & .895 & .895 & .895 & \textbf{.967} & \textit{.889} & \cellcolor[rgb]{ .725,  .788,  .898}.905 & \cellcolor[rgb]{ .741,  .796,  .894}.031 \\
    inferior vena cava & \textbf{.936} & .865 & \textit{.836} & .906 & \textit{.836} & .883 & \cellcolor[rgb]{ .737,  .796,  .894}.877 & \cellcolor[rgb]{ .761,  .808,  .886}.039 \\
    kidney left & .736 & .692 & .736 & \textbf{.748} & \textit{.610} & .711 & \cellcolor[rgb]{ .82,  .843,  .863}.705 & \cellcolor[rgb]{ .784,  .824,  .878}.051 \\
    kidney right & .646 & .735 & \textbf{.782} & .748 & \textit{.633} & .762 & \cellcolor[rgb]{ .812,  .839,  .867}.718 & \cellcolor[rgb]{ .808,  .835,  .871}.063 \\
    liver & .821 & \textbf{.870} & .741 & .698 & \textit{.611} & .815 & \cellcolor[rgb]{ .792,  .827,  .875}.759 & \cellcolor[rgb]{ .878,  .875,  .847}.095 \\
    lung lower lobe left & .860 & .889 & .883 & \textbf{.895} & \textit{.754} & \textit{.754} & \cellcolor[rgb]{ .757,  .808,  .886}.839 & \cellcolor[rgb]{ .82,  .843,  .867}.067 \\
    lung lower lobe right & .875 & .839 & .857 & \textbf{.893} & \textit{.720} & \textit{.720} & \cellcolor[rgb]{ .765,  .812,  .882}.817 & \cellcolor[rgb]{ .839,  .855,  .859}.077 \\
    lung middle lobe right & \textit{.707} & .762 & \textbf{.884} & .796 & .735 & .721 & \cellcolor[rgb]{ .788,  .827,  .875}.768 & \cellcolor[rgb]{ .816,  .839,  .867}.065 \\
    lung upper lobe left & .917 & .887 & \textbf{.946} & .839 & .839 & \textit{.815} & \cellcolor[rgb]{ .741,  .796,  .894}.874 & \cellcolor[rgb]{ .784,  .824,  .878}.051 \\
    lung upper lobe right & .855 & \textbf{.899} & .841 & .775 & .826 & \textit{.754} & \cellcolor[rgb]{ .765,  .812,  .882}.825 & \cellcolor[rgb]{ .788,  .824,  .878}.053 \\
    pancreas & .753 & .780 & \textbf{.787} & .733 & \textit{.527} & .660 & \cellcolor[rgb]{ .82,  .843,  .863}.707 & \cellcolor[rgb]{ .886,  .882,  .843}.099 \\
    portal and splenic vein & \textbf{.733} & .713 & .673 & .573 & .480 & \textit{.473} & \cellcolor[rgb]{ .863,  .871,  .847}.608 & \cellcolor[rgb]{ .918,  .902,  .831}.116 \\
    pulmonary artery & \textbf{.735} & .718 & \textbf{.735} & .692 & .632 & \textit{.530} & \cellcolor[rgb]{ .831,  .851,  .859}.674 & \cellcolor[rgb]{ .847,  .859,  .855}.080 \\
    rib left 1 & .833 & .825 & \textbf{.912} & \textit{.702} & .737 & .904 & \cellcolor[rgb]{ .765,  .812,  .882}.819 & \cellcolor[rgb]{ .859,  .863,  .851}.086 \\
    rib left 10 & \textbf{.850} & .732 & .739 & .778 & \textit{.621} & .712 & \cellcolor[rgb]{ .804,  .835,  .871}.739 & \cellcolor[rgb]{ .835,  .851,  .859}.075 \\
    rib left 11 & .732 & .778 & \textbf{.797} & .693 & \textit{.647} & .758 & \cellcolor[rgb]{ .804,  .835,  .871}.734 & \cellcolor[rgb]{ .796,  .827,  .875}.056 \\
    rib left 12 & .611 & .535 & \textbf{.729} & .549 & \textit{.465} & .528 & \cellcolor[rgb]{ .882,  .882,  .843}.569 & \cellcolor[rgb]{ .867,  .871,  .847}.091 \\
    rib left 2 & .833 & \textbf{.858} & \textbf{.858} & .667 & \textit{.633} & .792 & \cellcolor[rgb]{ .788,  .824,  .875}.774 & \cellcolor[rgb]{ .886,  .882,  .843}.099 \\
    rib left 3 & .789 & .805 & \textbf{.813} & .797 & .797 & \textit{.756} & \cellcolor[rgb]{ .776,  .82,  .878}.793 & \cellcolor[rgb]{ .722,  .784,  .902}.020 \\
    rib left 4 & \textbf{.917} & .867 & .892 & \textit{.742} & .842 & .783 & \cellcolor[rgb]{ .757,  .808,  .886}.840 & \cellcolor[rgb]{ .816,  .839,  .867}.067 \\
    rib left 5 & .788 & \textbf{.818} & .773 & .780 & .773 & \textit{.735} & \cellcolor[rgb]{ .784,  .824,  .875}.778 & \cellcolor[rgb]{ .737,  .792,  .898}.027 \\
    rib left 6 & .807 & .807 & \textbf{.847} & \textit{.713} & .753 & .720 & \cellcolor[rgb]{ .788,  .824,  .875}.774 & \cellcolor[rgb]{ .792,  .827,  .878}.054 \\
    rib left 7 & .803 & .837 & \textbf{.952} & .844 & \textit{.714} & .776 & \cellcolor[rgb]{ .765,  .812,  .882}.821 & \cellcolor[rgb]{ .843,  .859,  .859}.080 \\
    rib left 8 & .791 & .810 & \textbf{.830} & .765 & \textit{.634} & .804 & \cellcolor[rgb]{ .788,  .824,  .875}.772 & \cellcolor[rgb]{ .827,  .847,  .863}.071 \\
    rib left 9 & \textbf{.850} & \textbf{.850} & .758 & \textbf{.850} & \textit{.712} & .745 & \cellcolor[rgb]{ .776,  .82,  .878}.794 & \cellcolor[rgb]{ .808,  .835,  .871}.063 \\
    rib right 1 & .868 & .842 & \textbf{.886} & .693 & \textit{.675} & \textbf{.886} & \cellcolor[rgb]{ .773,  .816,  .882}.808 & \cellcolor[rgb]{ .882,  .878,  .843}.098 \\
    rib right 10 & \textbf{.843} & .712 & .699 & .771 & \textit{.634} & .712 & \cellcolor[rgb]{ .808,  .835,  .867}.729 & \cellcolor[rgb]{ .827,  .847,  .863}.071 \\
    rib right 11 & .758 & \textbf{.863} & .791 & .706 & \textit{.562} & .673 & \cellcolor[rgb]{ .808,  .839,  .867}.725 & \cellcolor[rgb]{ .894,  .886,  .839}.104 \\
    rib right 12 & .596 & .596 & \textbf{.667} & .596 & \textit{.355} & .504 & \cellcolor[rgb]{ .89,  .886,  .839}.552 & \cellcolor[rgb]{ .906,  .894,  .835}.110 \\
    rib right 2 & \textbf{.863} & .786 & .838 & \textit{.624} & .701 & .821 & \cellcolor[rgb]{ .788,  .824,  .875}.772 & \cellcolor[rgb]{ .871,  .871,  .847}.092 \\
    rib right 3 & .780 & .780 & \textbf{.829} & .748 & .797 & \textit{.659} & \cellcolor[rgb]{ .792,  .827,  .875}.766 & \cellcolor[rgb]{ .8,  .831,  .875}.059 \\
    rib right 4 & \textbf{.892} & .883 & .867 & \textit{.767} & .850 & .817 & \cellcolor[rgb]{ .753,  .804,  .886}.846 & \cellcolor[rgb]{ .776,  .816,  .882}.047 \\
    rib right 5 & .806 & \textbf{.853} & .798 & .767 & .791 & \textit{.674} & \cellcolor[rgb]{ .784,  .824,  .878}.782 & \cellcolor[rgb]{ .804,  .831,  .871}.060 \\
    rib right 6 & .803 & .741 & \textbf{.837} & .728 & .741 & \textit{.694} & \cellcolor[rgb]{ .796,  .827,  .875}.757 & \cellcolor[rgb]{ .788,  .824,  .878}.052 \\
    rib right 7 & .823 & .844 & \textbf{.898} & .823 & .735 & \textit{.714} & \cellcolor[rgb]{ .773,  .816,  .882}.806 & \cellcolor[rgb]{ .824,  .843,  .867}.069 \\
    rib right 8 & .758 & .765 & \textbf{.843} & .804 & \textit{.641} & .784 & \cellcolor[rgb]{ .792,  .827,  .875}.766 & \cellcolor[rgb]{ .824,  .843,  .867}.069 \\
    rib right 9 & .817 & \textbf{.856} & .752 & .817 & \textit{.641} & .765 & \cellcolor[rgb]{ .788,  .824,  .875}.775 & \cellcolor[rgb]{ .835,  .851,  .859}.076 \\
    sacrum & .871 & .871 & .886 & \textbf{.894} & \textit{.818} & .879 & \cellcolor[rgb]{ .741,  .8,  .89}.870 & \cellcolor[rgb]{ .737,  .792,  .898}.027 \\
    scapula left & \textbf{.927} & .894 & .886 & .846 & \textit{.821} & .902 & \cellcolor[rgb]{ .737,  .796,  .894}.879 & \cellcolor[rgb]{ .761,  .808,  .886}.039 \\
    scapula right & \textbf{.922} & .915 & \textbf{.922} & .752 & \textit{.744} & .899 & \cellcolor[rgb]{ .749,  .8,  .89}.859 & \cellcolor[rgb]{ .859,  .867,  .851}.087 \\
    small bowel & .931 & \textit{.840} & .882 & .903 & \textbf{.951} & .910 & \cellcolor[rgb]{ .725,  .788,  .898}.903 & \cellcolor[rgb]{ .761,  .808,  .886}.039 \\
    spleen & .767 & \textbf{.827} & .713 & .667 & \textit{.547} & .680 & \cellcolor[rgb]{ .82,  .843,  .863}.700 & \cellcolor[rgb]{ .878,  .875,  .847}.096 \\
    stomach & \textbf{.843} & .719 & .784 & .719 & \textit{.569} & .706 & \cellcolor[rgb]{ .812,  .839,  .867}.723 & \cellcolor[rgb]{ .871,  .871,  .847}.092 \\
    trachea & .870 & \textbf{.943} & .829 & \textit{.780} & .837 & .846 & \cellcolor[rgb]{ .749,  .804,  .89}.851 & \cellcolor[rgb]{ .792,  .827,  .875}.054 \\
    urinary bladder & .798 & .705 & .853 & \textbf{.860} & .729 & \textit{.659} & \cellcolor[rgb]{ .788,  .827,  .875}.767 & \cellcolor[rgb]{ .851,  .859,  .855}.082 \\
    vertebrae C1 & .524 & .595 & \textbf{.762} & \textit{.476} & .548 & .571 & \cellcolor[rgb]{ .878,  .878,  .843}.579 & \cellcolor[rgb]{ .882,  .878,  .843}.098 \\
    vertebrae C2 & .744 & .667 & \textbf{.846} & .692 & \textit{.513} & .718 & \cellcolor[rgb]{ .824,  .847,  .863}.697 & \cellcolor[rgb]{ .906,  .894,  .835}.109 \\
    vertebrae C3 & .595 & \textbf{.619} & .595 & .548 & \textit{.405} & .476 & \cellcolor[rgb]{ .894,  .89,  .835}.540 & \cellcolor[rgb]{ .851,  .863,  .855}.083 \\
    vertebrae C4 & .356 & .311 & .400 & .400 & \textbf{.444} & \textit{.222} & \cellcolor[rgb]{ .98,  .937,  .804}.356 & \cellcolor[rgb]{ .843,  .859,  .859}.080 \\
    vertebrae C5 & .544 & .491 & \textbf{.632} & .456 & .351 & \textit{.333} & \cellcolor[rgb]{ .929,  .91,  .824}.468 & \cellcolor[rgb]{ .918,  .898,  .831}.114 \\
    vertebrae C6 & \textbf{.483} & .437 & .437 & .391 & \textit{.241} & .437 & \cellcolor[rgb]{ .957,  .925,  .812}.404 & \cellcolor[rgb]{ .855,  .863,  .855}.085 \\
    vertebrae C7 & .676 & \textbf{.722} & .667 & .537 & \textit{.426} & .657 & \cellcolor[rgb]{ .863,  .867,  .847}.614 & \cellcolor[rgb]{ .91,  .894,  .835}.111 \\
    vertebrae L1 & \textbf{.728} & .558 & .687 & .680 & \textit{.442} & .612 & \cellcolor[rgb]{ .859,  .867,  .851}.618 & \cellcolor[rgb]{ .898,  .886,  .839}.105 \\
    vertebrae L2 & .636 & .538 & \textbf{.682} & .621 & \textit{.477} & .523 & \cellcolor[rgb]{ .878,  .878,  .843}.580 & \cellcolor[rgb]{ .843,  .855,  .859}.079 \\
    vertebrae L3 & \textbf{.721} & .496 & .566 & \textit{.488} & .667 & .698 & \cellcolor[rgb]{ .867,  .871,  .847}.606 & \cellcolor[rgb]{ .894,  .886,  .839}.103 \\
    vertebrae L4 & .636 & \textit{.379} & .568 & \textbf{.795} & .644 & .576 & \cellcolor[rgb]{ .867,  .871,  .847}.600 & \cellcolor[rgb]{ .961,  .925,  .816}.136 \\
    vertebrae L5 & .744 & \textit{.566} & \textbf{.767} & .729 & .643 & .628 & \cellcolor[rgb]{ .831,  .851,  .859}.680 & \cellcolor[rgb]{ .843,  .855,  .859}.079 \\
    vertebrae T1 & .622 & .604 & .658 & \textbf{.739} & \textit{.315} & .721 & \cellcolor[rgb]{ .863,  .871,  .847}.610 & \cellcolor[rgb]{ 1,  .949,  .8}\textit{.154} \\
    vertebrae T10 & .571 & .524 & \textbf{.619} & .599 & \textit{.483} & .490 & \cellcolor[rgb]{ .89,  .886,  .839}.548 & \cellcolor[rgb]{ .8,  .831,  .875}.057 \\
    vertebrae T11 & .472 & .458 & .493 & \textbf{.563} & .431 & \textit{.410} & \cellcolor[rgb]{ .925,  .906,  .824}.471 & \cellcolor[rgb]{ .792,  .827,  .878}.054 \\
    vertebrae T12 & .527 & .527 & \textbf{.613} & .587 & .533 & \textit{.513} & \cellcolor[rgb]{ .89,  .886,  .839}.550 & \cellcolor[rgb]{ .765,  .808,  .886}.040 \\
    vertebrae T2 & .632 & .623 & \textit{.509} & .728 & .675 & \textbf{.737} & \cellcolor[rgb]{ .843,  .859,  .855}.651 & \cellcolor[rgb]{ .855,  .863,  .855}.084 \\
    vertebrae T3 & .728 & .658 & .746 & .684 & \textit{.640} & \textbf{.754} & \cellcolor[rgb]{ .82,  .843,  .863}.702 & \cellcolor[rgb]{ .78,  .82,  .882}.048 \\
    vertebrae T4 & .761 & \textbf{.778} & .761 & \textit{.624} & .658 & .641 & \cellcolor[rgb]{ .82,  .843,  .863}.704 & \cellcolor[rgb]{ .824,  .847,  .863}.070 \\
    vertebrae T5 & .692 & .709 & \textbf{.786} & .692 & \textit{.547} & .581 & \cellcolor[rgb]{ .835,  .855,  .859}.668 & \cellcolor[rgb]{ .863,  .867,  .851}.088 \\
    vertebrae T6 & .676 & .667 & .519 & \textbf{.722} & .500 & \textit{.481} & \cellcolor[rgb]{ .871,  .875,  .847}.594 & \cellcolor[rgb]{ .898,  .89,  .839}.105 \\
    vertebrae T7 & .650 & .624 & \textbf{.709} & .581 & \textit{.453} & .590 & \cellcolor[rgb]{ .867,  .871,  .847}.601 & \cellcolor[rgb]{ .859,  .867,  .851}.086 \\
    vertebrae T8 & .576 & .545 & \textbf{.750} & .462 & .439 & \textit{.348} & \cellcolor[rgb]{ .906,  .894,  .831}.520 & \cellcolor[rgb]{ .965,  .929,  .812}.139 \\
    vertebrae T9 & \textit{.435} & .599 & \textbf{.721} & .537 & .497 & .592 & \cellcolor[rgb]{ .886,  .882,  .839}.563 & \cellcolor[rgb]{ .882,  .878,  .843}.098 \\ \hline
    Average  & .764 & .754 & \textbf{.790} & .736 & \textit{.662} & .720 &       &  \\
    STD   & .148 & .150 & \textbf{.142} & .145 & \textit{.171} & .166 &       &  \\
\label{tab:localization-ratio-104-regions-reranking}
\end{longtable}
\endgroup

\section{Discussion}

\subsection{Dataset}

\begin{figure*}[!htpb]
  \centering
  \begin{minipage}{ .9\textwidth}
    \centering
    \subcaptionbox{}{\includegraphics[scale= .48]{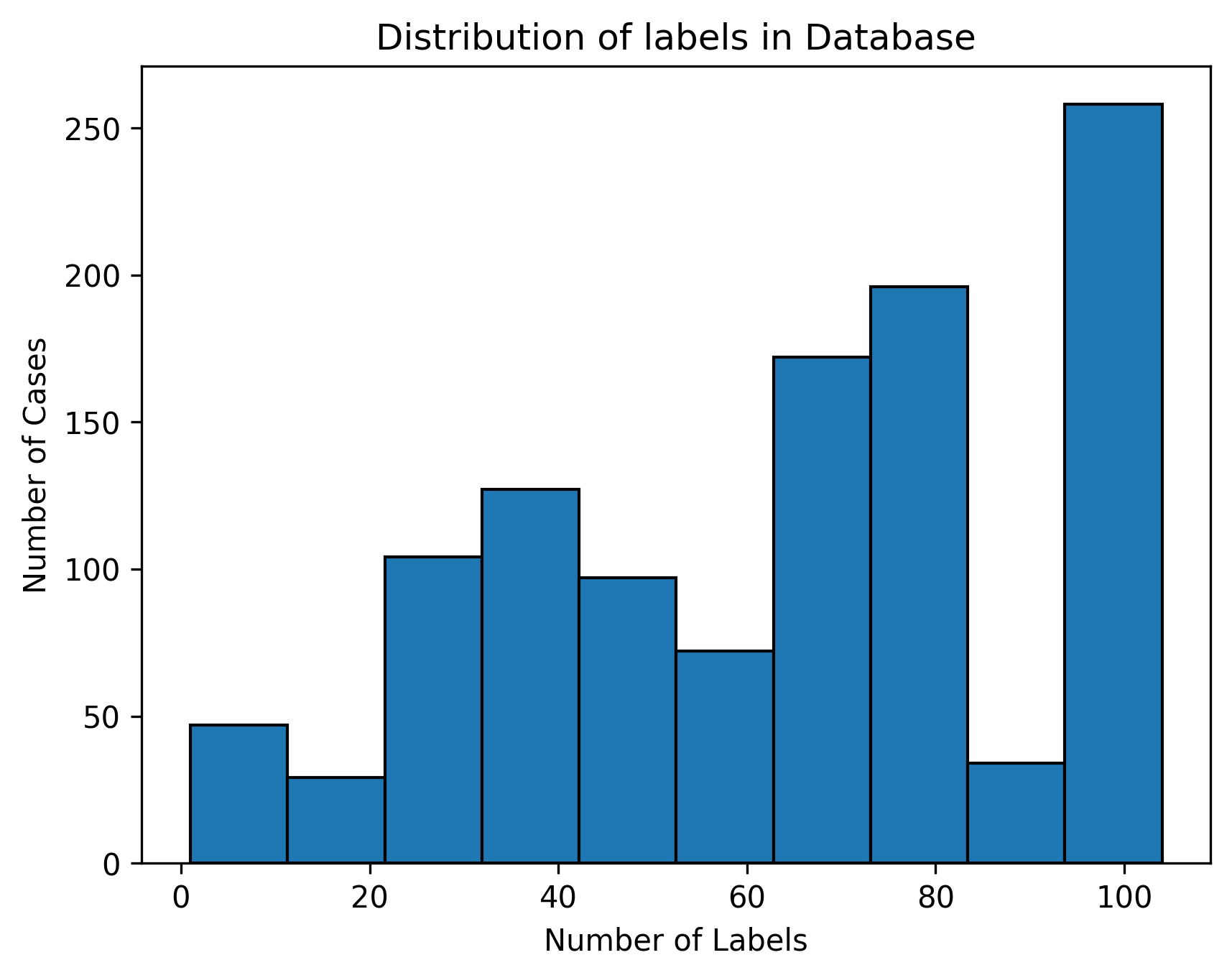}}
    \subcaptionbox{}{\includegraphics[scale= .48]{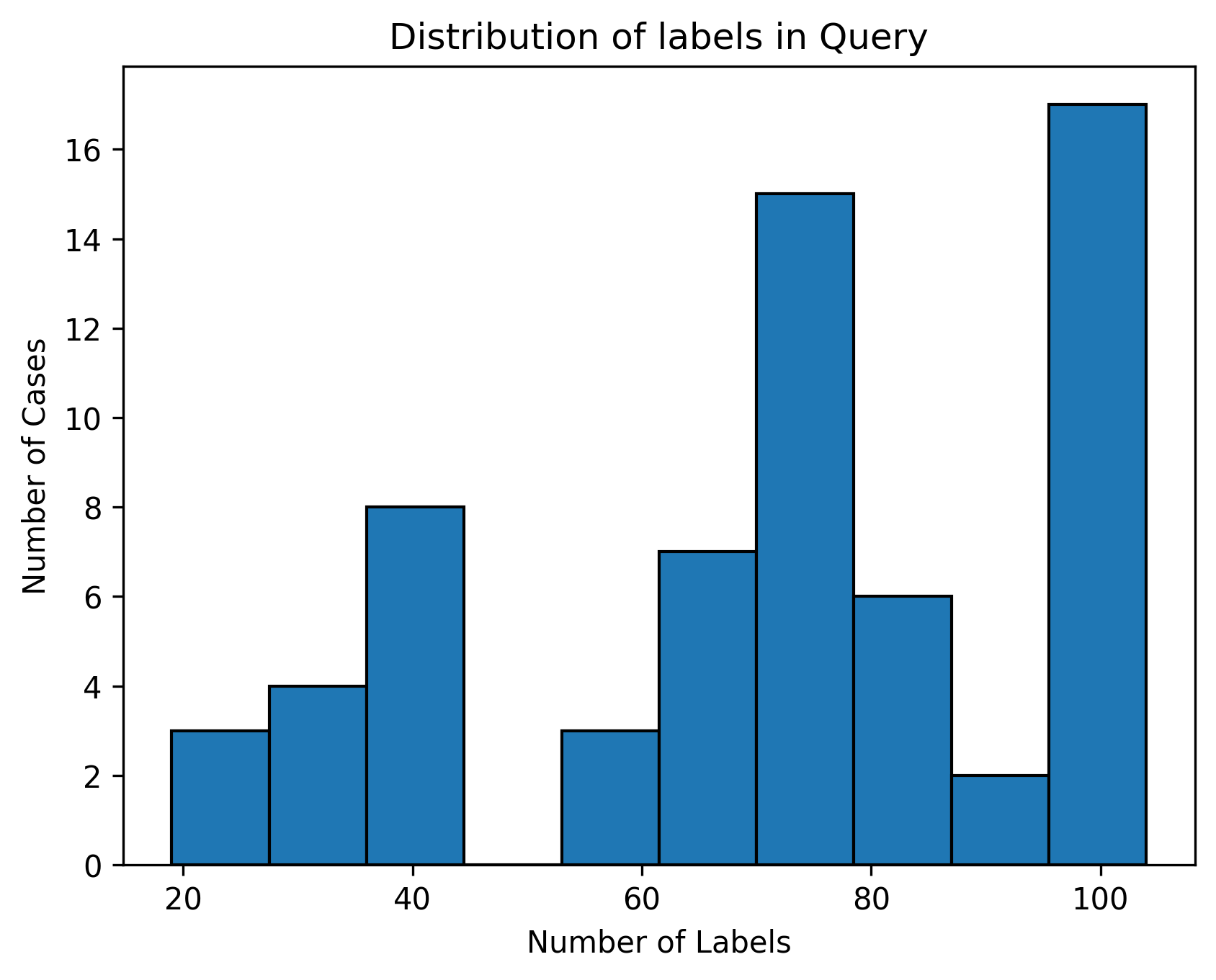}}
    \caption{Distribution of the classes in database (a) and query (b) volumes.}
    \label{fig:label-distribution}
  \end{minipage}
\end{figure*}

As depicted in \Cref{fig:label-distribution}, the labels inside the database and query subset (derived from TS train and test set, respectively) are not balanced. This should resemble a pattern as can be observed in future real-world scenarios of image retrieval. At the same time, this imbalance should be kept in mind when reading and interpreting recall values from the provided result tables.

\begin{figure*}[!htpb]
  \centering
  \begin{minipage}{ .9\textwidth}
    \centering
    \subcaptionbox{}{\includegraphics[scale= .15]{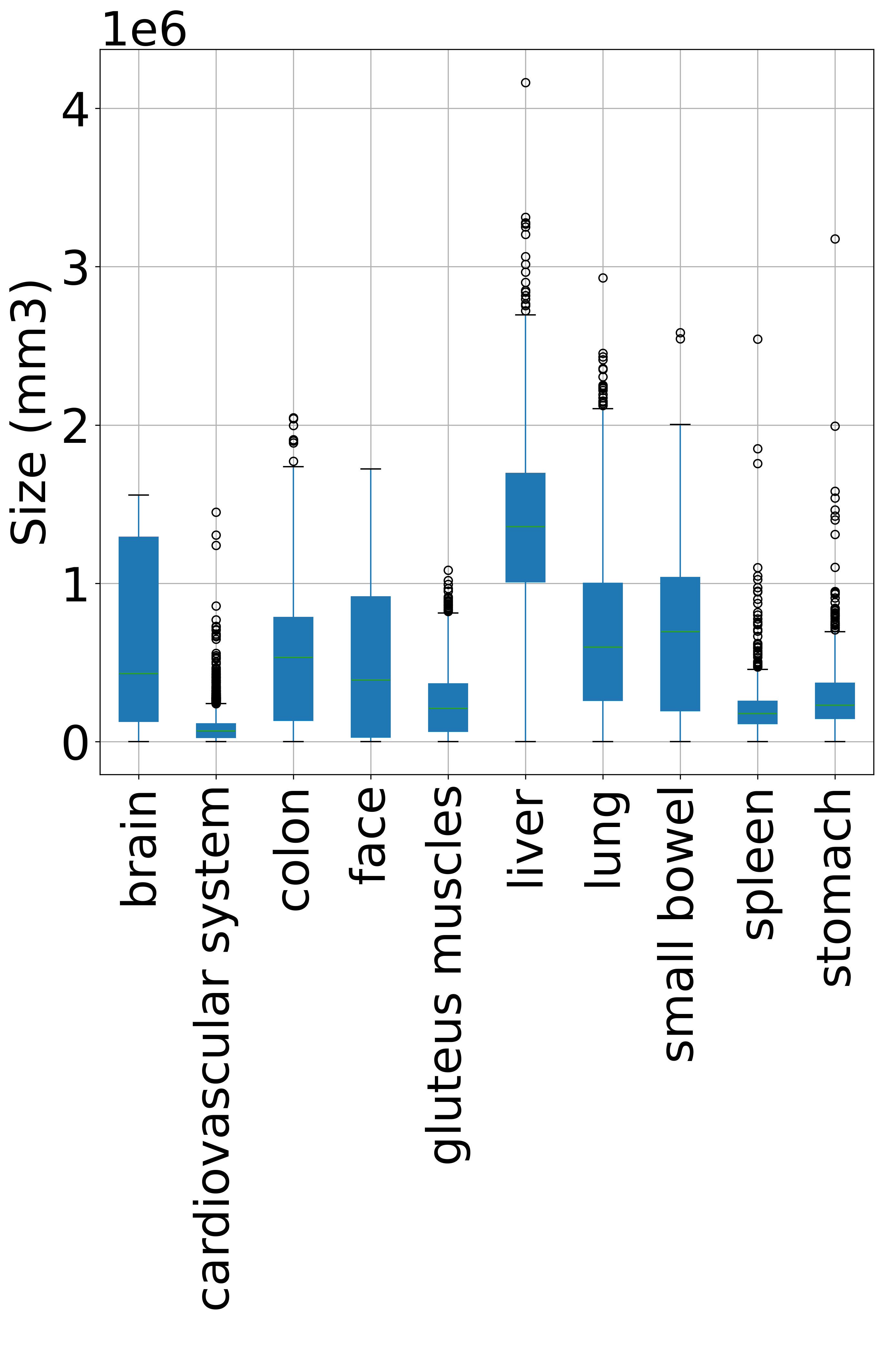}} 
    \subcaptionbox{}{\raisebox{ .65cm}{\includegraphics[scale= .15]{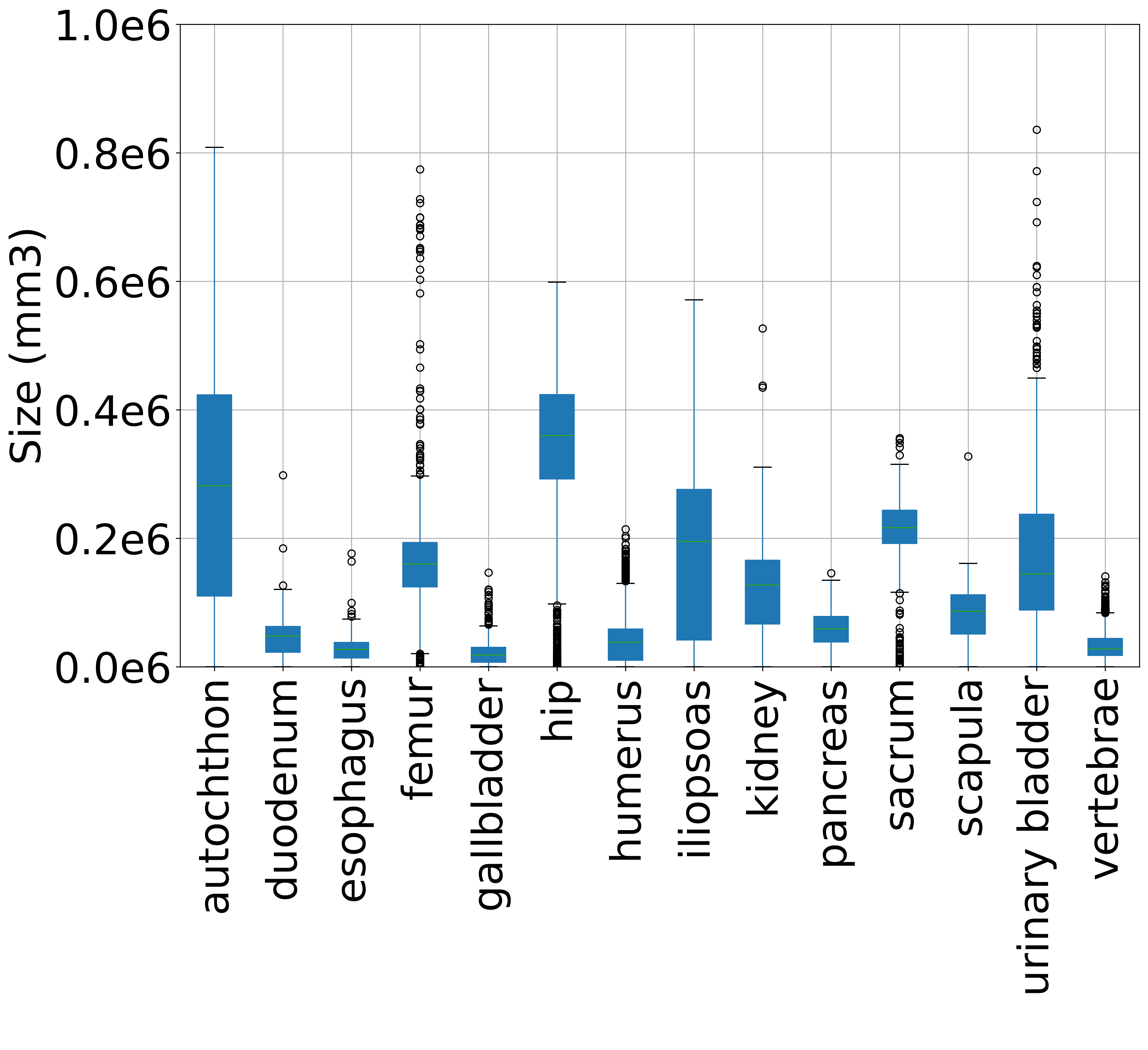}}}
    \subcaptionbox{}{\raisebox{ .2cm}{\includegraphics[scale= .15]{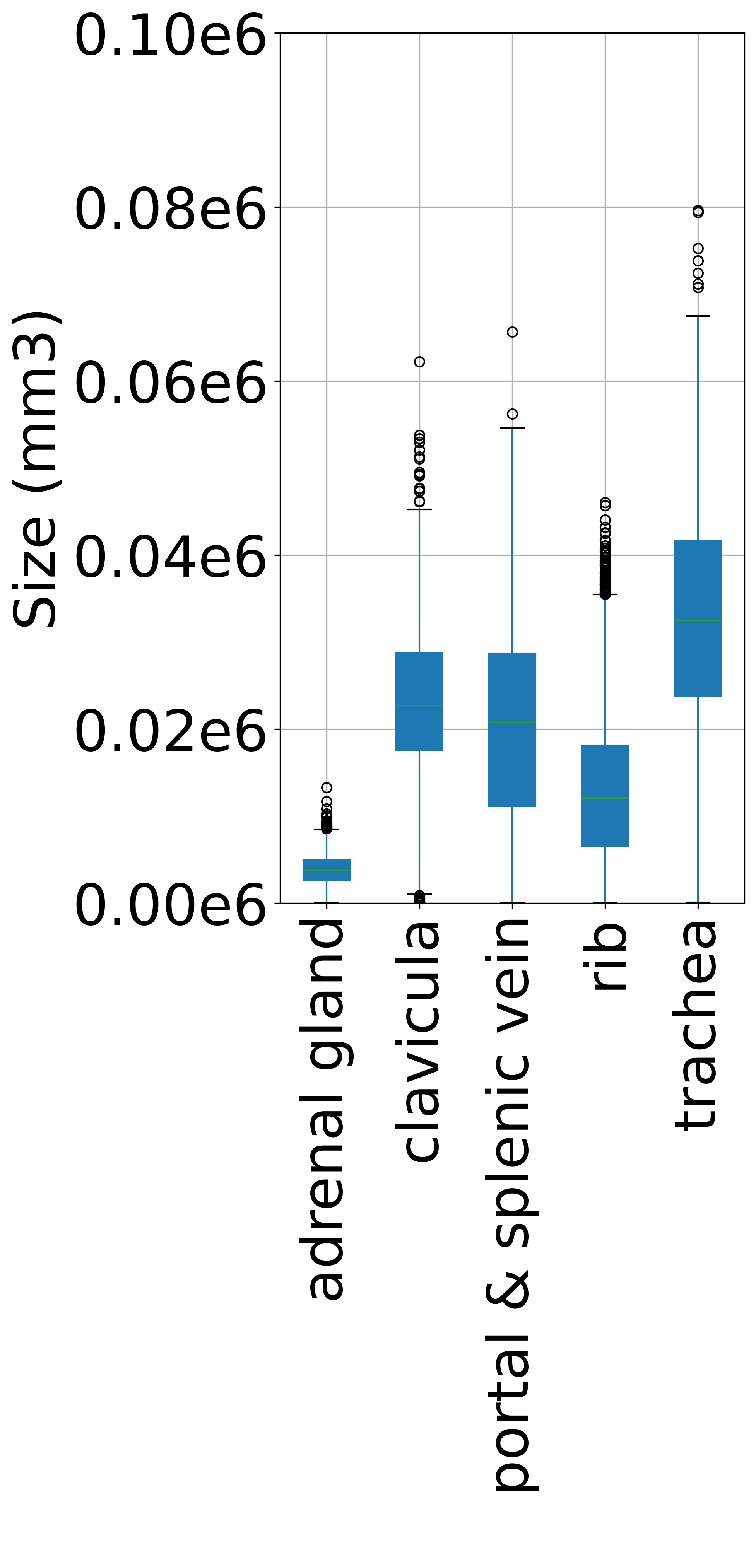}}}
    \caption{Distribution of the size of the anatomical regions for the 29 regions (a) big anatomical regions with a physical size range that exceeds $10^6$~$mm^3$, (b) mid-size anatomical regions with a physical size in range in the range $( .1:1)\times10^6$~$mm^3$ and (c) small anatomical regions with a physical size smaller than $ .1\times10^6$~$mm^3$.}
    \label{fig:coarser-size}
  \end{minipage}
\end{figure*}

\begin{figure*}[!htpb]
  \centering
  \begin{minipage}{ .9\textwidth}
    \centering
    \subcaptionbox{}{\includegraphics[scale= .13]{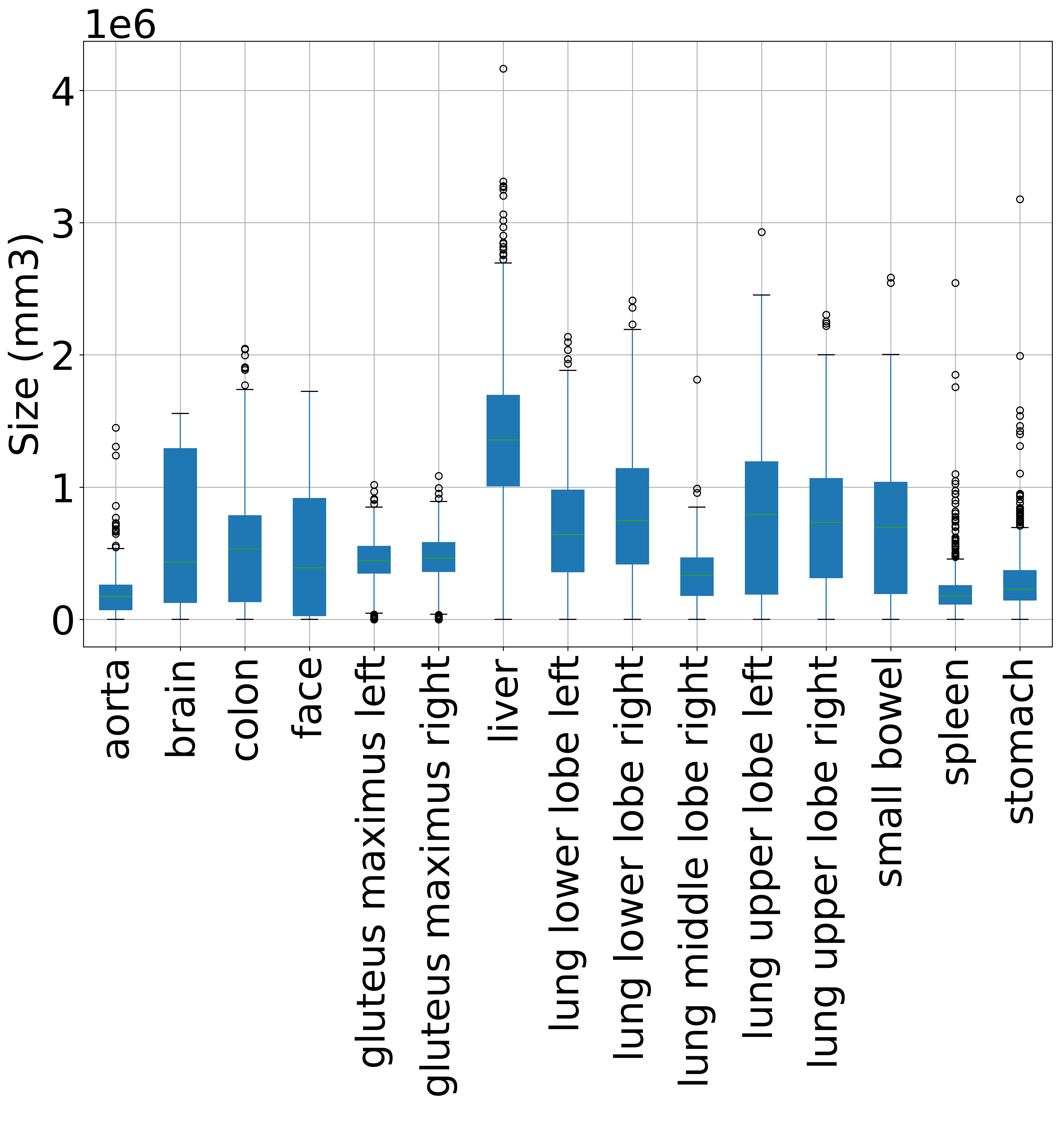}} 
    \subcaptionbox{}{{\includegraphics[scale= .13]{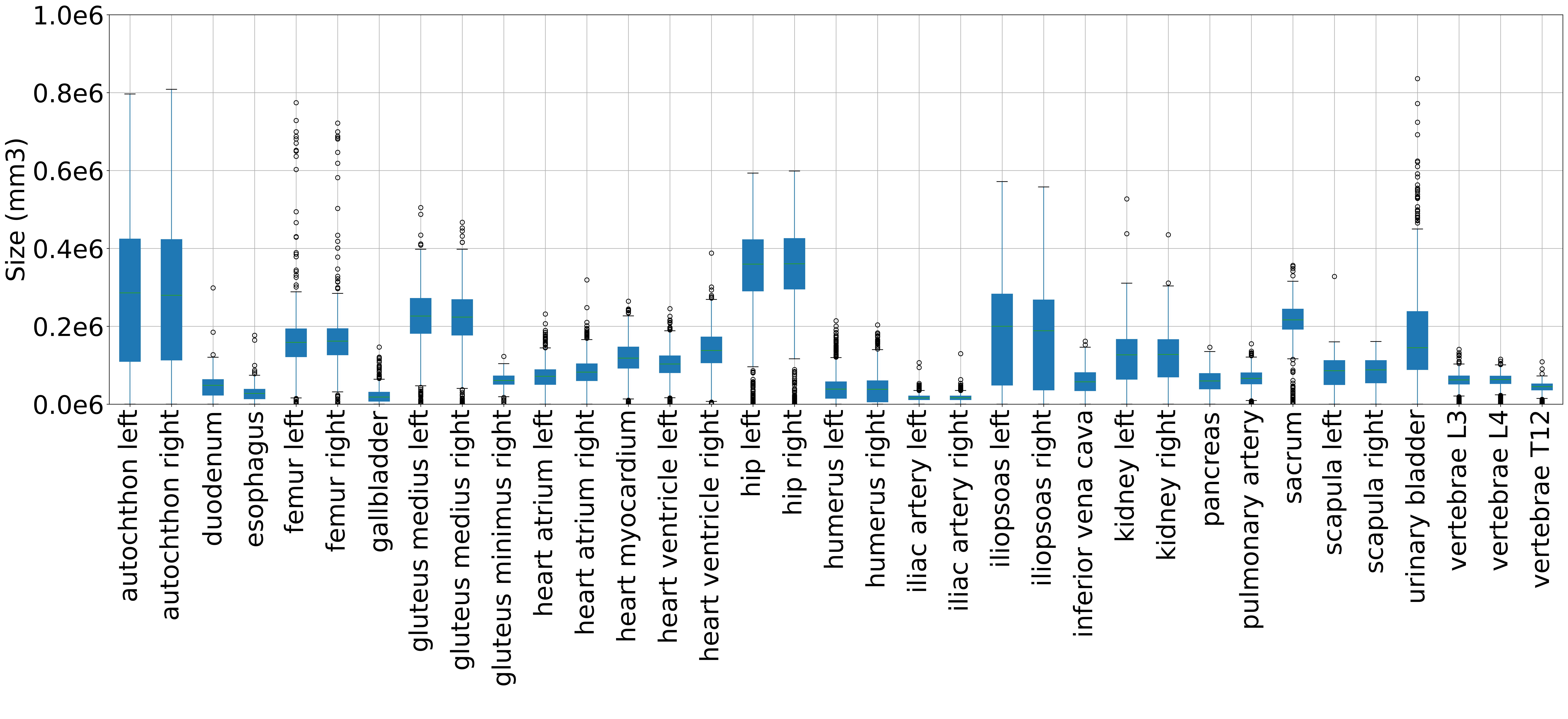}}} \\ 
    \subcaptionbox{}{{\includegraphics[width=\textwidth]{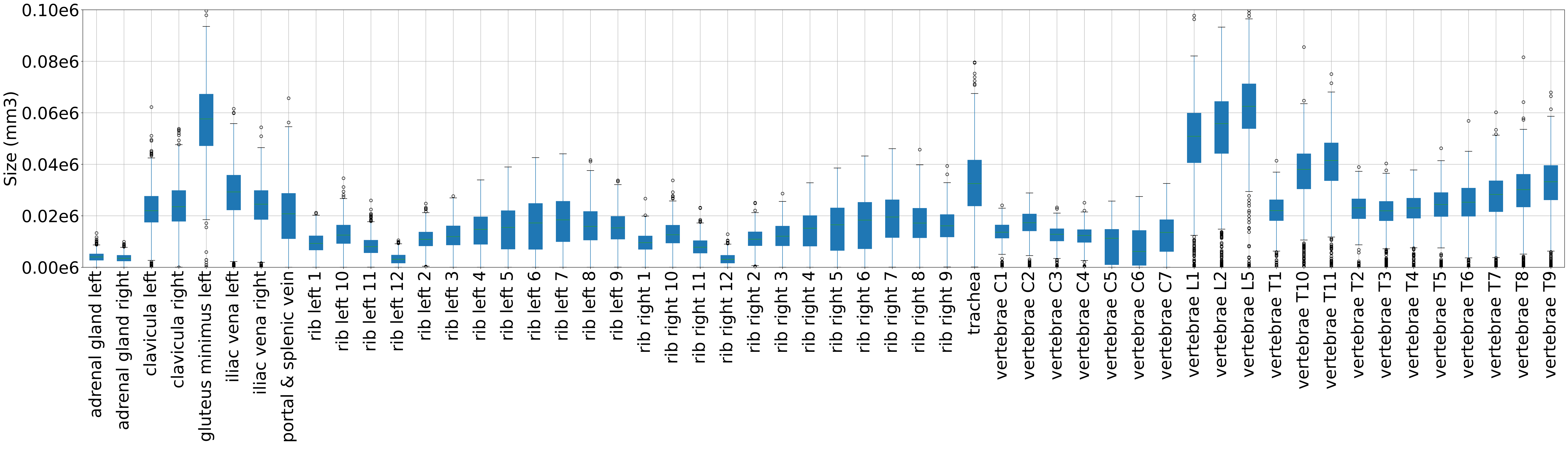}}}
    \caption{Distribution of the size of the anatomical regions for the 104 regions (a) big anatomical regions with a physical size range that exceeds $10^6$~$mm^3$, (b) mid-size anatomical regions with a physical size in range in the range $( .1:1)\times10^6$~$mm^3$ and (c) small anatomical regions with a physical size smaller than $ .1\times10^6$~$mm^3$.}
    \label{fig:all-size}
  \end{minipage}
\end{figure*}

Additionally, it is worth noting that the size and shape of organs can impact the probability of correctly predicting a given label by chance. For example, smaller organs can be less likely to collect "by-chance" true positive predictions compared to larger organs. Similarly, organs with elongated shapes aligned with the slice-wise sampling direction can increase the likelihood of "by-chance" hits. A volume and shape-adjusted representation of recall values does not seem reasonable and thus has not been performed in this work. However, organ volume as shown in \Cref{fig:coarser-size} and \Cref{fig:all-size} should be considered while interpreting result tables.

\begin{figure*}
  \centering
  \begin{minipage}{ .98\textwidth}
    \centering
    \subcaptionbox{}{\includegraphics[scale= .17]{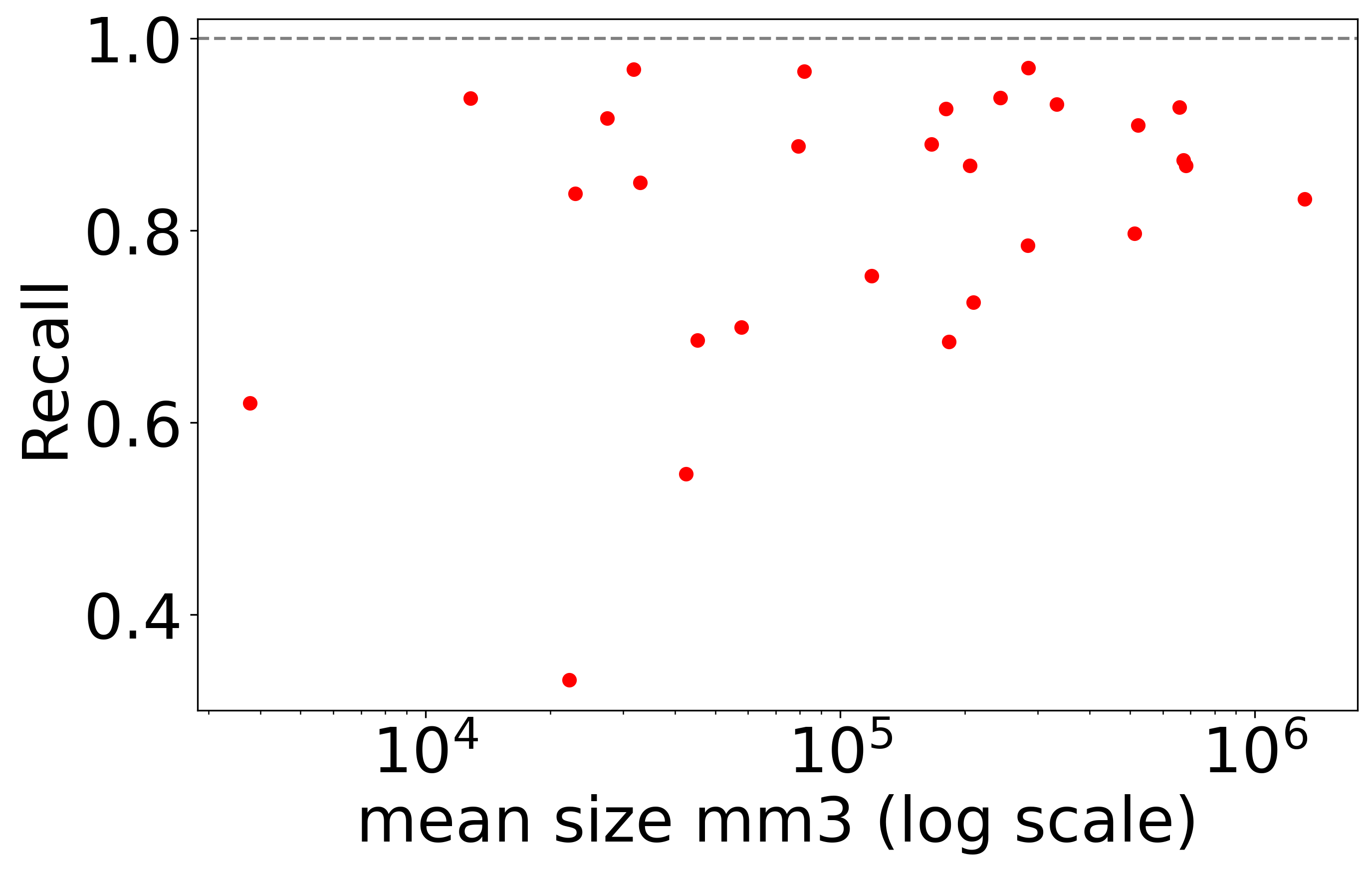}} 
    \subcaptionbox{}{{\includegraphics[scale= .17]{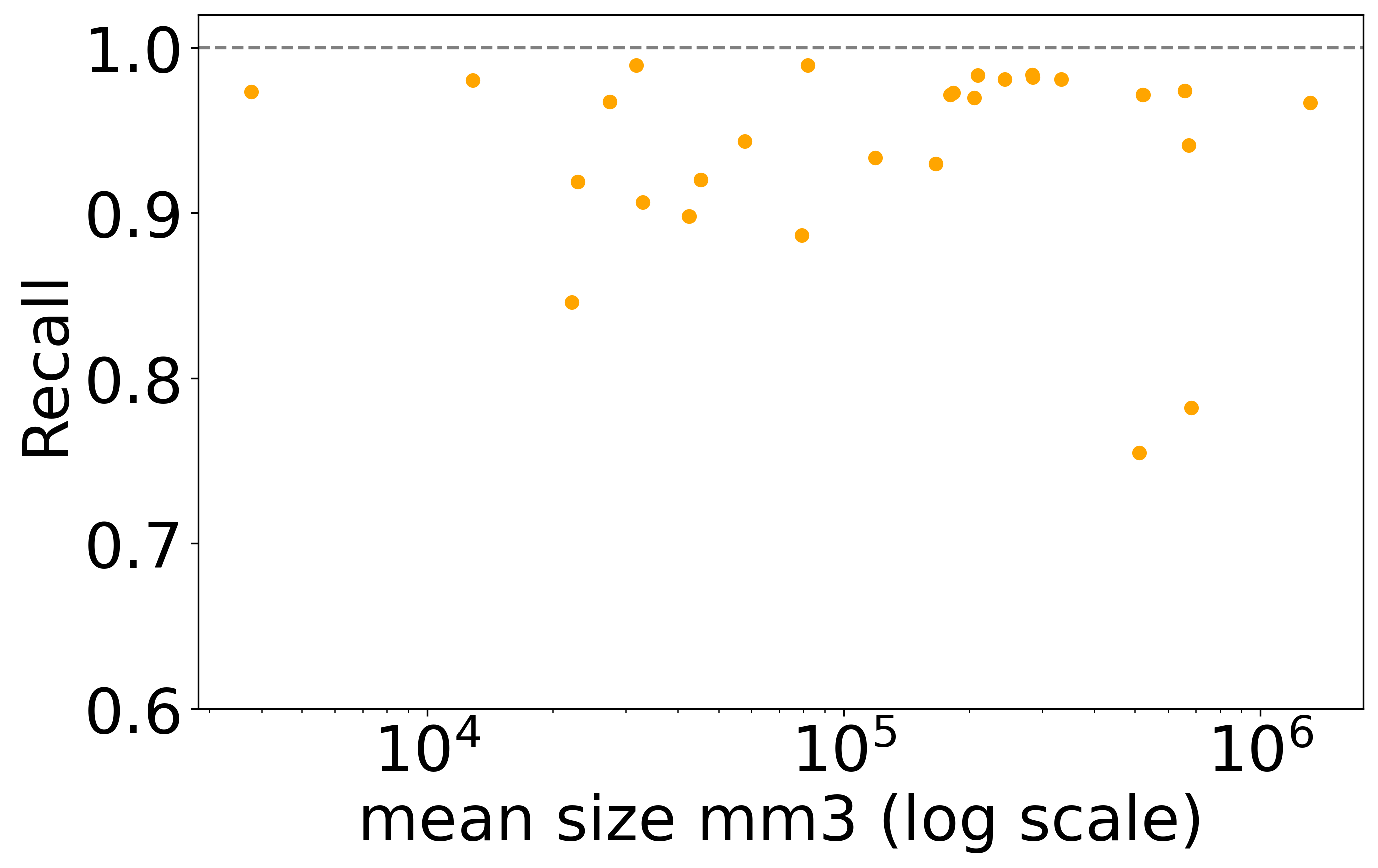}}} 
    \subcaptionbox{}{{\includegraphics[scale= .17]{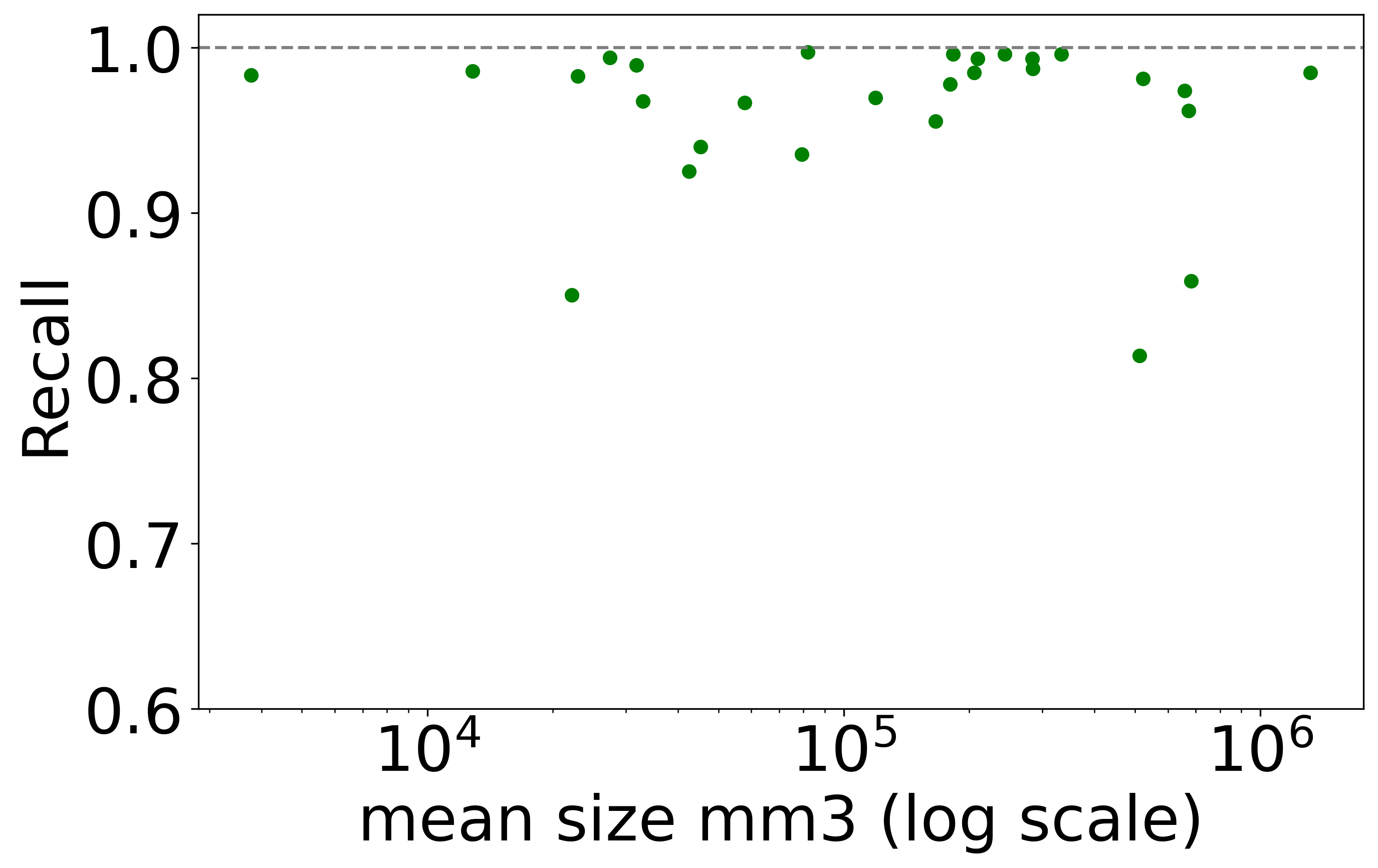}}} \\
    \subcaptionbox{}{{\includegraphics[scale= .17]{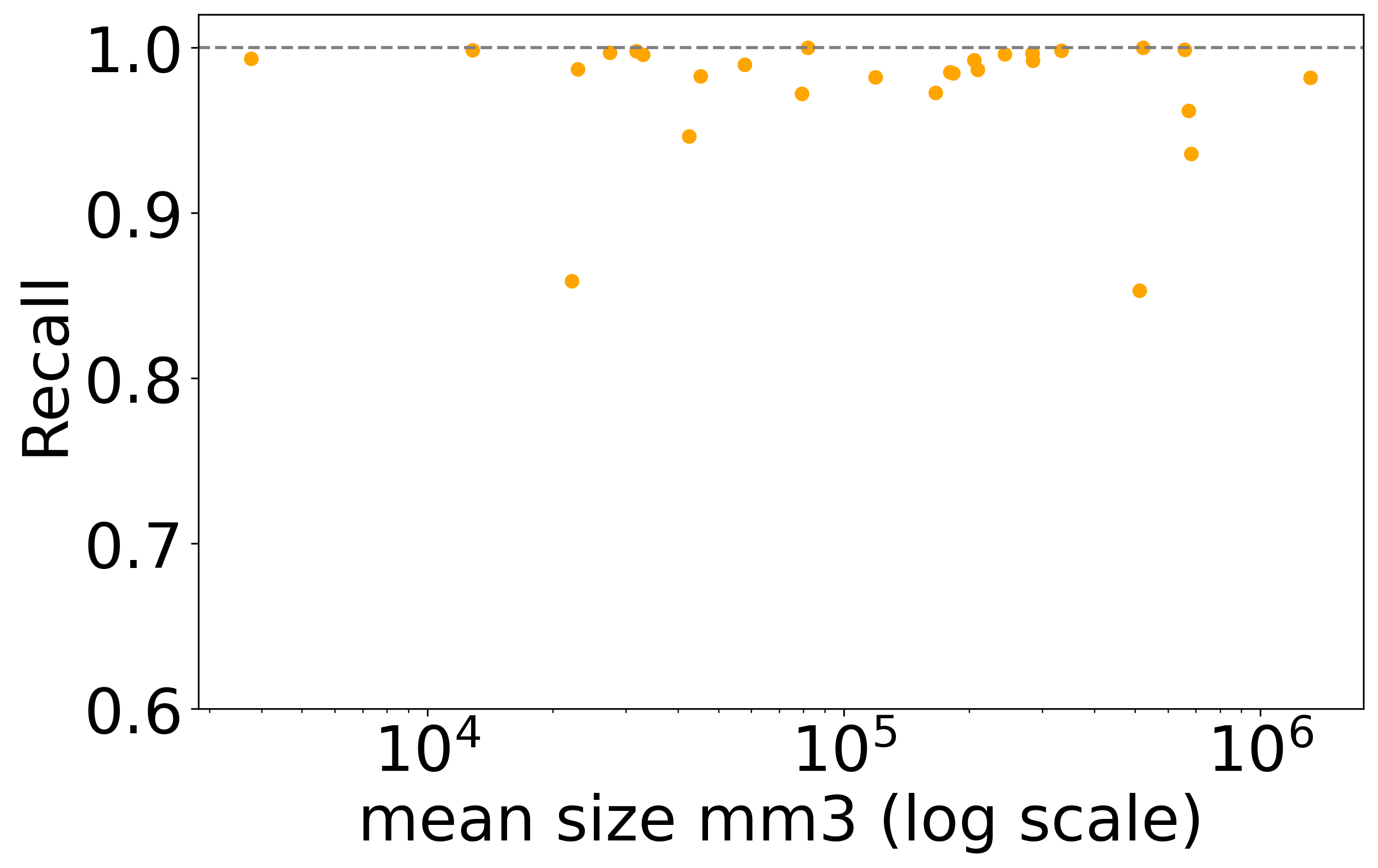}}}
    \subcaptionbox{}{{\includegraphics[scale= .17
    ]{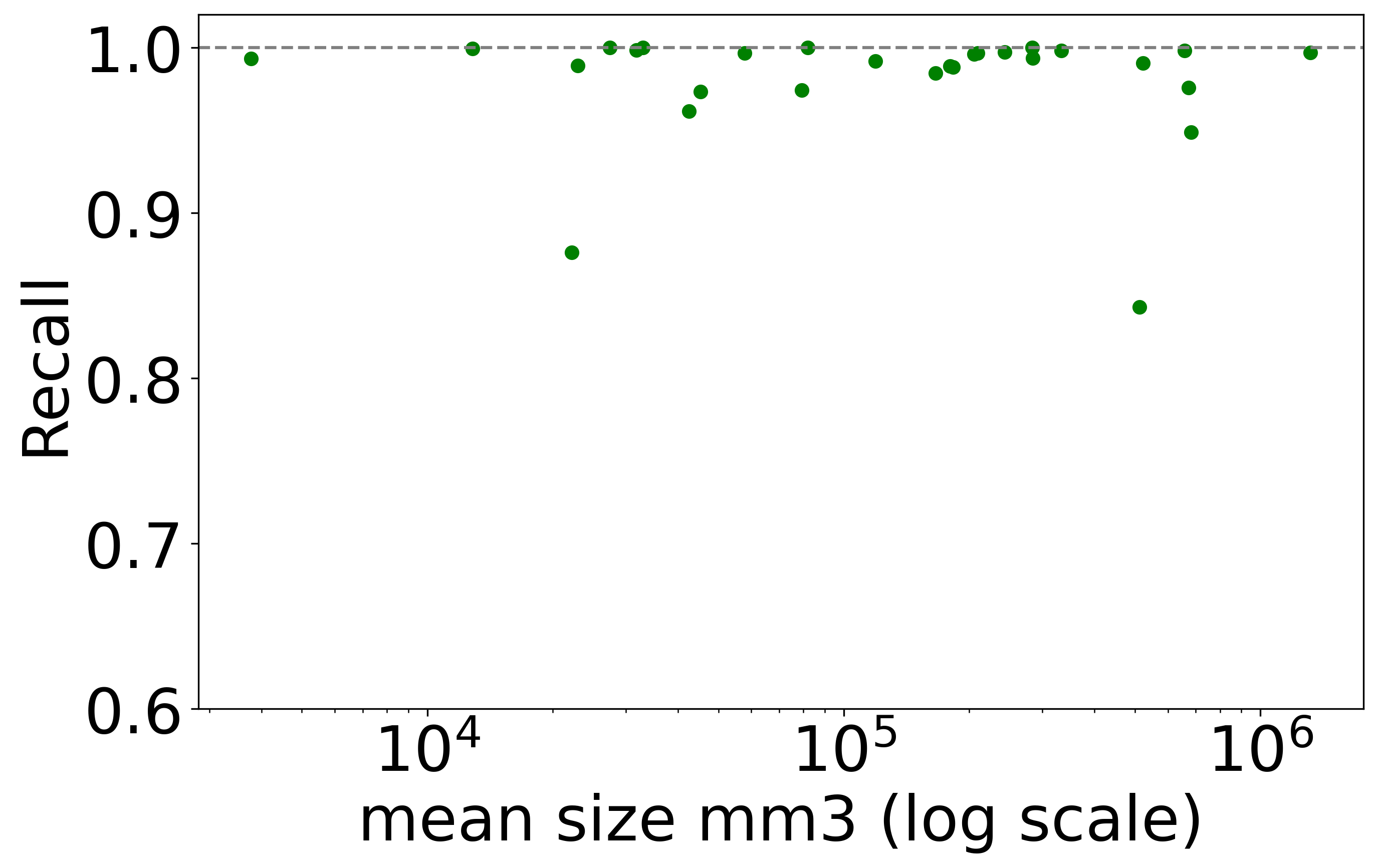}}}
    \subcaptionbox{}{{\includegraphics[scale= .17
    ]{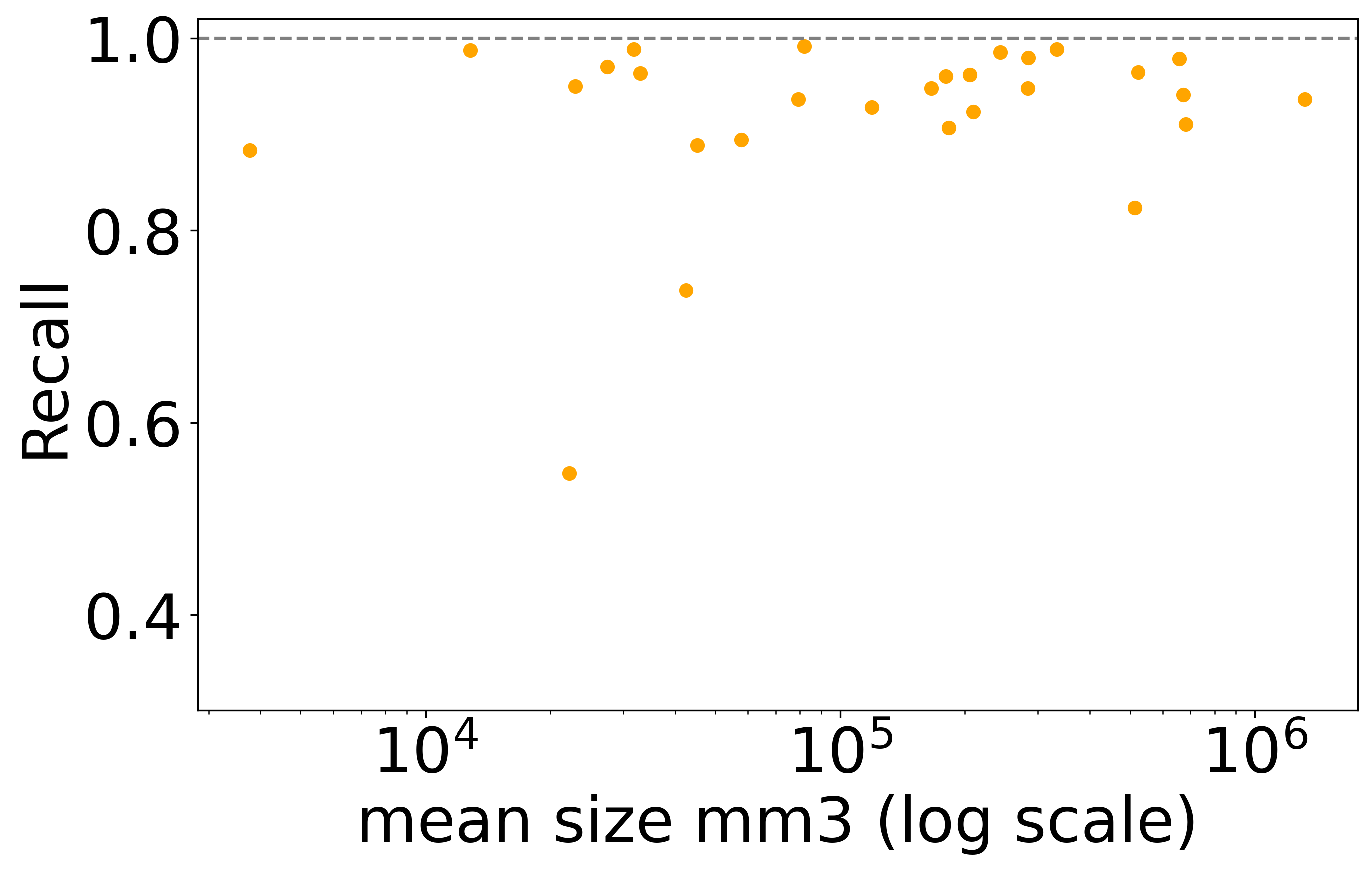}}}
    \subcaptionbox{}{{\includegraphics[scale= .17
    ]{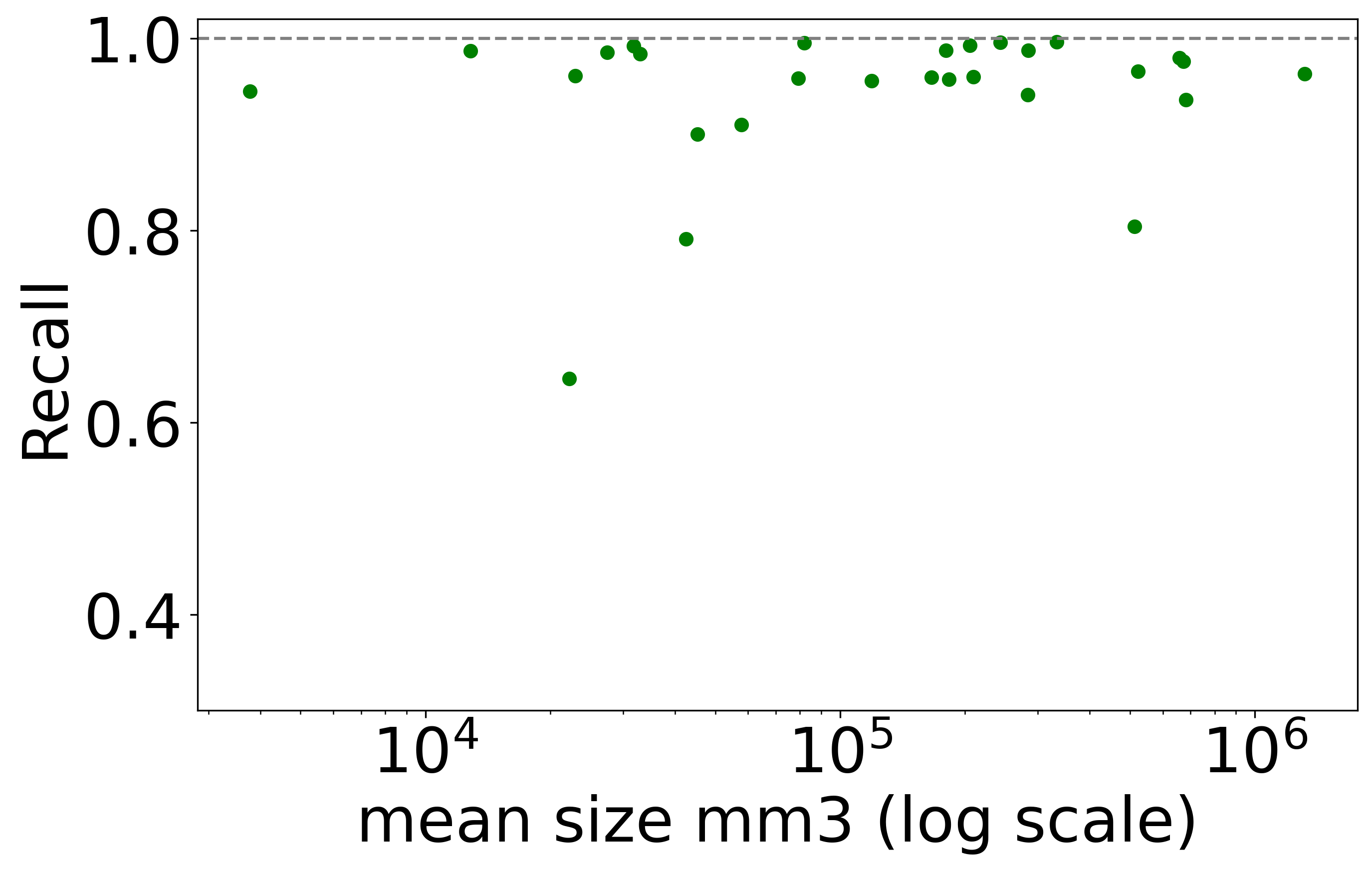}}}
    \caption{Overview of average recall vs. mean anatomical region size for 29 anatomical regions for (a) slice-wise, (b) volume-based, (c) volume-based and re-ranking, (d) region-based, (e) region-based and re-ranking, (f) localized, (g) localized and re-ranking retrieval.  }
    \label{fig:coarser-recall-vs-mean}
  \end{minipage}
  
  \centering
  \begin{minipage}{ .98\textwidth}
    \centering
    \subcaptionbox{}{\includegraphics[scale= .17]{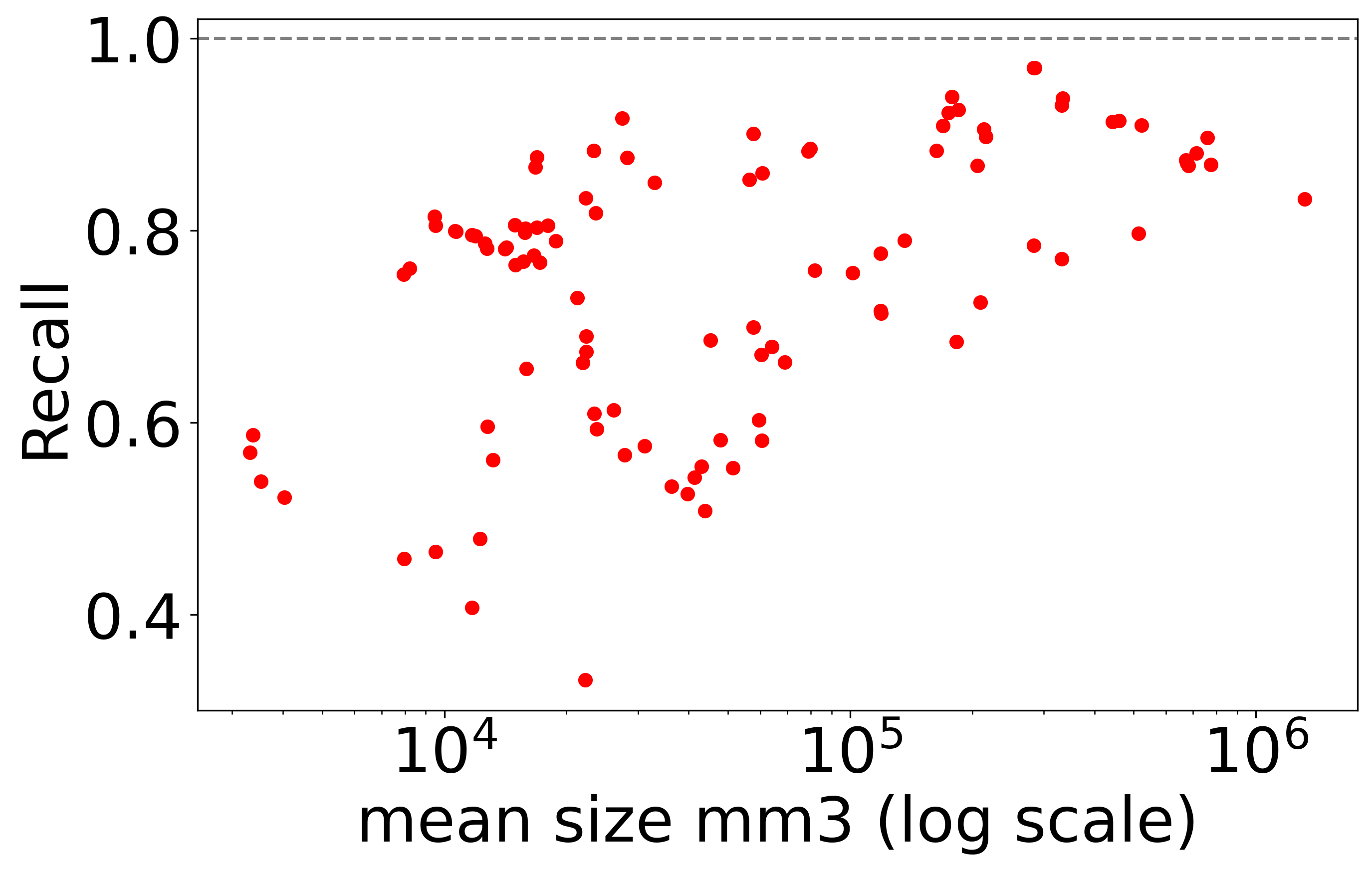}} 
    \subcaptionbox{}{{\includegraphics[scale= .17]{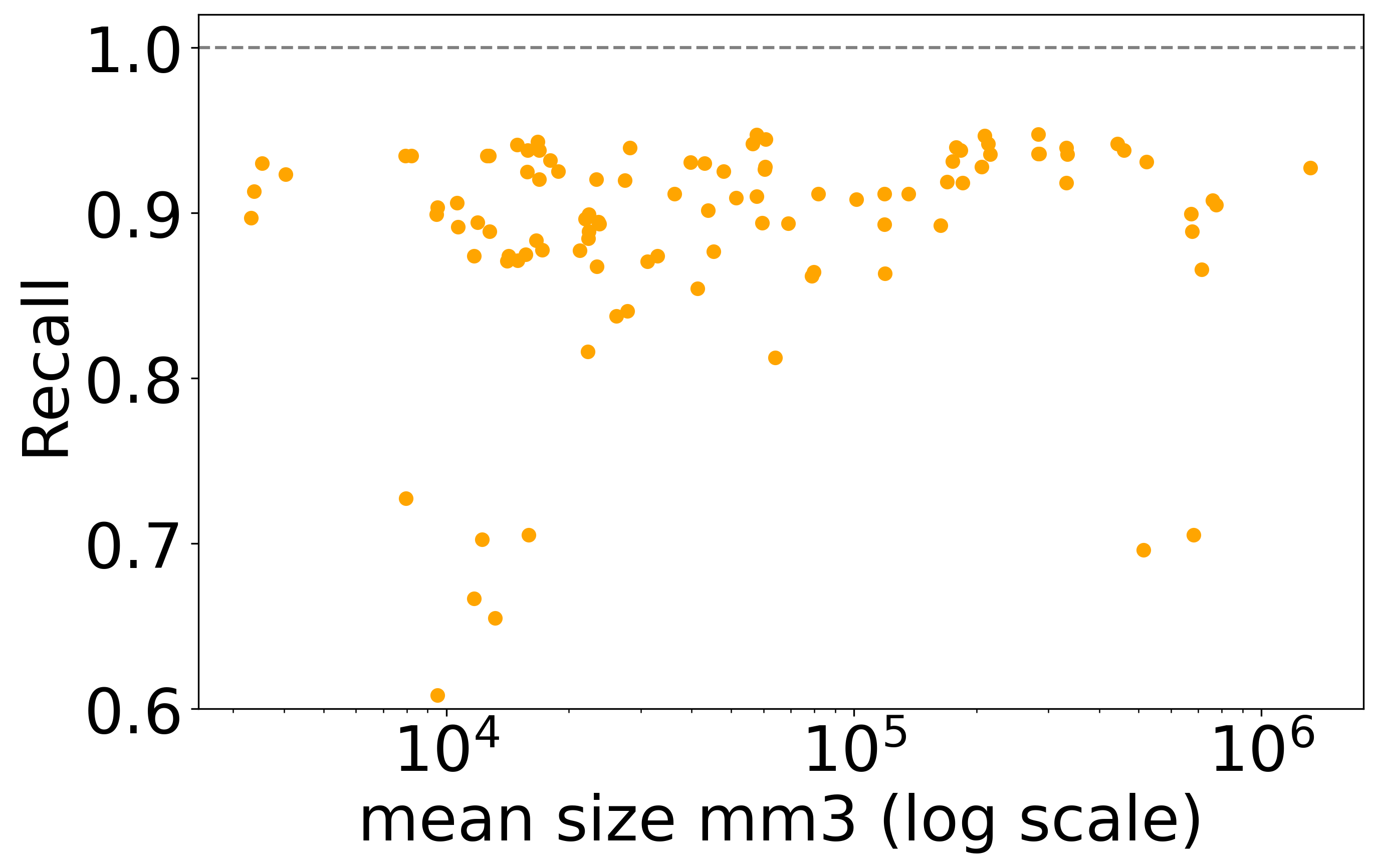}}} 
    \subcaptionbox{}{{\includegraphics[scale= .17]{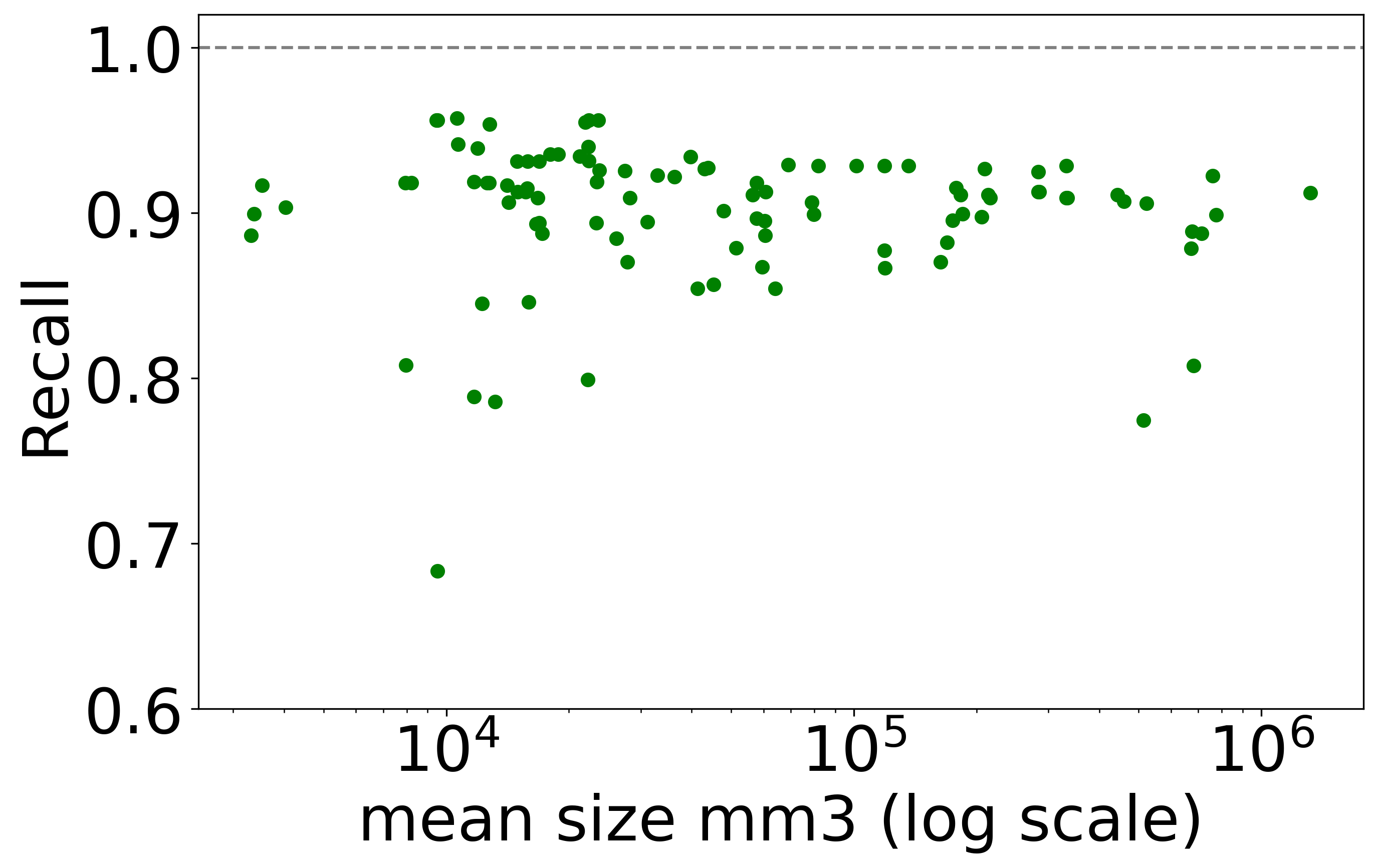}}} \\
    \subcaptionbox{}{{\includegraphics[scale= .17]{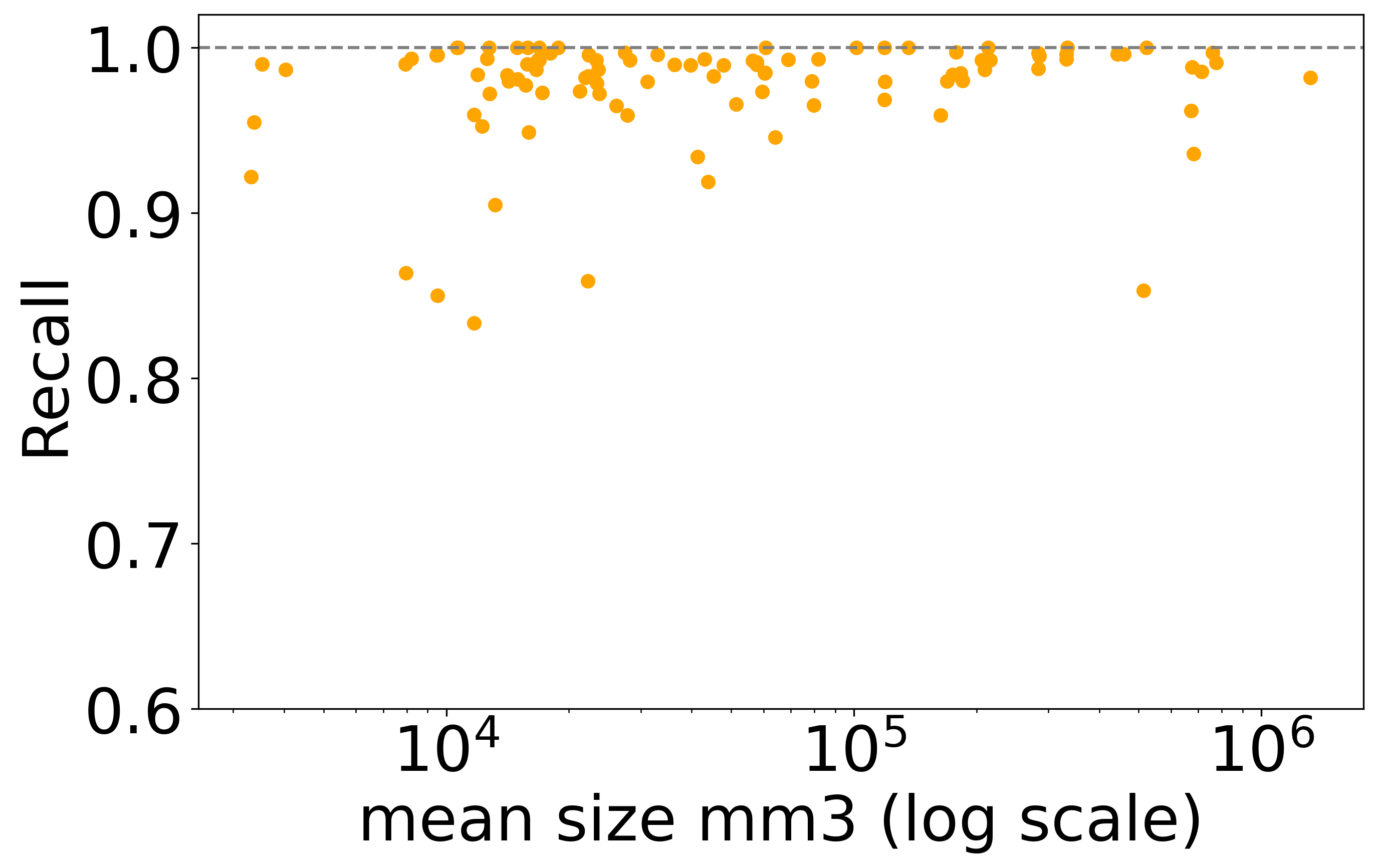}}}
    \subcaptionbox{}{{\includegraphics[scale= .17
    ]{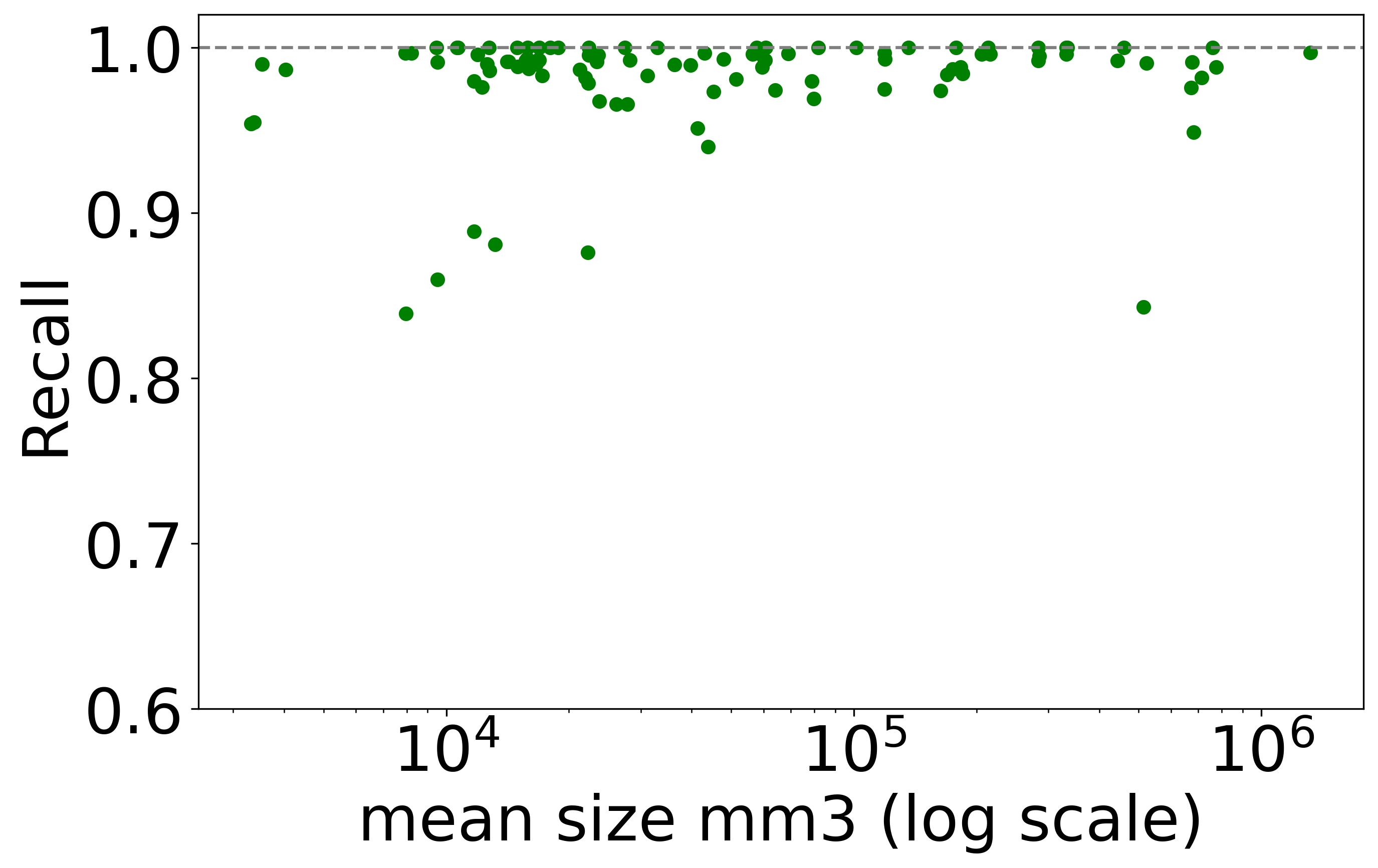}}}
    \subcaptionbox{}{{\includegraphics[scale= .17
    ]{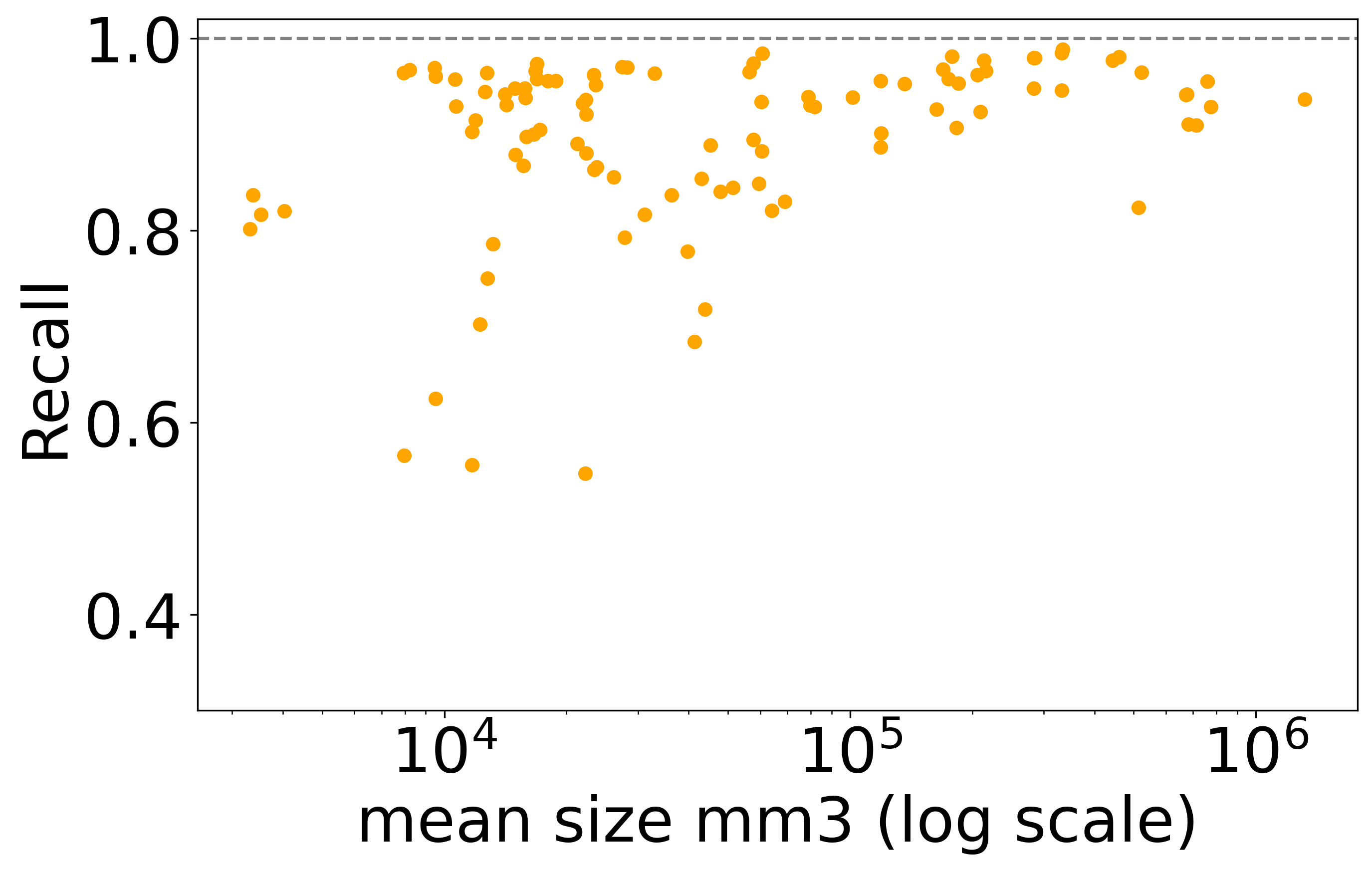}}}
    \subcaptionbox{}{{\includegraphics[scale= .17
    ]{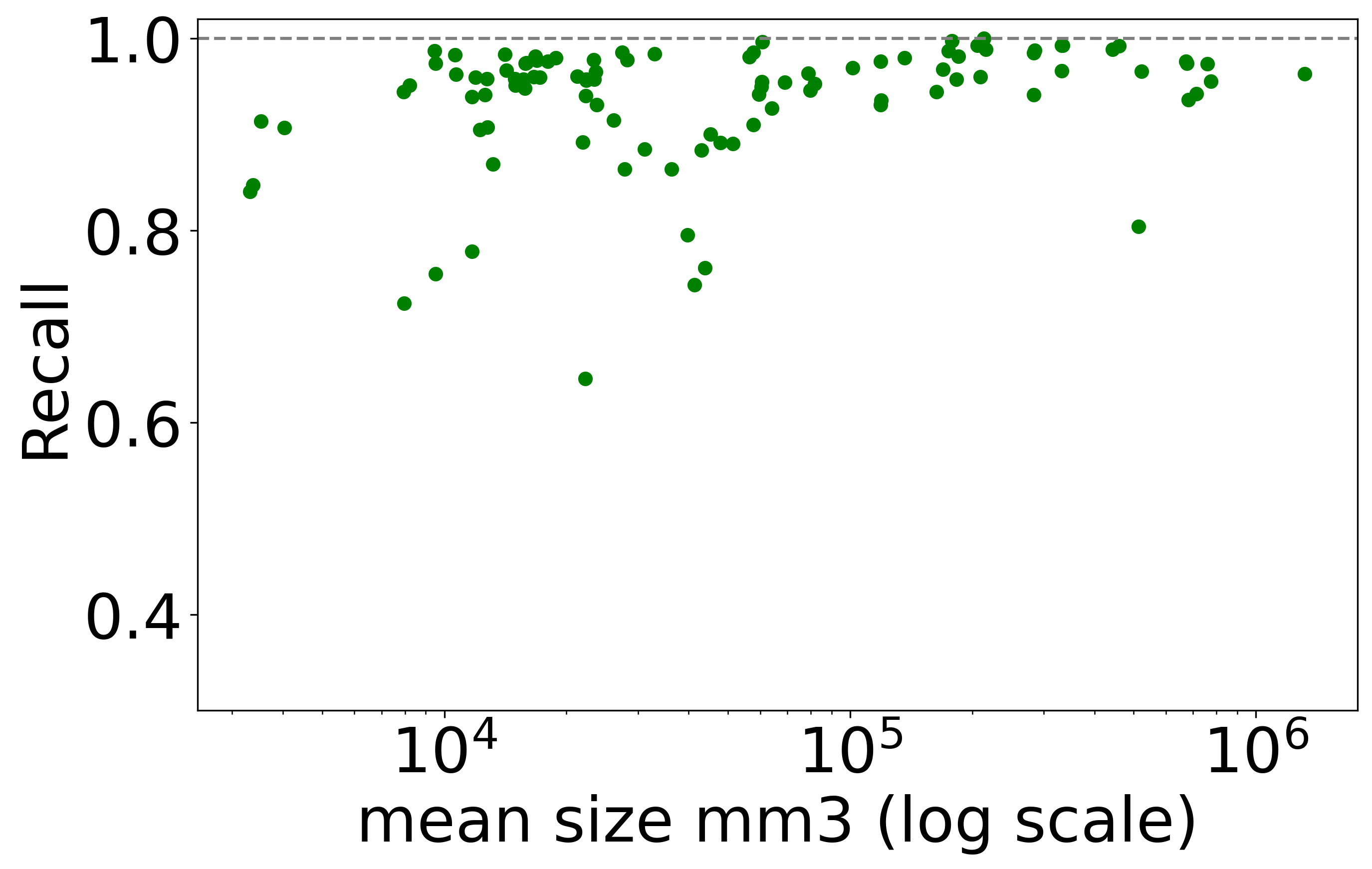}}}
    \caption{Overview of average recall vs. mean anatomical region size for 104 anatomical regions for (a) slice-wise, (b) volume-based, (c) volume-based and re-ranking, (d) region-based, (e) region-based and re-ranking, (f) localized, (g) localized and re-ranking retrieval.  }
    \label{fig:all-recall-vs-mean}
  \end{minipage}
\end{figure*}

\Cref{fig:coarser-recall-vs-mean} and \Cref{fig:all-recall-vs-mean} present an overview of mean recall for each of the retrieval methods (all models) versus the mean anatomical region size for 29 and 104 classes, respectively.
There is no pattern suggesting any correlation between the size of the anatomical region and the average retrieval recall. 

\subsection{Re-ranking}
For the first time, we could successfully adopt and show the feasibility of ColBERT-inspired re-ranking for an image retrieval task. 
In theory, this shows that CBIR results can be made subject to context-aware re-ranking. This is very important as it provides a conceptual entry point to use the information of a future retrieval solution in the real world. Concretely, observations such as user behavior on a graphical user interface, and temporal or medical relevance can be "factored in" to adjust the search results.
Further research will study the advantages and disadvantages of ColBERT-inspired re-ranking. 
In future works, further insights into balancing computational costs in the context of latency-accuracy trade-offs will be shared.

\subsection{Embeddings}
It was shown that embeddings generated from self-supervised models are slightly better for image retrieval tasks than those derived from regular supervised models. This is true for coarse anatomical regions with 29 labels (see \Cref{tab:recall-29-regions}) as well as fine-granular anatomical regions with 104 regions (see \Cref{tab:recall-104-regions}). This is roughly preserved for all modes of retrieval (i.e. slice-wise, volume-based, region-based, and localized retrieval). More generally, the differences in recall across differently pre-trained models (except pre-trained from fractal image) are very small.
Practically, the exact choice of the feature extractor should not be noticeable to a potential user in a downstream application.
Further, it can be concluded that pre-training on general natural images (i.e. ImageNet) resulted in slightly more performant embedding vectors than domain-specific images (i.e. RadImageNet). This is unexpected and subject to further research. 

Although, the model pre-trained of formula-derived synthetic images of fractals (i.e. Fractaldb) showed the lowest recall accuracy the absolute values are surprisingly high considering that the model learned visual primitives out of rendered fractals. This is very encouraging as the Formular-Driven Supervised Learning (FDSL) can easily be extended to very high number of data points per class and also several virtual classes within one family of formulas \citep{KataokaIJCV2022}. Additionally, the mathematical space of formulas for producing visual primitives is virtually infinite and thus it is the subject of further research whether radiology-specific visual primitives can be created that outperform natural image-based pre-training. Again, FDSL does not require the effort of data collection, curation, and annotation. It can scale to a large number of samples and classes which potentially results in a very smooth and evenly covered latent space.

Embeddings derived from DreamSim architecture showed the highest overall retrieval recall in region-based and localized evaluations. 
DreamSim is an ensemble architecture that uses multiple ViT embeddings with additional fine-tuning using synthetic images. 
It is plausible that an ensemble approach outperforms single-architecture embeddings (i.e. DINOv1, DINOv2, SwinTransformer, and ResNet50). 
Therefore, the usage of DreamSim is currently the preferred method of embedding generation.

Worth discussing is an observation that can be found in all tables presenting recall values. Across all model architectures (column) there are usually a few anatomies or regions (i.e. row) that show lower recall on average (see "Average" column). For example, in \Cref{tab:slicewise-recall-28-regions} "gallbladder" showed poor retrieval accuracy, whereas in Table \Cref{tab:volumewise-recall-29-regions} "brain" and "face" showed lower recall. The observation of isolated low-recall patterns can be seen across all modes of retrieval and aggregation. The authors of this paper cannot provide an explanation, as to why certain anatomies perform worse in certain retrieval configurations but gain high recall in many other retrieval configurations. This will be subject to future research.

\begin{table}
  \centering
  \caption{Summary of the average retrieval recall and standard deviation between classes for 29 anatomical regions, the boldfaced values highlight the highest recall across feature extractors. }
  \scriptsize
    \begin{tabular}{lcccccc}
    \hline
    Model & DINOv1 & DINOv2 & DreamSim & SwinTrans. & \multicolumn{2}{c}{ResNet50} \\ \hline
    Dataset (pre-trained) & (ImgNet) & (ImgNet) & (ImgNet)& (RadImg)& (Fractaldb) & (RadImg) \\ \hline
    slice-wise &  $ .855 \pm  .108$ & $ .832 \pm  .118$ & $\mathbf{ .863 \pm  .107}$ &  $ .837 \pm  .114$ & $ .51 \pm  .52$ &  $ .813 \pm   .124$ \\ 
    volume-based  & $ .949 \pm  .072$ & $ .932 \pm  .064$ & $ .936 \pm  .063$ &  $ .932 \pm   .067$ & $ .939 \pm   .078$ & $\mathbf{ .952 \pm  .043}$ \\
    volume-based re-ranked & $\mathbf{ .967 \pm  .040}$  & $ .961 \pm   .045$ &  $ .967 \pm  .045$ &  $ .961 \pm   .049$ &  $ .962 \pm   .086$ & $ .960 \pm  .050$ \\
    region-based &  $ .977 \pm  .047$ &  $ .976 \pm  .051$ & $\mathbf{ .979 \pm   .037}$ & $ .973 \pm  .042$ & $ .976 \pm   .033$ & $ .978 \pm  .039$ \\
    region-based re-ranked & $ .979 \pm   .045$ & $ .977 \pm   .050$  & $\mathbf{ .987 \pm  .027}$ & $ .977 \pm  .041$ &  $ .981 \pm   .031$ &  $ .979 \pm  .045$ \\
    localized &  $.932 \pm .105$ & $ \mathbf{.941 \pm .077}$ & $.936 \pm .089$ & $.922 \pm .099$ & $.884 \pm .111$ & $.908 \pm .099$  \\
    localized re-ranked &  $.944 \pm .099$ & $ .955 \pm .075$ & $ \mathbf{.955 \pm .062}$ & $.943 \pm .079$ & $.907 \pm .099$ & $.947 \pm .076$   \\ 
    \hline
    \label{tab:recall-29-regions}
    \end{tabular}
    
  \centering
  \caption{Summary of the average retrieval recall and standard deviation between classes for 104 anatomical regions, the boldfaced values highlight the highest recall across feature extractors.}
  \scriptsize
    \begin{tabular}{lcccccc}
    \hline
    Model & DINOv1 & DINOv2 & DreamSim & SwinTrans. & \multicolumn{2}{c}{ResNet50} \\ \hline
    Dataset (pre-trained) & (ImgNet) & (ImgNet) & (ImgNet)& (RadImg)& (Fractaldb) & (RadImg) \\ \hline
    slice-wise & $ .784 \pm  .137$ & $ .750 \pm  .144$ & $\mathbf{ .797 \pm  .129}$ &  $ .765 \pm  .140$ & $ .659 \pm  .172$ &  $ .726 \pm  .154$ \\ 
    volume-based  & $\mathbf{ .923 \pm  .077}$ & $ .887 \pm  .071$ & $ .892 \pm  .080$ &  $ .873 \pm  .082$ & $ .856 \pm  .054$ &  $ .908 \pm  .081$ \\
    volume-based re-ranked & $ .914 \pm  .040$ & $ .880 \pm  .041$ & $ .887 \pm  .040$ &  $\mathbf{ .924 \pm  .072}$ & $ .901 \pm  .061$ &  $ .902 \pm  .055$ \\
    region-based  & $ .979 \pm  .037$ & $ .972 \pm  .050$ & $\mathbf{ .983 \pm  .032}$ &  $ .978 \pm  .032$ & $ .973 \pm  .046$ &  $ .974 \pm  .042$  \\
    region-based re-ranked & $ .980 \pm  .036$ & $ .978 \pm  .042$ & $\mathbf{ .987 \pm  .024}$ &  $ .981 \pm  .033$ & $ .982 \pm  .042$ &  $ .980 \pm  .041$  \\
    localized &  $\mathbf{.929 \pm .085}$ & $.917 \pm .091$ & $.924 \pm .078$ & $.906 \pm .092$ & $.840 \pm .145$ & $.866 \pm .118$  \\
    localized re-ranked &  $ .944 \pm .075$ & $.951 \pm .071$ & $ \mathbf{.956 \pm .055}$ & $.940 \pm .066$ & $.890 \pm .105$ & $.935 \pm .070$   \\
    \hline
    \label{tab:recall-104-regions}
    \end{tabular}
\end{table}

\begin{figure*}[!htpb]
  \centering
  \begin{minipage}{ .95\textwidth}
    \centering
    \subcaptionbox{}{\includegraphics[scale= .18]{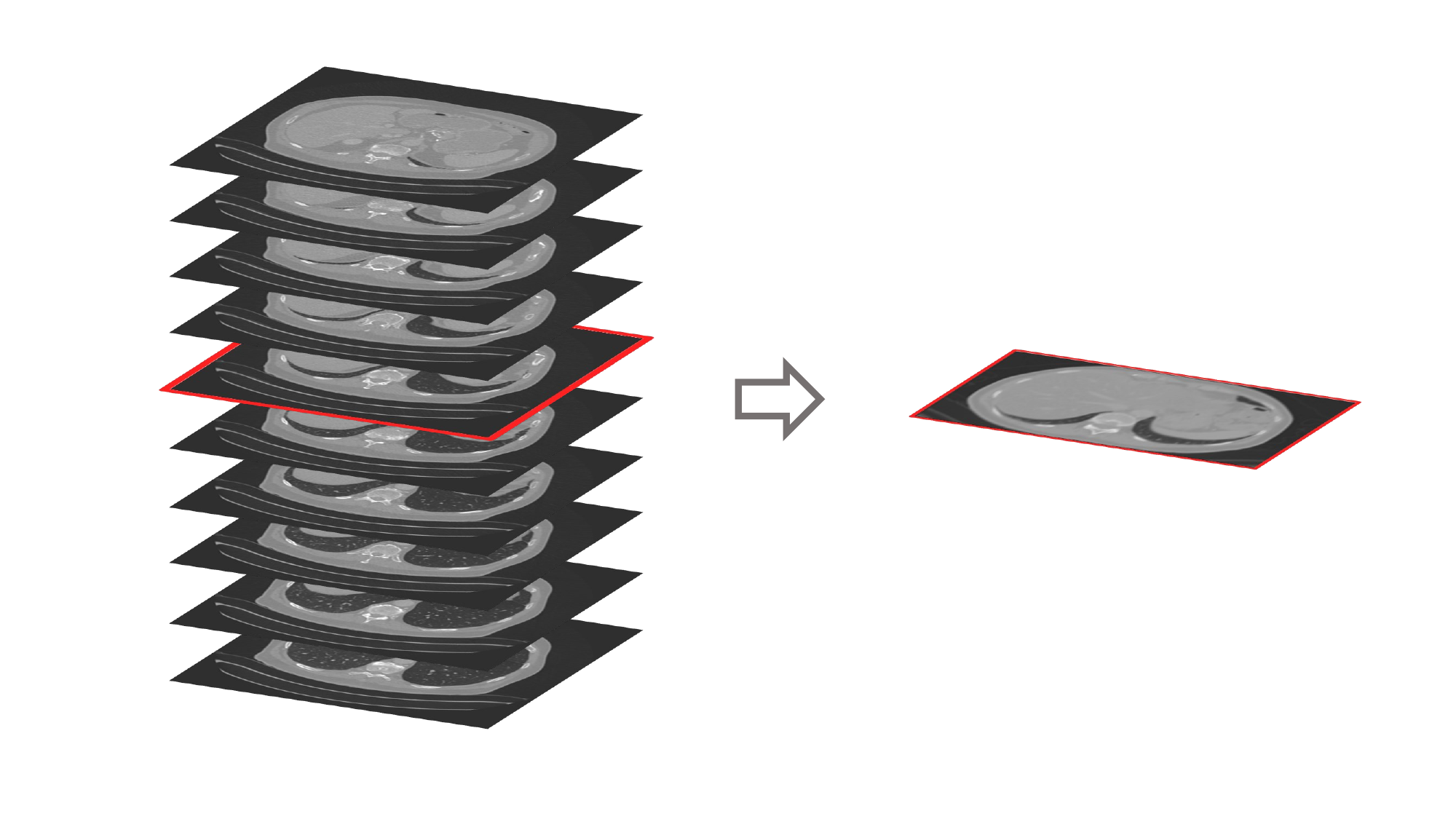}} 
    \subcaptionbox{}{{\includegraphics[scale= .18]{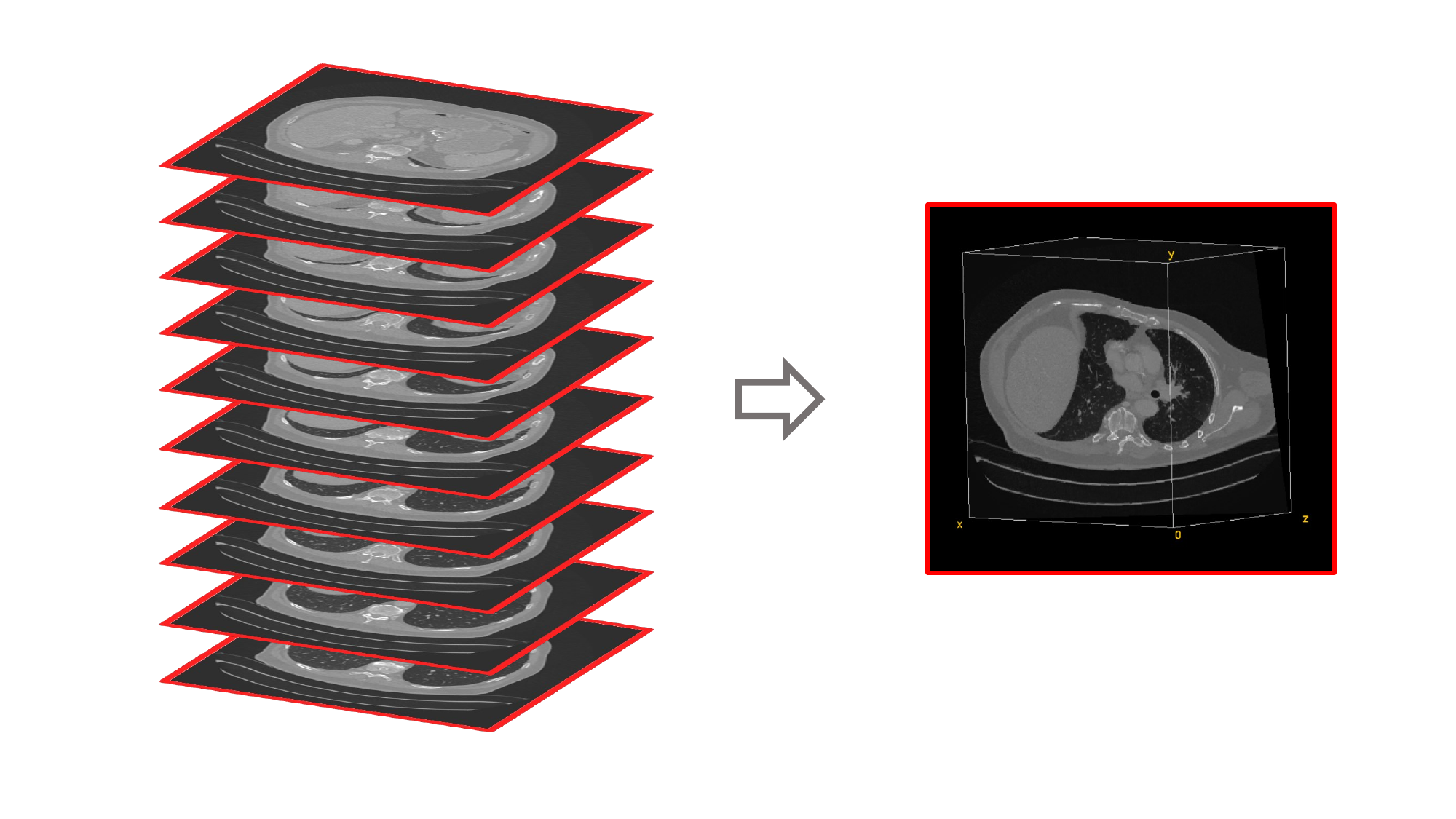}}} \\
    \subcaptionbox{}{{\includegraphics[scale= .18]{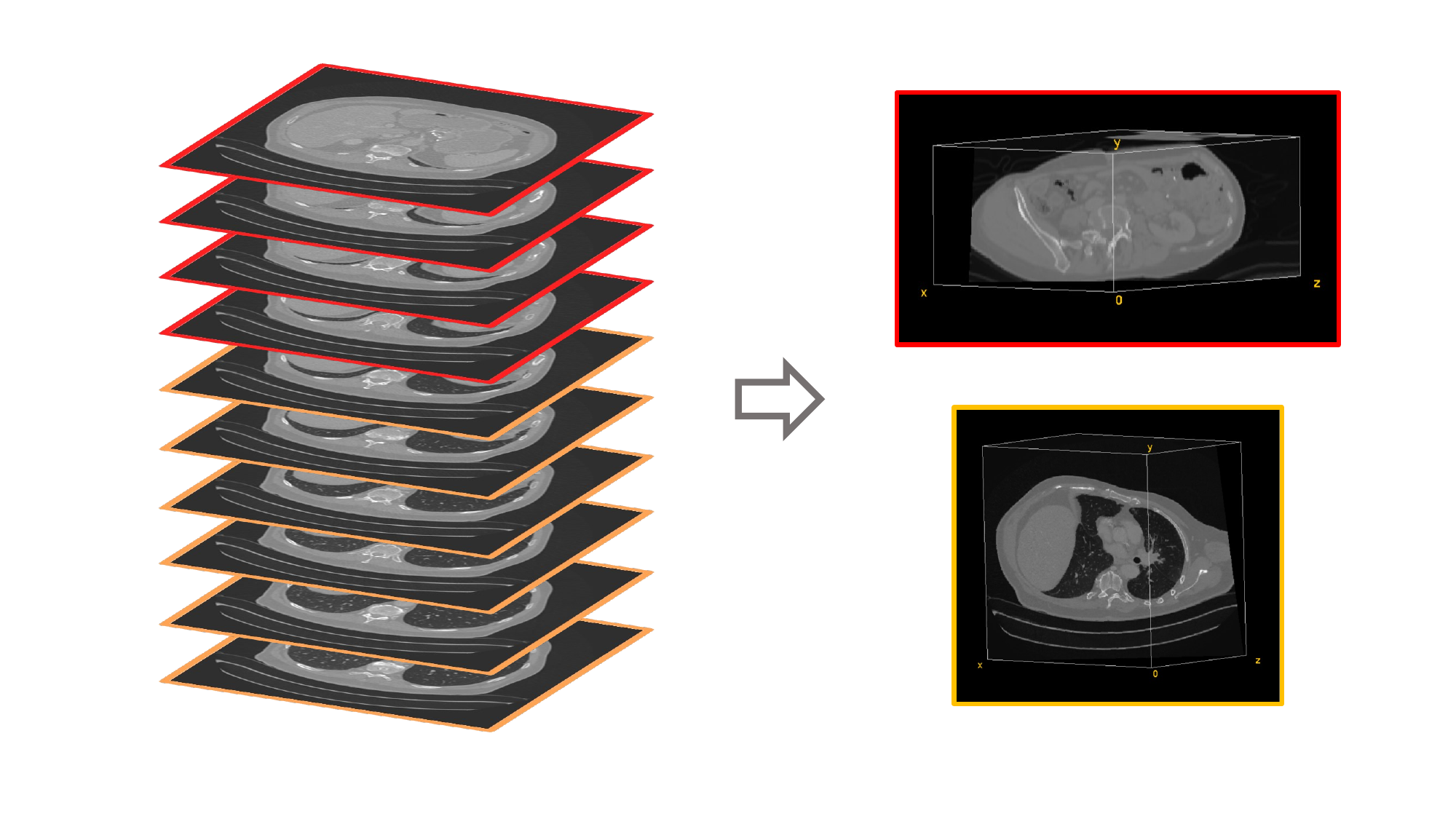}}}
    \subcaptionbox{}{{\includegraphics[scale= .18]{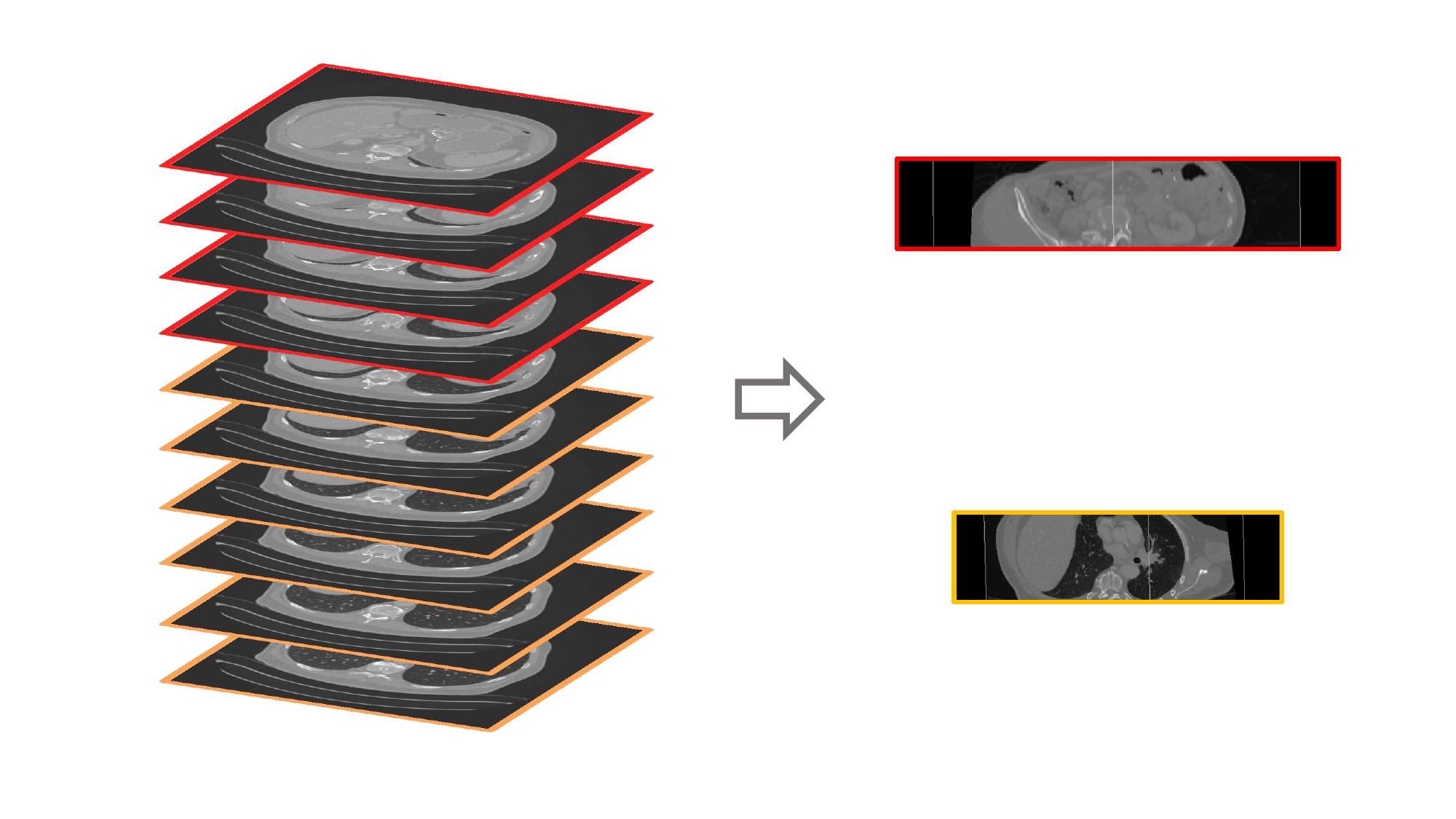}}}
    \caption{An overview of three retrieval methods: (a) Slice-wise, where the retrieval is based on one selected slice e.g., the user zooms to a slice and retrieves the most similar slice (b) volume-based, where the retrieval is based on a complete volume, e.g., the user would like to retrieve similar volumes to the volumes under examination or simply filter the database (c) region-based, where the retrieval is based on the selected organ (or sub-volume), e.g, the user zooms in to a specific region and the most similar volume containing that region is retrieved, (d) localized, where the retrieval is based on the selected organ (or sub-volume) but the region in the retrieved volume is localized to the desired organ (or sub-volume), e.g, the user zooms in to a specific region and the system returns the localized region in the retrieved volume.  }
    \label{fig:discussion-retrieval}
  \end{minipage}
\end{figure*}

\subsection{Volume-based, Region-based and Localized Retrieval}
Since multiple organs (i.e. labels) are present in each query volume, there are essentially different ways in which image retrieval can be performed. 
The preferred choice depends on the context of the retrieval task in the real world. 
If the goal is to find a scan out of a database that is most similar to a complete query scan with the entirety of all present organs (think scan-id to scan-id but visual), then volume-based retrieval is the right choice. 
In contrast, if the experimenter is interested in a particular organ and its most similar counterpart in the database (and all other organs just happen to be in the same scan due to proximity), then region-based retrieval or localized retrieval is advised. 
Slice-wise retrieval can find the most similar slice of a volume regardless of the number of other slices. 
This is not usually a practical choice in real scenarios.
\Cref{fig:discussion-retrieval} visualizes the options.

\subsection{Localization-ratio}
\label{sec:discussion,localization-ratio}

\begin{figure}
  \centering
  \includegraphics[trim=5cm 3.5cm 5cm 3.5cm, clip, width=0.70\textwidth]{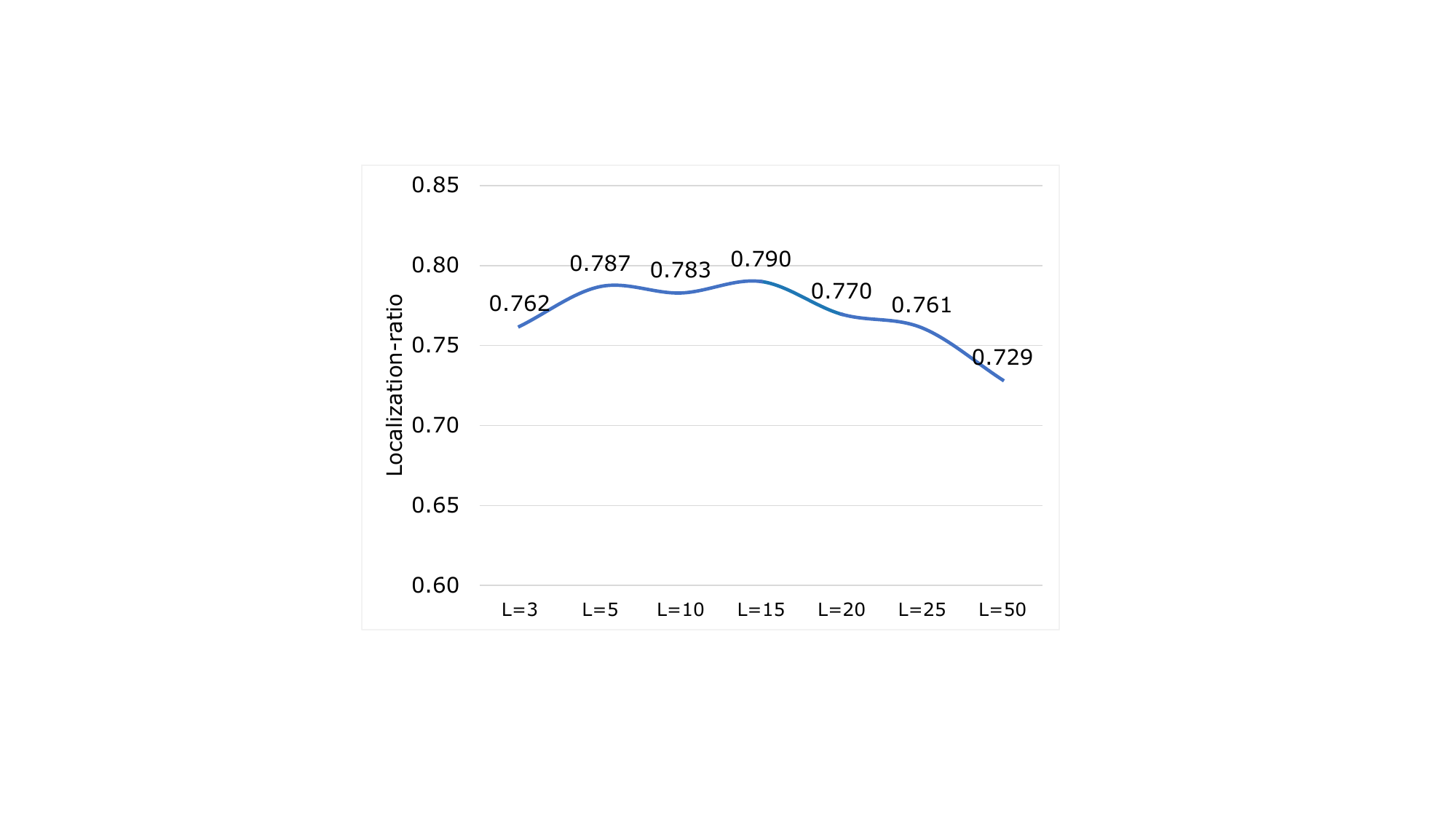}
  \caption{Average localization-ratio for different L values for re-ranking evaluation using DreamSim model (best-performing embedding) based on \eqref{eq:normalized-location-ratio-rerank-version}.}
  \label{fig:localization-ratio-L-experiment}
\end{figure}

\Cref{tab:localization-ratio-summary} shows the average localization-ratio for with and without re-ranking ($L=15$). 
There is a drop for both 29 coarse and 104 original TS classes after re-ranking. 
However, both with and without re-ranking the localization-ratio is high, indicating that most of the slices contributing to the final retrieval of the volumes actually contain the desired anatomical region. 
However, based on \eqref{eq:normalized-location-ratio-rerank-version} the choice of $L$ can impact this measure. 
\Cref{fig:localization-ratio-L-experiment} shows localization-ratio for different values of $L$ as shown increasing the $L$ to more that $L=15$ decreases the localization-ratio. 
This indicates that the first highest values in vector $m_{SIM}$ based on \eqref{eq:msim} points to the exact slices that contain the desired anatomical structure. 
To improve the re-ranking localization-ratio, further studies can focus on optimizing the $L$ value based on heuristic, organ size, etc. 

\begin{table}
  \centering
  \caption{Summary of the average localization-ratio and standard deviation between classes for 29 and 104 anatomical regions for $L=15$, the boldfaced values highlight the highest localization-ratio across feature extractors. }
  \scriptsize
    \begin{tabular}{lcccccc}
    \hline
    Model & DINOv1 & DINOv2 & DreamSim & SwinTrans. & \multicolumn{2}{c}{ResNet50} \\ \hline
    Dataset (pre-trained) & (ImgNet) & (ImgNet) & (ImgNet)& (RadImg)& (Fractaldb) & (RadImg) \\ \hline
    localization-ratio (29 anatomical regions) & $0.842 \pm 0.161$ & $0.834 \pm 0.144$ & $\mathbf{0.864 \pm 0.145}$ & $0.835 \pm 0.155$ & $0.800 \pm 0.168$ & $0.815 \pm 0.152$   \\
    localization-ratio (104 anatomical regions) &  $0.793 \pm 0.140$ & $0.763 \pm 0.148$ & $\mathbf{0.803 \pm 0.130}$ & $0.773 \pm 0.143$ & $0.722 \pm 0.178$ & $0.736 \pm 0.162$  \\ 
    localization-ratio re-ranked (29 anatomical regions) & $0.823 \pm 0.169$ & $0.803 \pm 0.157$ & $\mathbf{0.837 \pm 0.159}$ & $0.792 \pm 0.164$ & $0.727 \pm 0.195$ & $0.787 \pm 0.171$   \\
    localization-ratio re-ranked (104 anatomical regions) & $0.764 \pm 0.148$ & $0.754 \pm 0.150$ & $\mathbf{0.790 \pm 0.142}$ & $0.736 \pm 0.145$ & $0.662 \pm 0.171$ & $0.720 \pm 0.166$ \\ 
    \hline
    \label{tab:localization-ratio-summary}
    \end{tabular}
\end{table}

\section{Conclusion}

Our study establishes a new benchmark for the retrieval of anatomical structures within 3D medical volumes, utilizing the TotalSegmentator dataset to facilitate targeted queries of volumes or sub-volumes for specific anatomical structures. 
The results highlight the potential of leveraging pre-trained vision embeddings, originally trained on non-medical images, for medical image retrieval across diverse anatomical regions with a wide size range. 

We introduced a re-ranking method based on a late interaction model from text retrieval, i.e. ColBERT \cite{khattab2020colbert}. The proposed ColBERT-inspired method enhances the retrieval recall of all anatomical regions. 
Future investigations can focus on refining and optimizing the computational efficiency of the proposed re-ranking method.

We evaluated the performance of different embeddings pre-trained supervised and self-supervised on medical and non-medical data. 
The results indicate that pre-training on general natural images (e.g., ImageNet) yields slightly more effective embedding vectors than domain-specific natural images (e.g., RadImageNet). However, given the marginal difference, the choice of embeddings is unlikely to impact the user experience in downstream tasks significantly.

The retrieval of certain anatomical structures, such as the brain and face, demonstrates low recall across all embedding and retrieval methods. 
Subsequent research can explore the prevalence of such patterns and potential solutions.

This benchmark sets the stage for future advancements in content-based medical image retrieval, particularly in localizing specific organs or areas within scans.

\section*{Acknowledgement}
The authors like to thank the Bayer team of the internal ML innovation platform for providing compute infrastructure and technical support. 

We thank Timothy Deyer and his RadImageNet team for providing the RadImageNet pre-trained model weights for the SwinTransformer architecture.

\bibliographystyle{unsrtnat}
\bibliography{references}  

\begin{thebibliography}{28}
\providecommand{\natexlab}[1]{#1}
\providecommand{\url}[1]{\texttt{#1}}
\expandafter\ifx\csname urlstyle\endcsname\relax
  \providecommand{\doi}[1]{doi: #1}\else
  \providecommand{\doi}{doi: \begingroup \urlstyle{rm}\Url}\fi

\bibitem[Dubey(2021)]{dubey2021decade}
Shiv~Ram Dubey.
\newblock A decade survey of content based image retrieval using deep learning.
\newblock \emph{IEEE Transactions on Circuits and Systems for Video Technology}, 32\penalty0 (5):\penalty0 2687--2704, 2021.

\bibitem[Wang et~al.(2022)Wang, Jiao, Liu, Ma, and Shang]{wang2022two}
Wenqing Wang, Pengfei Jiao, Han Liu, Xiao Ma, and Zhuo Shang.
\newblock Two-stage content based image retrieval using sparse representation and feature fusion.
\newblock \emph{Multimedia Tools and Applications}, 81\penalty0 (12):\penalty0 16621--16644, 2022.

\bibitem[Qayyum et~al.(2017)Qayyum, Anwar, Awais, and Majid]{qayyum2017medical}
Adnan Qayyum, Syed~Muhammad Anwar, Muhammad Awais, and Muhammad Majid.
\newblock Medical image retrieval using deep convolutional neural network.
\newblock \emph{Neurocomputing}, 266:\penalty0 8--20, 2017.

\bibitem[Khun~Jush et~al.(2023)Khun~Jush, Truong, Vogler, and Lenga]{jush2023medical}
Farnaz Khun~Jush, Tuan Truong, Steffen Vogler, and Matthias Lenga.
\newblock Medical image retrieval using pretrained embeddings.
\newblock \emph{arXiv preprint arXiv:2311.13547}, 2023.

\bibitem[Abacha et~al.(2023)Abacha, Santamaria-Pang, Lee, Merkow, Cai, Devarakonda, Islam, Gong, Lungren, Lin, et~al.]{abacha20233d}
Asma~Ben Abacha, Alberto Santamaria-Pang, Ho~Hin Lee, Jameson Merkow, Qin Cai, Surya~Teja Devarakonda, Abdullah Islam, Julia Gong, Matthew~P Lungren, Thomas Lin, et~al.
\newblock 3d-mir: A benchmark and empirical study on 3d medical image retrieval in radiology.
\newblock \emph{arXiv preprint arXiv:2311.13752}, 2023.

\bibitem[Denner et~al.(2024)Denner, Zimmerer, Bounias, Bujotzek, Xiao, Kausch, Schader, Penzkofer, J{\"a}ger, and Maier-Hein]{denner2024leveraging}
Stefan Denner, David Zimmerer, Dimitrios Bounias, Markus Bujotzek, Shuhan Xiao, Lisa Kausch, Philipp Schader, Tobias Penzkofer, Paul~F J{\"a}ger, and Klaus Maier-Hein.
\newblock Leveraging foundation models for content-based medical image retrieval in radiology.
\newblock \emph{arXiv preprint arXiv:2403.06567}, 2024.

\bibitem[Truong et~al.(2023)Truong, Jush, and Lenga]{truong2023benchmarking}
Tuan Truong, Farnaz~Khun Jush, and Matthias Lenga.
\newblock Benchmarking pretrained vision embeddings for near-and duplicate detection in medical images.
\newblock \emph{arXiv preprint arXiv:2312.07273}, 2023.

\bibitem[Antonelli et~al.(2022)Antonelli, Reinke, Bakas, Farahani, Kopp-Schneider, Landman, Litjens, Menze, Ronneberger, Summers, et~al.]{antonelli2022medical}
Michela Antonelli, Annika Reinke, Spyridon Bakas, Keyvan Farahani, Annette Kopp-Schneider, Bennett~A Landman, Geert Litjens, Bjoern Menze, Olaf Ronneberger, Ronald~M Summers, et~al.
\newblock The medical segmentation decathlon.
\newblock \emph{Nature communications}, 13\penalty0 (1):\penalty0 4128, 2022.

\bibitem[Wasserthal et~al.(2023)Wasserthal, Breit, Meyer, Pradella, Hinck, Sauter, Heye, Boll, Cyriac, Yang, et~al.]{wasserthal2023totalsegmentator}
Jakob Wasserthal, Hanns-Christian Breit, Manfred~T Meyer, Maurice Pradella, Daniel Hinck, Alexander~W Sauter, Tobias Heye, Daniel~T Boll, Joshy Cyriac, Shan Yang, et~al.
\newblock Totalsegmentator: Robust segmentation of 104 anatomic structures in ct images.
\newblock \emph{Radiology: Artificial Intelligence}, 5\penalty0 (5), 2023.

\bibitem[Khattab and Zaharia(2020)]{khattab2020colbert}
Omar Khattab and Matei Zaharia.
\newblock Colbert: Efficient and effective passage search via contextualized late interaction over bert.
\newblock In \emph{Proceedings of the 43rd International ACM SIGIR conference on research and development in Information Retrieval}, pages 39--48, 2020.

\bibitem[Aum{\"u}ller et~al.(2020)Aum{\"u}ller, Bernhardsson, and Faithfull]{aumuller2020ann}
Martin Aum{\"u}ller, Erik Bernhardsson, and Alexander Faithfull.
\newblock Ann-benchmarks: A benchmarking tool for approximate nearest neighbor algorithms.
\newblock \emph{Information Systems}, 87:\penalty0 101374, 2020.

\bibitem[Charikar(2002)]{charikar2002similarity}
Moses~S Charikar.
\newblock Similarity estimation techniques from rounding algorithms.
\newblock In \emph{Proceedings of the thiry-fourth annual ACM symposium on Theory of computing}, pages 380--388, 2002.

\bibitem[Malkov and Yashunin(2018)]{malkov2018efficient}
Yu~A Malkov and Dmitry~A Yashunin.
\newblock Efficient and robust approximate nearest neighbor search using hierarchical navigable small world graphs.
\newblock \emph{IEEE transactions on pattern analysis and machine intelligence}, 42\penalty0 (4):\penalty0 824--836, 2018.

\bibitem[Taha et~al.(2024)Taha, Lissandrini, Simitsis, and Ioannidis]{taha2024study}
Ibraheem Taha, Matteo Lissandrini, Alkis Simitsis, and Yannis Ioannidis.
\newblock A study on efficient indexing for table search in data lakes.
\newblock In \emph{2024 IEEE 18th International Conference on Semantic Computing (ICSC)}, pages 245--252. IEEE, 2024.

\bibitem[Johnson et~al.(2019)Johnson, Douze, and J{\'e}gou]{johnson2019billion}
Jeff Johnson, Matthijs Douze, and Herv{\'e} J{\'e}gou.
\newblock Billion-scale similarity search with gpus.
\newblock \emph{IEEE Transactions on Big Data}, 7\penalty0 (3):\penalty0 535--547, 2019.

\bibitem[Deng et~al.(2009)Deng, Dong, Socher, Li, Li, and Fei-Fei]{deng2009imagenet}
Jia Deng, Wei Dong, Richard Socher, Li-Jia Li, Kai Li, and Li~Fei-Fei.
\newblock Imagenet: A large-scale hierarchical image database.
\newblock In \emph{2009 IEEE conference on computer vision and pattern recognition}, pages 248--255. Ieee, 2009.

\bibitem[Caron et~al.(2021)Caron, Touvron, Misra, J{\'e}gou, Mairal, Bojanowski, and Joulin]{caron2021emerging}
Mathilde Caron, Hugo Touvron, Ishan Misra, Herv{\'e} J{\'e}gou, Julien Mairal, Piotr Bojanowski, and Armand Joulin.
\newblock Emerging properties in self-supervised vision transformers.
\newblock In \emph{Proceedings of the IEEE/CVF international conference on computer vision}, pages 9650--9660, 2021.

\bibitem[Oquab et~al.(2023)Oquab, Darcet, Moutakanni, Vo, Szafraniec, Khalidov, Fernandez, Haziza, Massa, El-Nouby, et~al.]{oquab2023dinov2}
Maxime Oquab, Timoth{\'e}e Darcet, Th{\'e}o Moutakanni, Huy Vo, Marc Szafraniec, Vasil Khalidov, Pierre Fernandez, Daniel Haziza, Francisco Massa, Alaaeldin El-Nouby, et~al.
\newblock Dinov2: Learning robust visual features without supervision.
\newblock \emph{arXiv preprint arXiv:2304.07193}, 2023.

\bibitem[Fu et~al.(2023)Fu, Tamir, Sundaram, Chai, Zhang, Dekel, and Isola]{fu2023dreamsim}
Stephanie Fu, Netanel Tamir, Shobhita Sundaram, Lucy Chai, Richard Zhang, Tali Dekel, and Phillip Isola.
\newblock Dreamsim: Learning new dimensions of human visual similarity using synthetic data.
\newblock \emph{arXiv preprint arXiv:2306.09344}, 2023.

\bibitem[Liu et~al.(2021)Liu, Lin, Cao, Hu, Wei, Zhang, Lin, and Guo]{liu2021swin}
Ze~Liu, Yutong Lin, Yue Cao, Han Hu, Yixuan Wei, Zheng Zhang, Stephen Lin, and Baining Guo.
\newblock Swin transformer: Hierarchical vision transformer using shifted windows.
\newblock In \emph{Proceedings of the IEEE/CVF international conference on computer vision}, pages 10012--10022, 2021.

\bibitem[He et~al.(2016)He, Zhang, Ren, and Sun]{he2016deep}
Kaiming He, Xiangyu Zhang, Shaoqing Ren, and Jian Sun.
\newblock Deep residual learning for image recognition.
\newblock In \emph{Proceedings of the IEEE conference on computer vision and pattern recognition}, pages 770--778, 2016.

\bibitem[Mei et~al.(2022)Mei, Liu, Robson, Marinelli, Huang, Doshi, Jacobi, Cao, Link, Yang, et~al.]{mei2022radimagenet}
Xueyan Mei, Zelong Liu, Philip~M Robson, Brett Marinelli, Mingqian Huang, Amish Doshi, Adam Jacobi, Chendi Cao, Katherine~E Link, Thomas Yang, et~al.
\newblock Radimagenet: an open radiologic deep learning research dataset for effective transfer learning.
\newblock \emph{Radiology: Artificial Intelligence}, 4\penalty0 (5):\penalty0 e210315, 2022.

\bibitem[Kataoka et~al.(2022)Kataoka, Okayasu, Matsumoto, Yamagata, Yamada, Inoue, Nakamura, and Satoh]{KataokaIJCV2022}
Hirokatsu Kataoka, Kazushige Okayasu, Asato Matsumoto, Eisuke Yamagata, Ryosuke Yamada, Nakamasa Inoue, Akio Nakamura, and Yutaka Satoh.
\newblock Pre-training without natural images.
\newblock \emph{International Journal of Computer Vision (IJCV)}, 2022.

\bibitem[Ai et~al.(2018)Ai, Mao, Liu, and Croft]{ai2018unbiased}
Qingyao Ai, Jiaxin Mao, Yiqun Liu, and W~Bruce Croft.
\newblock Unbiased learning to rank: Theory and practice.
\newblock In \emph{Proceedings of the 27th ACM International Conference on Information and Knowledge Management}, pages 2305--2306, 2018.

\bibitem[Guo et~al.(2020)Guo, Fan, Pang, Yang, Ai, Zamani, Wu, Croft, and Cheng]{guo2020deep}
Jiafeng Guo, Yixing Fan, Liang Pang, Liu Yang, Qingyao Ai, Hamed Zamani, Chen Wu, W~Bruce Croft, and Xueqi Cheng.
\newblock A deep look into neural ranking models for information retrieval.
\newblock \emph{Information Processing \& Management}, 57\penalty0 (6):\penalty0 102067, 2020.

\bibitem[MacAvaney et~al.(2019)MacAvaney, Yates, Cohan, and Goharian]{macavaney2019cedr}
Sean MacAvaney, Andrew Yates, Arman Cohan, and Nazli Goharian.
\newblock Cedr: Contextualized embeddings for document ranking.
\newblock In \emph{Proceedings of the 42nd international ACM SIGIR conference on research and development in information retrieval}, pages 1101--1104, 2019.

\bibitem[Devlin et~al.(2018)Devlin, Chang, Lee, and Toutanova]{devlin2018bert}
Jacob Devlin, Ming-Wei Chang, Kenton Lee, and Kristina Toutanova.
\newblock Bert: Pre-training of deep bidirectional transformers for language understanding.
\newblock \emph{arXiv preprint arXiv:1810.04805}, 2018.

\bibitem[Santhanam et~al.(2021)Santhanam, Khattab, Saad-Falcon, Potts, and Zaharia]{santhanam2021colbertv2}
Keshav Santhanam, Omar Khattab, Jon Saad-Falcon, Christopher Potts, and Matei Zaharia.
\newblock Colbertv2: Effective and efficient retrieval via lightweight late interaction.
\newblock \emph{arXiv preprint arXiv:2112.01488}, 2021.

\end{thebibliography}

\end{document}